\documentclass[letterpaper]{article}
\usepackage[preprint]{aaai2027}

\usepackage[hyphens]{url}
\usepackage{graphicx}
\usepackage{natbib}
\usepackage{caption}
\usepackage[dvipsnames,table]{xcolor}
\usepackage{algorithm}
\usepackage{algorithmic}
\usepackage{amsmath}
\usepackage{amssymb}
\usepackage{booktabs}
\usepackage{makecell}
\usepackage{multirow}
\usepackage{pifont}
\usepackage{placeins}
\usepackage{soul}
\usepackage{subcaption}
\usepackage{tabularx}
\usepackage[capitalize]{cleveref}

\newcommand{\best}[1]{\textbf{#1}}
\newcommand{\second}[1]{\ul{#1}}
\newcommand{\declareappendixref}[2]{\expandafter\def\csname appendixletter@#1\endcsname{#2}}
\declareappendixref{sec:suppl_setup}{A}
\declareappendixref{sec:suppl_method_impl}{B}
\declareappendixref{sec:suppl_results}{C}
\declareappendixref{sec:suppl_efficiency}{C}
\declareappendixref{sec:suppl_ablation}{D}
\declareappendixref{sec:suppl_empirical}{E}
\newcommand{\appendixref}[1]{Appendix~\csname appendixletter@#1\endcsname}

\title{STAR-Pro: Stage-Wise Token Adaptive Reduction with Progressive Refinement for Efficient Large Vision-Language Models}
\author{
Yichen Guo\textsuperscript{\rm 1,\rm 2}\equalcontrib,
Tinghao Wang\textsuperscript{\rm 1,\rm 3}\equalcontrib,
Qizhe Zhang\textsuperscript{\rm 1}\equalcontrib,
Lingbei Meng\textsuperscript{\rm 5},
Yuan Zhang\textsuperscript{\rm 1},
Jiajun Cao\textsuperscript{\rm 1},\\
Hao Jiang\textsuperscript{\rm 3},
Chenwei Wu\textsuperscript{\rm 4},
Jixian Wu\textsuperscript{\rm 1},
Sixiang Chen\textsuperscript{\rm 1},
Tao Luo\textsuperscript{\rm 3},
Hongyang Cheng\textsuperscript{\rm 1},\\
Kai Tang\textsuperscript{\rm 1,\rm 2},
Chenxi Li\textsuperscript{\rm 5},
Renyuan Li\textsuperscript{\rm 3},
Xiande Huang\textsuperscript{\rm 5},
Wenya Wang\textsuperscript{\rm 2},
Shanghang Zhang\textsuperscript{\rm 1}\thanks{Corresponding author: shanghang@pku.edu.cn}
}
\affiliations{
\textsuperscript{\rm 1}State Key Laboratory of Multimedia Information Processing, School of Computer Science, Peking University\\
\textsuperscript{\rm 2}Nanyang Technological University, Singapore\quad
\textsuperscript{\rm 3}University of Electronic Science and Technology of China\\
\textsuperscript{\rm 4}University of Michigan, Ann Arbor\quad
\textsuperscript{\rm 5}De Artificial Intelligence Lab\\
\textbf{Code:} \url{https://github.com/EasonAI-5589/starpro}
}

\begin{document}

\maketitle

\begin{abstract}
Large vision-language models (LVLMs) achieve strong multimodal understanding, but the hundreds to thousands of visual tokens they process impose substantial computational overhead, motivating training-free visual token pruning. In this work, we conduct two complementary analyses of visual token pruning. First, we measure the feature-space coverage of tokens retained before cross-modal fusion and find that aggressive pruning discards substantial visual information. Second, we track text-to-visual attention across decoder layers and find that the visual tokens considered important change substantially with depth, making one-shot pruning decisions unreliable. Together, these findings show that effective pruning should preserve broad visual coverage before fusion and progressively refine the retained tokens as cross-modal evidence evolves during fusion. We therefore propose \textbf{STAR-Pro} (\textbf{ST}age-Wise \textbf{A}daptive Token \textbf{R}eduction with \textbf{Pro}gressive Refinement), a training-free two-stage framework. Its Adaptive Stage applies pivoted QR to construct an over-budget feature-coverage candidate pool, while its Progressive Stage uses evolving text-to-visual attention at selected decoder layers to prune a nested survivor set under a target layer-average token budget. Extensive experiments across seven LVLMs spanning multiple architectures and 18 image and video benchmarks demonstrate the effectiveness of STAR-Pro under aggressive pruning. On LLaVA-Video-7B, STAR-Pro reduces visual tokens by \textbf{90.5\%}, retains \textbf{92.7\%} of baseline performance, and achieves a \textbf{2.24$\times$} measured inference speedup. Code is available at \url{https://github.com/EasonAI-5589/starpro}.
\end{abstract}

\begin{figure}[!t]
    \centering
    \includegraphics[width=\linewidth]{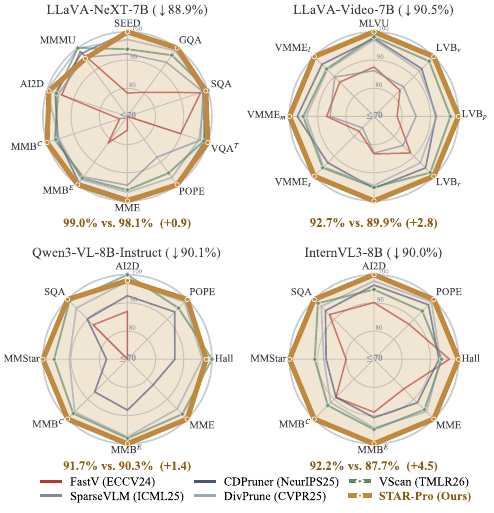}
    \caption{STAR-Pro at ${\sim}90\%$ visual token pruning. Each radar axis is normalised to the strongest plotted method (100); below each panel: STAR-Pro vs.\ the strongest baseline in the corresponding main table, both as percentages of baseline performance.}
    \label{fig:teaser}
    \vspace{-8pt}
\end{figure}

\section{Introduction}

\begin{figure*}[t]
    \centering
    \includegraphics[width=\linewidth]{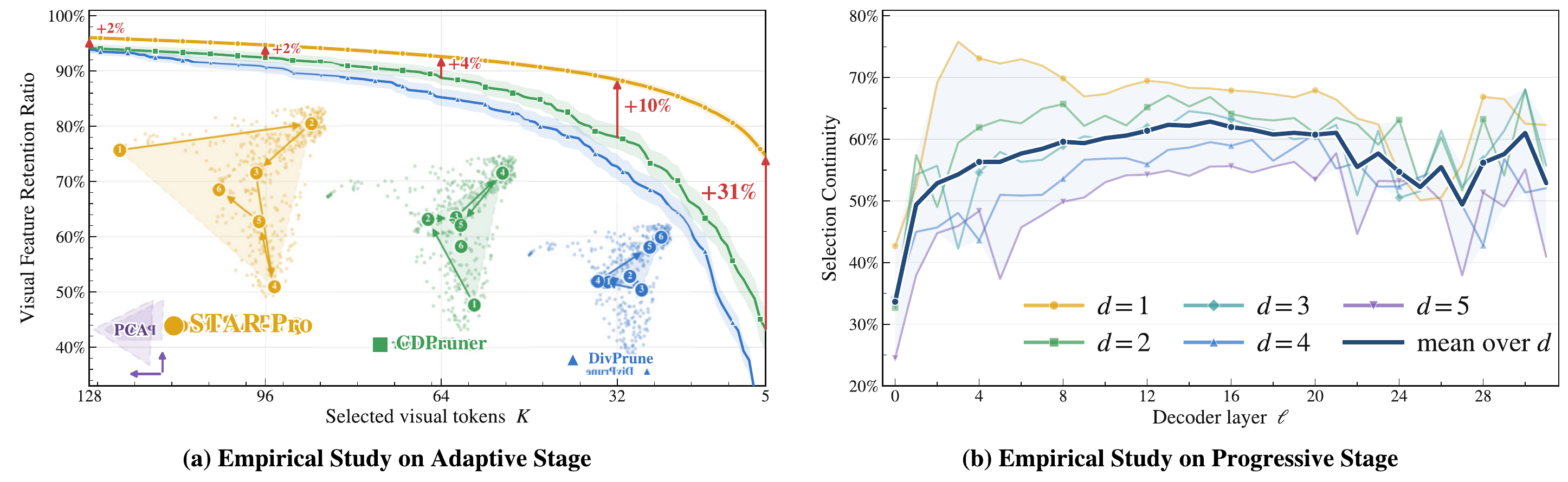}
    \caption{Empirical Study.
    (a) Empirical Study on Adaptive Stage. We measure the visual feature retention ratio $\eta{=}\|P_{S}X\|_F^{2}/\|X\|_F^{2}$ of the $K$ retained tokens. Across all budgets, STAR-Pro preserves more visual information than CDPruner and DivPrune, with a larger advantage under more aggressive pruning. The PCA insets visualize the first six selections of each method for one image: STAR-Pro spans the largest region of the feature space and reaches the most distinct feature clusters.
    (b) Empirical Study on Progressive Stage. At each decoder layer $\ell$, we measure the IoU continuity of the top-$K$ attended set $S_\ell$ with layers a distance $d$ away, defined as $\tfrac{1}{2}[\mathrm{IoU}(S_\ell,S_{\ell-d})+\mathrm{IoU}(S_\ell,S_{\ell+d})]$. Across gaps $d{=}1$--$5$, continuity stays well below~$1$, confirming visual token importance changes with depth (LLaVA-1.5-7B, $K{=}64$, mean over 500 MME images).}
    \label{fig:empirical_study}
    \vspace{-8pt}
\end{figure*}

With the rapid development of Large Language Models (LLMs) \cite{grattafiori2024llama3,achiam2023gpt4,team2023gemini,yang2025qwen3,liu2024deepseek}, their outstanding reasoning abilities have enabled remarkable advances in Large Vision-Language Models (LVLMs) \cite{liu2023llava, liu2024llavanext, bai2025qwen25vl, zhu2025internvl3}. By integrating vision encoders (e.g., ViT \cite{dosovitskiy2020imageViT}) with LLMs, recent LVLMs demonstrate strong performance on multimodal tasks \cite{goyal2017makingVQAV2,hudson2019gqa,gurariVizWizGrandChallenge2018,lu2022learnScienceQA,singhVQAModelsThat2019,fu2023mme,liuMMBenchYourMultimodal2024,yu2023mmvet}. However, a typical LVLM generates hundreds to thousands of visual tokens per image, introducing substantial computational overhead. This problem is exacerbated in high-resolution and video scenarios, where token number dramatically escalates.

Existing training-free visual token pruning methods mainly operate either before or within LLM. Pre-LLM approaches, such as LLaVA-PruMerge~\cite{shang2024llava-prumerge}, VisionZip~\cite{yangVisionZipLongerBetter2024}, DivPrune~\cite{alvarDivPruneDiversitybasedVisual2025}, and CDPruner~\cite{zhangAttentionSimilarityMaximizing2025}, compress the visual sequence before decoder interaction. We assess whether their retained tokens preserve the original visual information from two complementary perspectives. Quantitatively, visual feature retention ratio $\eta{=}\|P_{S}X\|_F^{2}/\|X\|_F^{2}$ measures the fraction of the full visual feature information kept by the subspace spanned by the retained tokens (applicability is discussed in \appendixref{sec:suppl_empirical}). Geometrically, a 2D PCA of the visual token features reveals whether the selected tokens span distinct feature groups. As shown in \cref{fig:empirical_study}(a), representative pre-LLM selectors exhibit a sharp decline in feature retention ratio as the token budget shrinks and, under aggressive compression, increasingly fail to span different groups in the PCA space. Together, these observations show that aggressive pre-LLM pruning loses complementary visual token information before cross-modal reasoning begins, motivating a coverage-oriented solution. Within-LLM methods, including FastV~\cite{chen2024imagefastv}, PyramidDrop~\cite{xing2024pyramiddrop}, and SparseVLM~\cite{zhang2024sparsevlm}, exploit cross-modal attention to prune visual tokens within the decoder. However, the visual evidence they preserve largely depends on the depth at which importance is estimated. To test whether an importance estimate at one layer remains valid across depth, we compare its top-$K$ attended visual token set with those at layers a distance $d$ away using intersection over union (IoU). Since IoU compares set membership rather than attention magnitudes, it directly measures how much the two layers agree on which tokens matter. Our results reveal \textbf{Visual Token Importance Evolution}: as the distance $d$ between two decoder layers increases, the overlap between their selected visual token sets steadily decreases. As shown in \cref{fig:empirical_study}(b), continuity remains well below~$1$, indicating persistent token turnover. A visual token judged unimportant at an early layer may therefore become important later, so a one-shot decision can irreversibly remove visual evidence required by subsequent reasoning. Together, these findings motivate the Adaptive Stage for broad feature-space coverage before fusion and the Progressive Stage for finer-grained refinement inside the decoder via evolving text-to-visual attention. Although some recent works combine pre-LLM and in-LLM pruning~\cite{liu2024mustdrop,zhang2025vscan}, they do not jointly address both issues.

In this paper, we propose \textbf{STAR-Pro} (\textbf{ST}age-Wise \textbf{A}daptive Token \textbf{R}eduction with \textbf{Pro}gressive Refinement), a training-free two-stage framework that follows this coverage-to-refinement principle. Before fusion, the Adaptive Stage applies pivoted QR factorization to construct an over-budget feature-coverage candidate pool. Once cross-modal evidence emerges, the Progressive Stage uses evolving text-to-visual attention at selected decoder layers to progressively prune a nested survivor set under a schedule matched to the target layer-average budget. By matching each pruning decision to the signal available at that stage, STAR-Pro preserves complementary visual evidence in the candidate pool and progressively refines it as decoder grounding evolves. The exact architecture-specific pruning schedules are provided in \appendixref{sec:suppl_method_impl}. Extensive experiments across image and video LVLMs demonstrate strong accuracy--efficiency trade-offs under aggressive token budgets. On LLaVA-Video-7B, STAR-Pro prunes \textbf{90.5\%} of visual tokens while retaining \textbf{92.7\%} of baseline performance and delivers a \textbf{2.24$\times$} inference speedup, all without architectural modifications.

In summary, our contributions are as follows:
\begin{enumerate}

    \item We reveal existing pre-LLM selections collapse visual feature coverage, while within-LLM importance drifts across layers (\textbf{Visual Token Importance Evolution}), motivating an Adaptive Stage for pre-fusion feature coverage and a Progressive Refinement Stage via evolving text-to-visual attention inside the decoder.

    \item We propose \textbf{STAR-Pro}, a training-free two-stage framework in which the Adaptive Stage constructs an over-budget feature-coverage candidate pool and the Progressive Stage refines a nested survivor set via evolving text-to-visual attention.

    \item We evaluate STAR-Pro across multiple model architectures, benchmarks, and pruning ratios, demonstrating strong accuracy--efficiency trade-offs, especially under aggressive pruning. Ablation studies further demonstrate the benefit of each stage.

\end{enumerate}

\section{Related Work}

\subsection{Large Vision-Language Models (LVLMs)}

Building upon the success of large language models (LLMs) \cite{grattafiori2024llama3,achiam2023gpt4,yang2025qwen3}, recent research has extended these powerful architectures to multimodal domains, giving rise to large vision-language models (LVLMs) \cite{liu2023llava,liu2024llavanext,zhang2024videollava-video,li2024llavaonevision,bai2025qwen25vl,zhu2025internvl3}. These models encode visual information into token sequences that can be processed by LLMs alongside textual inputs. However, visual tokenization introduces a significant computational challenge: the number of visual tokens substantially exceeds that of their textual counterparts. Classical models like LLaVA-1.5 \cite{liu2023llava} convert a 336$\times$336 image to 576 tokens, while high-resolution variants such as LLaVA-NeXT \cite{liu2024llavanext} produce 2,880 tokens at 672$\times$672 resolution. Advanced architectures like Qwen3-VL \cite{Qwen3-VL}, and InternVL3 \cite{zhu2025internvl3} employ dynamic resolution strategies that partition images into multiple patches, further amplifying token counts. The challenge becomes more severe in video understanding, where models such as LLaVA-Video~\cite{zhang2024videollava-video} process plenty of frames, resulting in token sequences exceeding 10K or even 100K tokens. Such extensive token sequences impose substantial computational overhead and severely limit inference speed, making efficient token reduction essential for practical deployment of LVLMs.

\subsection{Token Reduction for LVLMs}

To reduce visual token redundancy, some approaches introduce learned token compressors or pruning modules \cite{li2024tokenpacker,li2024inferencequecc,zhangLLaVAMiniEfficientImage2025}, requiring additional training and, in several cases, architectural modifications. Training-free methods fall into three families by pruning location. \emph{Pre-LLM} methods select tokens before the LLM, using encoder attention \cite{shang2024llava-prumerge,yangVisionZipLongerBetter2024,zhangTextVisualAttentionExploiting2025,fitprune}, feature diversity or subspace reconstruction \cite{alvarDivPruneDiversitybasedVisual2025,zhangAttentionSimilarityMaximizing2025,wenStopLookingImportant2025}, or combinations of saliency with diversity or spatial coverage \cite{deng2025scope,zou2025holov}. Across these baselines, ``coverage,'' when used, denotes spatial or set-level coverage over retained tokens, distinct from the feature-subspace coverage that defines our candidate pool. \emph{In-LLM} methods prune inside decoder layers using cross-modal attention \cite{chen2024imagefastv,xing2024pyramiddrop,zhang2024sparsevlm}, withdraw visual tokens at a selected layer \cite{lin2025vtw}, or balance prompt alignment and visual preservation at a selected decoder layer \cite{li2025mob}; recent variants re-derive a text-guided prior at every pruning layer \cite{zhang2025adaptinfer}. \emph{Hybrid or multi-stage} methods combine pruning before and inside the decoder \cite{liu2024mustdrop,zhang2025vscan}. \textbf{STAR-Pro} is a training-free two-stage visual pruning method whose Adaptive Stage preserves feature-space coverage through an over-budget feature-coverage candidate pool before fusion and whose Progressive Stage refines the nested survivor set across decoder layers using evolving text-to-visual attention.

\section{Method}
\label{sec:method}

STAR-Pro comprises two complementary stages. The \textbf{Adaptive Stage} constructs an over-budget feature-coverage candidate pool from vision-side token representations using pivoted QR (\cref{sec:visual_self_attention}). At selected decoder layers, the \textbf{Progressive Stage} refines this pool using evolving text-to-visual attention (\cref{sec:cross_modal_attention}).

\subsection{Preliminaries}

An LVLM typically consists of a vision encoder $f_v$, a projector $g$, and an LLM $f_\phi$ with $L$ decoder layers. Given an image $X_v$ and a text query $X_q$, the encoder extracts visual features $\mathbf{h}_v = f_v(X_v) \in \mathbb{R}^{n \times d_v}$, which the projector maps into visual tokens $\mathbf{H}_v = g(\mathbf{h}_v) \in \mathbb{R}^{n \times d}$. The text query is embedded into $\mathbf{H}_q \in \mathbb{R}^{n_q \times d}$, and the two are concatenated and fed into $f_\phi$ for autoregressive generation. Since $n \gg n_q$, the visual tokens dominate the per-layer cost, and this overhead compounds across all $L$ layers. Because a pruning method may retain different numbers of visual tokens at different depths, we use $T$ to denote the visual token count averaged over all $L$ decoder layers, referred to hereafter as the layer-average budget.

\begin{figure*}[!t]
  \centering
  \includegraphics[width=\linewidth,trim=8pt 39pt 10pt 24pt,clip]{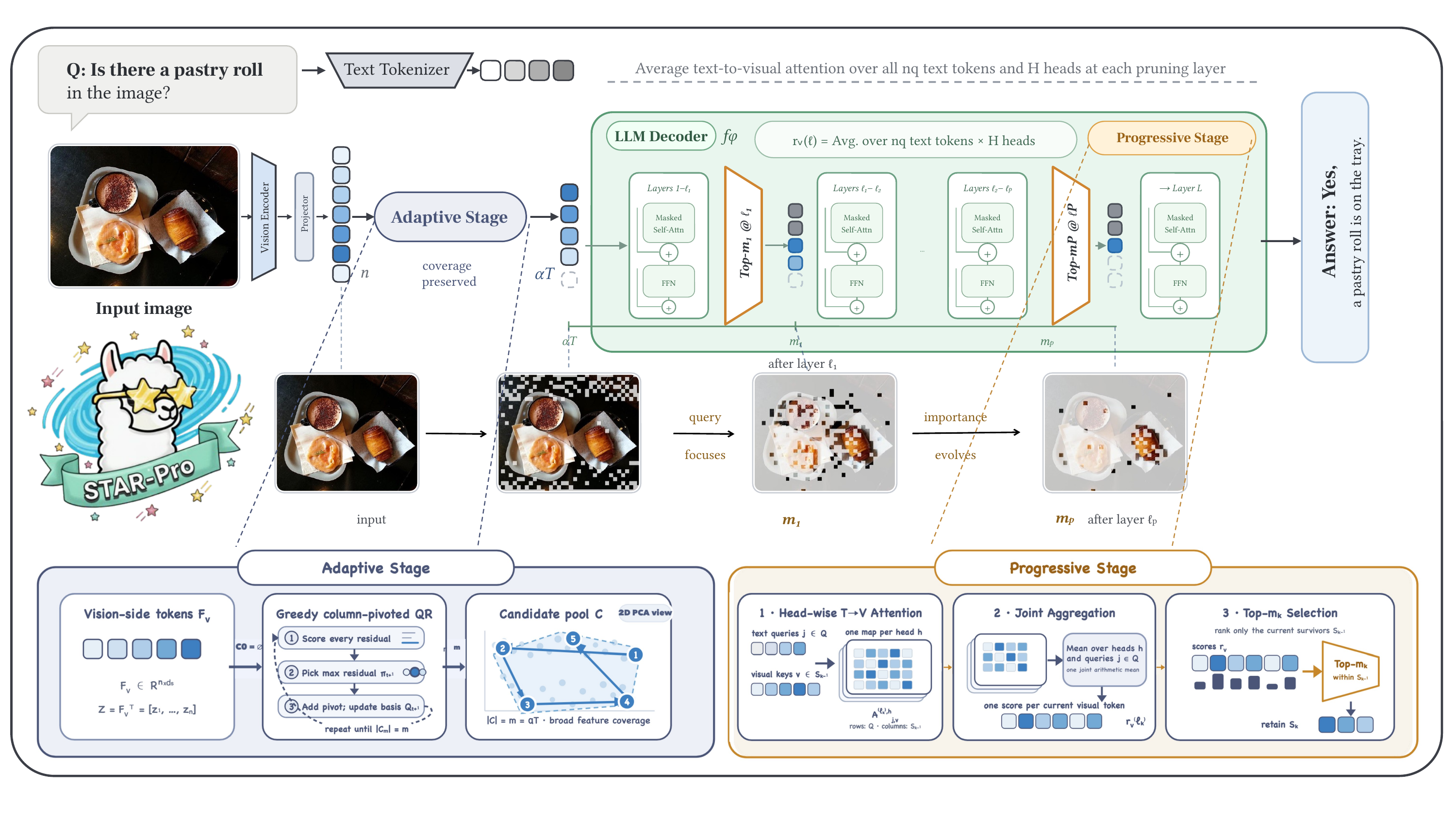}
  \caption{Overview of STAR-Pro. The Adaptive Stage (\S\ref{sec:visual_self_attention}) uses pivoted QR to build a feature-coverage candidate pool. At scheduled pruning layers, the Progressive Stage (\S\ref{sec:cross_modal_attention}) uses evolving text-to-visual attention to progressively refine the pool.}
  \label{fig:overall_structure}
  \vspace{-4pt}
\end{figure*}

\subsection{Adaptive Stage: Feature-Coverage Candidate Pool Construction}
\label{sec:visual_self_attention}

Let $\mathbf{F}_v\in\mathbb{R}^{n\times d_s}$ denote the vision-side token representation on which pivoted QR is applied at the Adaptive Stage pruning point. Given $\mathbf{F}_v$ and target layer-average budget $T$, we set the candidate-pool size to $m=\alpha T$, where $\alpha$ is a hyperparameter. The objective is to preserve broad feature-space coverage in the candidate pool. Let $\mathbf{Z}=\mathbf{F}_v^\top=[\mathbf{z}_1,\ldots,\mathbf{z}_n]$, so that each column $\mathbf{z}_i$ represents one visual token in the selection feature space. For a subset $\mathcal{S}$, the square root of its Gram determinant is the volume spanned by the selected feature vectors. This volume is large when tokens contribute complementary directions and collapses toward zero when they are redundant, providing a natural coverage objective. We formulate candidate selection as
\begin{equation}
\mathcal{S}^{\star}
=
\arg\max_{\substack{\mathcal{S}\subseteq\{1,\ldots,n\}\\|\mathcal{S}|=m}}
\det\!\left(\mathbf{Z}_{\mathcal{S}}^\top
\mathbf{Z}_{\mathcal{S}}\right)^{1/2}.
\label{eq:max_volume}
\end{equation}
Exact maximum-volume subset selection is combinatorial, so we approximate it with greedy column-pivoted QR~\cite{horn2012matrix}. Starting from $\mathcal{C}_0=\varnothing$ and an empty basis $\mathbf{Q}_0$, let $\mathcal{C}_t$ contain the visual token indices selected after $t$ iterations, and let $\mathbf{Q}_t$ be an orthonormal basis for their span in the visual selection feature space. Next \emph{pivot} is chosen as
\begin{equation}
\pi_{t+1}
=
\arg\max_{i\notin\mathcal{C}_t}
\left\|
(\mathbf{I}-\mathbf{Q}_t\mathbf{Q}_t^\top)\mathbf{z}_i
\right\|_2.
\label{eq:qr_pivot}
\end{equation}
The pivot is thus the remaining visual token with the largest residual orthogonal to the selected span, equivalently maximizing the greedy increase in volume. After $m$ iterations, pivoted QR yields a coverage-oriented candidate pool $\mathcal{C}=\mathcal{C}_m$ that retains complementary directions in the visual selection feature space. See \appendixref{sec:suppl_method_impl} for details.

\begin{table*}[t]
\centering
\small
\begin{tabular}{l|ccc|ccc|ccc|ccc}
\toprule
\multirow{2}{*}{\textbf{Method}}
  & \multicolumn{3}{c|}{\textbf{LLaVA-1.5-7B}}
  & \multicolumn{3}{c|}{\textbf{LLaVA-1.5-13B}}
  & \multicolumn{3}{c|}{\textbf{LLaVA-NeXT-7B}}
  & \multicolumn{3}{c}{\textbf{LLaVA-NeXT-13B}} \\
  & \multicolumn{3}{c|}{576 tokens}
  & \multicolumn{3}{c|}{576 tokens}
  & \multicolumn{3}{c|}{2880 tokens}
  & \multicolumn{3}{c}{2880 tokens} \\
\cmidrule(lr){2-4}\cmidrule(lr){5-7}\cmidrule(lr){8-10}\cmidrule(lr){11-13}
Compression
  & 77.8\% & 88.9\% & 94.4\%
  & 77.8\% & 88.9\% & 94.4\%
  & 77.8\% & 88.9\% & 94.4\%
  & 77.8\% & 88.9\% & 94.4\% \\
\midrule
\rowcolor[rgb]{.90, .96, .90} FastV (ECCV24)         & 92.5 & 75.6 & --   & 94.1 & 85.0 & --   & 95.4 & 80.4 & --   & 96.4 & 86.9 & --   \\
\rowcolor[rgb]{.87, .94, .88} PDrop (CVPR25)         & 94.8 & 75.2 & --   & 97.8 & 88.7 & --   & 97.0 & 83.1 & --   & 98.5 & 91.5 & --   \\
\rowcolor[rgb]{.84, .92, .86} SparseVLM (ICML25)     & 96.4 & 87.8 & --   & \best{98.0} & 93.1 & --   & 98.9 & 94.0 & --   & 99.9 & 96.8 & --   \\
\rowcolor[rgb]{.83, .90, .96} VisionZip (CVPR25)     & 96.7 & 93.1 & 87.9 & 96.3 & 93.1 & 87.4 & 99.4 & 95.8 & 90.5 & 94.8 & 96.4 & 92.6 \\
\rowcolor[rgb]{.78, .87, .95} DivPrune (CVPR25)      & 96.7 & 94.3 & 90.9 & 96.7 & 94.8 & 91.6 & 99.0 & 97.2 & 94.2 & 99.1 & 97.1 & 94.9 \\
\rowcolor[rgb]{.75, .85, .95} CDPruner (NeurIPS25)   & \second{97.8} & \second{96.2} & \second{94.1} & 97.2 & 95.7 & \second{94.0} & \second{99.5} & \second{98.1} & \second{96.6} & \second{100.0} & \second{98.2} & \second{96.9} \\
\rowcolor[rgb]{.80, .88, .96} SCOPE (NeurIPS25)      & 97.7 & 96.0 & 93.5 & 97.2 & \second{96.0} & 93.2 & \best{99.6} & \second{98.1} & 95.6 & 99.8 & \second{98.2} & 96.6 \\
\rowcolor[rgb]{.87, .94, .88} VScan (TMLR26)         & \second{97.8} & 96.0 & 91.8 & 97.3 & \second{96.0} & 88.7 & \second{99.5} & 96.2 & 90.3 & 99.2 & 95.9 & 89.9 \\
\rowcolor[rgb]{.90, .96, .90} HoloV (NeurIPS25)      & 96.0 & 92.5 & 88.8 & 95.6 & 92.8 & 88.9 & 97.1 & 96.1 & 92.5 & 97.2 & 96.1 & 93.7 \\
\rowcolor[rgb]{1.0, .92, .80} \textbf{STAR-Pro} & \best{98.9} & \best{97.1} & \best{94.9} & \second{97.9} & \best{97.0} & \best{96.1} & \best{99.6} & \best{99.0} & \best{97.0} & \best{100.3} & \best{98.9} & \best{97.5} \\
\bottomrule
\end{tabular}
\normalsize
\caption{Evaluation on LLaVA Series. All numbers report Rel. (\%), performance relative to the corresponding baseline. Ties at the displayed one-decimal precision are marked jointly best. ``--'': not applicable.}
\label{tab:llava_series_summary}
\vspace{-8pt}
\end{table*}

\subsection{Progressive Stage: Refinement via Evolving Text-to-Visual Attention}
\label{sec:cross_modal_attention}

Our empirical study (\cref{fig:empirical_study}(b)) shows that visual-token importance evolves across decoder layers, yet pruning is irreversible: a one-shot decision may discard visual tokens that later layers rely on for reasoning. We therefore distribute token reduction across multiple depths and re-estimate importance before each pruning step. Let $\mathcal{S}_0=\mathcal{C}$ denote the candidate pool entering the LLM decoder. Given $P$ pruning layers $\mathcal{L}=\{\ell_1,\ldots,\ell_P\}$ and their non-increasing budgets $\{m_1,\ldots,m_P\}$, at each layer $\ell_k$ we compute text-to-visual attention scores over the currently surviving visual tokens $\mathcal{S}_{k-1}$. We compute the evolving score of each $v\in\mathcal{S}_{k-1}$ by averaging over all heads and text tokens:
\begin{equation}
r_v^{(\ell_k)}
=\frac{1}{n_qH}\sum_{j=1}^{n_q}\sum_{h=1}^{H}
\mathbf{A}^{(\ell_k),h}_{j,v}.
\label{eq:stage2_agg}
\end{equation}
Here, $\mathbf{A}^{(\ell_k),h}_{j,v}$ denotes the text-to-visual attention score between text position $j$ and visual token $v$ at head $h$ of layer $\ell_k$. The resulting $r_v^{(\ell_k)}$ provides a layer-specific relevance score for ranking the currently surviving visual tokens.

At each pruning layer $\ell_k$, the Progressive Stage retains the $m_k$ highest-scoring visual tokens, producing nested survivor sets $\mathcal{C}=\mathcal{S}_0\supseteq\mathcal{S}_1\supseteq\cdots\supseteq\mathcal{S}_P$; removed tokens are not reintroduced. The non-increasing budgets are chosen so that the visual-token count averages to $T$ across decoder layers. See \appendixref{sec:suppl_method_impl} for settings.

\section{Experiments}

\subsection{Experimental Setup}

\noindent\textbf{Model architectures.} We evaluate STAR-Pro on 7 LVLMs: LLaVA-1.5-7B/13B~\cite{liu2023llava}, LLaVA-NeXT-7B/13B~\cite{liu2024llavanext}, LLaVA-Video-7B~\cite{zhang2024videollava-video}, Qwen3-VL-8B-Instruct~\cite{Qwen3-VL}, and InternVL3-8B~\cite{zhu2025internvl3}.

\noindent\textbf{Evaluation benchmarks.} We use 15 image benchmarks: VQAv2~\cite{goyal2017makingVQAV2}, GQA~\cite{hudson2019gqa}, VizWiz~\cite{gurariVizWizGrandChallenge2018}, ScienceQA-IMG~\cite{lu2022learnScienceQA}, TextVQA~\cite{singhVQAModelsThat2019}, POPE~\cite{liEvaluatingObjectHallucination2023a}, MME~\cite{fu2023mme}, MMBench-EN/CN~\cite{liuMMBenchYourMultimodal2024}, MM-Vet~\cite{yu2023mmvet}, MMStar~\cite{chen2024we}, AI2D~\cite{kembhaviDiagramWorthDozen2016}, HallusionBench~\cite{guanHallusionBenchAdvancedDiagnostic2024}, MMMU~\cite{yue2024mmmu}, and SEED-Bench~\cite{li2024seedbench}. We also use 3 video benchmarks: MLVU~\cite{zhouMLVUBenchmarkingMultitask2025}, LongVideoBench~\cite{wuLongVideoBenchBenchmarkLongcontext2024}, and Video-MME~\cite{fuVideoMMEFirstEverComprehensive2025}.

\noindent\textbf{Comparison methods.} We compare STAR-Pro with 9 training-free baselines: FastV~\cite{chen2024imagefastv}, PyramidDrop~\cite{xing2024pyramiddrop}, SparseVLM~\cite{zhang2024sparsevlm}, VisionZip~\cite{yangVisionZipLongerBetter2024}, DivPrune~\cite{alvarDivPruneDiversitybasedVisual2025}, CDPruner~\cite{zhangAttentionSimilarityMaximizing2025}, SCOPE~\cite{deng2025scope}, VScan~\cite{zhang2025vscan}, and HoloV~\cite{zou2025holov}. See 
\appendixref{sec:suppl_setup} for the full experimental setup.

\noindent\textbf{Implementation settings.} For both LLaVA-1.5-7B and LLaVA-NeXT-7B, we use $\alpha{=}2$ for the Adaptive Stage and pruning layers $\{12,20\}$ for the Progressive Stage; LLaVA-NeXT budgets count the concatenated tokens from its five image crops. See \appendixref{sec:suppl_method_impl} for details.

\subsection{Main Results}

\Cref{tab:llava_series_summary} compares STAR-Pro with other training-free visual token pruning methods on LLaVA-1.5 and LLaVA-NeXT at two model scales (7B and 13B) under the same three compression ratios. At the table's one-decimal precision, STAR-Pro is best or tied for best in 11 of the 12 model--compression settings, trailing SparseVLM by only 0.1 points solely on LLaVA-1.5-13B at 77.8\% compression. On LLaVA-1.5, STAR-Pro remains strongest at the most aggressive 94.4\% compression, retaining 94.9\% and 96.1\% of baseline performance on the 7B and 13B models, respectively. The same trend extends to LLaVA-NeXT: at 94.4\% compression, STAR-Pro retains 97.0\% and 97.5\% on the two model scales and ranks first on both. Thus, across all four models, STAR-Pro preserves at least 94.9\% of baseline performance under the strongest compression. These results validate the complementary of our design across architectures and model scales. See \appendixref{sec:suppl_results} for details.

\subsection{STAR-Pro for Video Understanding}

\begin{table}[t]
\centering
\small
\setlength{\tabcolsep}{1mm}
\begin{tabularx}{\columnwidth}{@{}l|
>{\hsize=1.05\hsize\centering\arraybackslash}X
>{\hsize=.85\hsize\centering\arraybackslash}X
>{\hsize=1.30\hsize\centering\arraybackslash}X|
>{\hsize=.85\hsize\centering\arraybackslash}X
>{\hsize=.95\hsize\centering\arraybackslash}X@{}}
\toprule
\textbf{Method} & \textbf{MLVU} & \textbf{LVB} & \textbf{VMME} &
\textbf{Acc.} & \textbf{Rel.} \\
\midrule
\multicolumn{6}{c}{\textit{Retain $64 \times 32$ Tokens ($\downarrow$ 81.1\%)}} \\
\midrule
\rowcolor[rgb]{.90, .96, .90}
FastV (ECCV24) & 58.5 & 52.4 & 56.0 & 55.6 & 87.7 \\
\rowcolor[rgb]{.84, .92, .86}
SparseVLM (ICML25) & 60.7 & 53.7 & 59.0 & 57.8 & 91.1 \\
\rowcolor[rgb]{.80, .88, .95}
CDPruner (NeurIPS25) & \second{63.0} & 56.5 & 60.5 & 60.0 & 94.6 \\
\rowcolor[rgb]{.78, .87, .95}
DivPrune (CVPR25) & 61.5 & 56.4 & 59.3 & 59.1 & 93.1 \\
\rowcolor[rgb]{.87, .94, .88}
VScan (TMLR26) & 62.4 & \best{57.8} & \second{60.9} & \second{60.4} & \second{95.2} \\
\rowcolor[rgb]{1.0, .95, .88}
\textbf{STAR-Pro (Ours)} & \best{65.4} & \second{57.0} & \best{61.6} & \best{61.3} & \best{96.7} \\
\midrule
\multicolumn{6}{c}{\textit{Retain $64 \times 16$ Tokens ($\downarrow$ 90.5\%)}} \\
\midrule
\rowcolor[rgb]{.90, .96, .90}
FastV (ECCV24) & 52.8 & 46.6 & 50.0 & 49.8 & 78.5 \\
\rowcolor[rgb]{.84, .92, .86}
SparseVLM (ICML25) & 52.0 & 47.6 & 49.8 & 49.8 & 78.5 \\
\rowcolor[rgb]{.80, .88, .95}
CDPruner (NeurIPS25) & 58.9 & 52.7 & \second{57.3} & 56.3 & 88.8 \\
\rowcolor[rgb]{.78, .87, .95}
DivPrune (CVPR25) & 58.6 & 52.1 & 56.7 & 55.8 & 88.0 \\
\rowcolor[rgb]{.87, .94, .88}
VScan (TMLR26) & \second{59.2} & \second{54.8} & 57.1 & \second{57.0} & \second{89.9} \\
\rowcolor[rgb]{1.0, .95, .88}
\textbf{STAR-Pro (Ours)} & \best{60.3} & \best{56.1} & \best{60.0} & \best{58.8} & \best{92.7} \\
\bottomrule
\end{tabularx}
\normalsize
\caption{Evaluation on Video. MLVU/LVB/VMME report m-avg/val/w/o-subtitle. Acc. is the equal average of the three reported benchmark scores; Rel. (\%) normalizes Acc. by the unpruned baseline's three-benchmark mean (63.4).}
\label{tab:video_comparison}
\vspace{-16pt}
\end{table}

For video understanding, we evaluate STAR-Pro on LLaVA-Video-7B using 64 frames per video. As shown in \cref{tab:video_comparison}, at a compression ratio of 81.1\%, STAR-Pro retains 32 tokens per frame (2,048 total), preserves 96.7\% of baseline performance, and outperforms the second-best method by 1.5 percentage points. It achieves the best aggregate accuracy, ranking first on MLVU and Video-MME and second on LongVideoBench. Under the more aggressive compression ratio of 90.5\%, only 16 tokens per frame (1,024 total) remain; nevertheless, STAR-Pro preserves 92.7\% of baseline performance, with its margin over the second-best method widening to 2.8 percentage points. At this budget, STAR-Pro ranks first on all three reported video benchmarks. These consistent gains demonstrate that combining the Adaptive Stage with the Progressive Stage remains robust at extreme compression. We further analyze the inference speedups in \cref{sec:computational_efficiency}. See \appendixref{sec:suppl_results} for detailed comparisons.

\begin{table*}[t]
\centering
\small
\setlength{\tabcolsep}{1mm}
\begin{tabularx}{\textwidth}{l|
  >{\hsize=.95\hsize\centering\arraybackslash}X
  >{\hsize=.95\hsize\centering\arraybackslash}X
  >{\hsize=.95\hsize\centering\arraybackslash}X
  >{\hsize=1.20\hsize\centering\arraybackslash}X
  >{\hsize=1.00\hsize\centering\arraybackslash}X
  >{\hsize=1.00\hsize\centering\arraybackslash}X
  >{\hsize=1.00\hsize\centering\arraybackslash}X
  >{\hsize=.95\hsize\centering\arraybackslash}X|
  >{\hsize=.95\hsize\centering\arraybackslash}X
  >{\hsize=1.05\hsize\centering\arraybackslash}X}
\toprule
\textbf{Method} & \textbf{AI2D} & \textbf{POPE}
  & \textbf{HallBench} & \textbf{MME}
  & \textbf{MMB}$^{\text{EN}}$ & \textbf{MMB}$^{\text{CN}}$
  & \textbf{MMStar} & \textbf{SQA} & \textbf{Acc.} & \textbf{Rel.} \\
\specialrule{\lightrulewidth}{0pt}{0pt}
\rowcolor[gray]{.94}
\multicolumn{11}{c}{\textit{Qwen3-VL-8B-Instruct (Total: 1,296 Tokens)}} \\
\specialrule{\lightrulewidth}{0pt}{0pt}
\rowcolor[gray]{.97} \textbf{Baseline} & 84.1 & 89.7 & 56.2 & 2406 & 86.3 & 86.1 & 67.1 & 94.6 & 85.6 & 100 \\
\specialrule{\lightrulewidth}{0pt}{0pt}
\rowcolor[gray]{.97}
\multicolumn{11}{c}{\rule[-0.4ex]{0pt}{2.8ex}\textit{Layer-Average Budget $T{=}256$ ($\downarrow$80.2\%)}} \\
\specialrule{\lightrulewidth}{0pt}{0pt}
\rowcolor[rgb]{.90, .96, .90} FastV (ECCV24)
  & 69.3 & 81.7 & 43.0 & 1913 & 77.3 & 76.7 & 51.1 & 82.0 & 72.1 & 84.3 \\
\rowcolor[rgb]{.75, .85, .95} CDPruner (NeurIPS25)
  & 75.1 & 87.6 & 41.9 & 2034 & 79.7 & 78.7 & 50.9 & 83.7 & 74.9 & 87.6 \\
\rowcolor[rgb]{.78, .87, .95} DivPrune (CVPR25)
  & 80.0 & \best{89.4} & 46.9 & 2236 & 83.6 & 82.3 & 58.3 & 88.0 & 80.0 & 93.6 \\
\rowcolor[rgb]{.87, .94, .88} VScan (TMLR26)
  & \second{80.2} & 87.8 & \best{51.8} & \second{2272}
  & \second{84.6} & 82.6 & 61.1 & \second{91.0} & \second{81.6} & \second{95.4} \\
\rowcolor[rgb]{.90, .96, .90} HoloV (NeurIPS25)
  & \best{80.9} & 89.0 & \second{50.6} & 2254
  & 83.3 & \second{83.2} & \second{61.5} & \best{91.6} & \second{81.6} & \second{95.4} \\
\rowcolor[rgb]{1.0, .92, .80} \textbf{STAR-Pro (Ours)}
  & 79.9 & \second{89.2} & 49.0 & \best{2308}
  & \best{84.8} & \best{83.8} & \best{62.1} & 90.0 & \best{81.8} & \best{95.6} \\
\specialrule{\lightrulewidth}{0pt}{0pt}
\rowcolor[gray]{.97}
\multicolumn{11}{c}{\rule[-0.4ex]{0pt}{2.8ex}\textit{Layer-Average Budget $T{=}128$ ($\downarrow$90.1\%)}} \\
\specialrule{\lightrulewidth}{0pt}{0pt}
\rowcolor[rgb]{.90, .96, .90} FastV (ECCV24)
  & 66.3 & 57.2 & 30.3 & 1312 & 51.6 & 50.0 & 38.3 & 76.4 & 54.5 & 63.7 \\
\rowcolor[rgb]{.75, .85, .95} CDPruner (NeurIPS25)
  & 70.5 & 83.9 & 38.3 & 1824 & 72.9 & 71.2 & 44.6 & 79.0 & 68.9 & 80.6 \\
\rowcolor[rgb]{.78, .87, .95} DivPrune (CVPR25)
  & 74.4 & \second{88.3} & 42.6 & 2089 & 80.8 & 79.5 & 52.2 & 83.8 & 75.8 & 88.6 \\
\rowcolor[rgb]{.87, .94, .88} VScan (TMLR26)
  & \second{76.2} & 85.6 & \best{44.1} & \second{2146}
  & 81.2 & \second{80.1} & 55.3 & \second{86.8} & 77.1 & 90.1 \\
\rowcolor[rgb]{.90, .96, .90} HoloV (NeurIPS25)
  & \best{79.5} & 87.2 & 42.1 & 2126
  & \second{81.4} & 79.8 & \second{55.9} & 85.7 & \second{77.2} & \second{90.3} \\
\rowcolor[rgb]{1.0, .92, .80} \textbf{STAR-Pro (Ours)}
  & 74.6 & \best{89.5} & \second{43.2} & \best{2200}
  & \best{82.8} & \best{82.4} & \best{57.6} & \best{87.6} & \best{78.5} & \best{91.7} \\
\specialrule{\lightrulewidth}{0pt}{0pt}
\rowcolor[gray]{.94}
\multicolumn{11}{c}{\textit{InternVL3-8B (Total: 1,280 Tokens)}} \\
\specialrule{\lightrulewidth}{0pt}{0pt}
\rowcolor[gray]{.97} \textbf{Baseline} & 85.1 & 90.7 & 49.4 & 2369 & 85.7 & 85.1 & 68.3 & 97.8 & 85.1 & 100 \\
\specialrule{\lightrulewidth}{0pt}{0pt}
\rowcolor[gray]{.97}
\multicolumn{11}{c}{\rule[-0.4ex]{0pt}{2.8ex}\textit{Layer-Average Budget $T{=}256$ ($\downarrow$80.0\%)}} \\
\specialrule{\lightrulewidth}{0pt}{0pt}
\rowcolor[rgb]{.90, .96, .90} FastV (ECCV24)
  & \second{80.5} & 89.1 & \second{44.0} & \second{2289}
  & \second{83.6} & \second{83.7} & \second{61.2} & 93.3 & \second{81.2} & \second{95.5} \\
\rowcolor[rgb]{.75, .85, .95} CDPruner (NeurIPS25)
  & 78.8 & 89.4 & 41.5 & 2130 & 79.2 & 78.3 & 55.7 & 90.2 & 77.5 & 91.0 \\
\rowcolor[rgb]{.78, .87, .95} DivPrune (CVPR25)
  & 80.3 & \second{89.8} & 43.0 & 2178 & 81.9 & 80.5 & 58.9 & 91.8 & 79.4 & 93.3 \\
\rowcolor[rgb]{.87, .94, .88} VScan (TMLR26)
  & 76.7 & 87.6 & 41.4 & 2230 & 82.4 & 80.5 & 58.5 & \second{93.9} & 79.1 & 92.9 \\
\rowcolor[rgb]{.90, .96, .90} HoloV (NeurIPS25)
  & 77.9 & 87.5 & 41.7 & 2194 & 82.1 & 80.8 & 57.0 & 91.8 & 78.6 & 92.4 \\
\rowcolor[rgb]{1.0, .92, .80} \textbf{STAR-Pro (Ours)}
  & \best{81.3} & \best{90.6} & \best{46.4} & \best{2331}
  & \best{85.1} & \best{85.6} & \best{63.2} & \best{95.8} & \best{83.1} & \best{97.7} \\
\specialrule{\lightrulewidth}{0pt}{0pt}
\rowcolor[gray]{.97}
\multicolumn{11}{c}{\rule[-0.4ex]{0pt}{2.8ex}\textit{Layer-Average Budget $T{=}128$ ($\downarrow$90.0\%)}} \\
\specialrule{\lightrulewidth}{0pt}{0pt}
\rowcolor[rgb]{.90, .96, .90} FastV (ECCV24)
  & 68.4 & 78.7 & \second{38.9} & 1807 & 73.7 & 73.5 & 47.3 & 83.8 & 69.3 & 81.5 \\
\rowcolor[rgb]{.75, .85, .95} CDPruner (NeurIPS25)
  & 73.1 & 87.5 & 37.4 & 1953 & 75.3 & 73.5 & 51.5 & 85.6 & 72.7 & 85.4 \\
\rowcolor[rgb]{.78, .87, .95} DivPrune (CVPR25)
  & \second{74.2} & \second{88.8} & 37.6 & \second{2051}
  & 78.3 & 75.7 & 52.0 & 87.5 & \second{74.6} & \second{87.7} \\
\rowcolor[rgb]{.87, .94, .88} VScan (TMLR26)
  & 71.9 & 84.7 & 37.8 & 2028
  & \second{78.8} & 76.9 & \second{53.9} & \second{89.0} & 74.3 & 87.3 \\
\rowcolor[rgb]{.90, .96, .90} HoloV (NeurIPS25)
  & 72.4 & 84.1 & 37.9 & 1968 & 78.4 & \second{77.4} & 53.2 & 86.7 & 73.6 & 86.5 \\
\rowcolor[rgb]{1.0, .92, .80} \textbf{STAR-Pro (Ours)}
  & \best{75.9} & \best{89.9} & \best{40.2} & \best{2128}
  & \best{83.0} & \best{82.5} & \best{59.2} & \best{90.7} & \best{78.5} & \best{92.2} \\
\bottomrule
\end{tabularx}
\normalsize
\caption{Evaluation on Advanced Architectures. Budgets are layer-average visual-token counts. Acc. denotes the average across eight benchmarks (MME divided by 20); Rel. (\%) is performance relative to the corresponding baseline.}
\label{tab:advanced_vlm_summary}
\vspace{-12pt}
\end{table*}

\subsection{STAR-Pro for Advanced Architectures}

Beyond the LLaVA series, we evaluate STAR-Pro on Qwen3-VL-8B-Instruct and InternVL3-8B, whose vision encoders and visual-token processing pipelines differ from those of LLaVA. \Cref{tab:advanced_vlm_summary} reports eight-benchmark results at layer-average visual-token budgets of 256 and 128. On Qwen3-VL, STAR-Pro achieves the best relative performance at both budgets, retaining 95.6\% and 91.7\% of baseline performance; its margin over the second-best method increases from 0.2 to 1.4 percentage points under stronger compression. On InternVL3, STAR-Pro retains 97.7\% and 92.2\% of baseline performance, exceeding the runners-up by 2.2 and 4.5 percentage points. It also leads across all eight benchmarks at both budgets. These widening margins show that STAR-Pro becomes more advantageous as token budgets tighten across architectures. See \appendixref{sec:suppl_results} for details.

\subsection{Efficiency Studies}
\label{sec:computational_efficiency}
\begin{table}[!t]
\centering
\small
\setlength{\tabcolsep}{3pt}
\begin{tabularx}{\columnwidth}{@{}>{\raggedright\arraybackslash}Xrrrr@{}}
\toprule
\textbf{Timed component}
& \multicolumn{2}{c}{\textbf{Latency (ms)}}
& \multicolumn{2}{c}{\textbf{$\Delta$}} \\
\cmidrule(lr){2-3}\cmidrule(l){4-5}
& \textbf{Full} & \textbf{STAR-Pro} & \textbf{ms} & \textbf{\%} \\
\midrule
Vision encoder + projector & 662.8 & 659.1 &    3.7 &   0.37 \\
Adaptive Stage: pivoted QR        &   0.0 &  40.4 &  $-40.4$ &  $-3.99$ \\
Auxiliary selection ops.         &   0.0 &   0.2 &   $-0.2$ &  $-0.02$ \\
Decoder prefill             & 968.8 &  95.9 &  872.9 &  86.15 \\
Residual runtime            & 188.9 &  11.9 &  177.0 &  17.47 \\
\midrule
\textbf{End-to-end}
                   & 1820.6 & \textbf{807.4} & \textbf{1013.2} & \textbf{100.00} \\
\bottomrule
\end{tabularx}
\normalsize
\caption{Adaptive-Stage overhead on LLaVA-Video-7B at $T{=}16$.
End-to-end speedup: 2.24$\times$.}
\label{tab:video_qr_overhead}
\vspace{-20pt}
\end{table}

\begin{figure*}[!t]
\begin{minipage}{\textwidth}
\centering
\includegraphics[width=\textwidth]{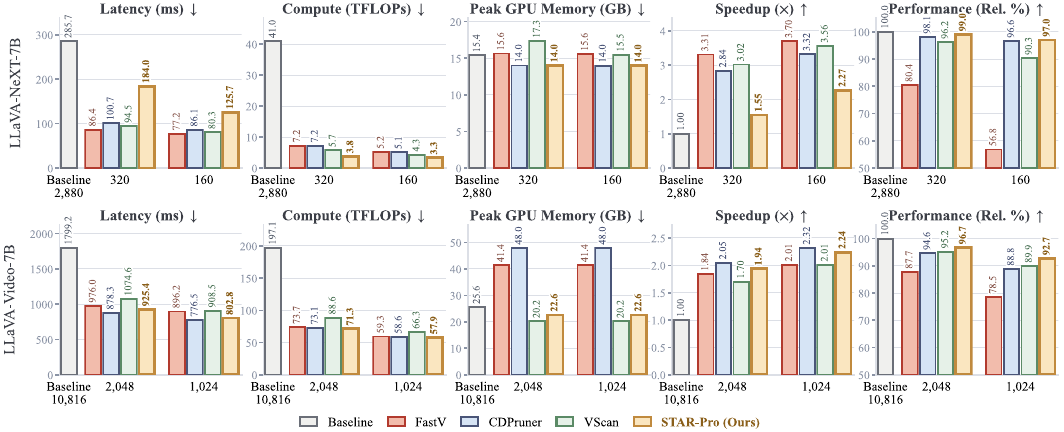}
\captionof{figure}{Efficiency Studies.
Latency, operator-counted TFLOPs, peak allocated GPU memory, speedup, and
relative performance on LLaVA-NeXT-7B (320/160 retained tokens) and
LLaVA-Video-7B (32/16 tokens per frame, 64 frames).}
\label{fig:efficiency_bars}

\includegraphics[width=\textwidth]{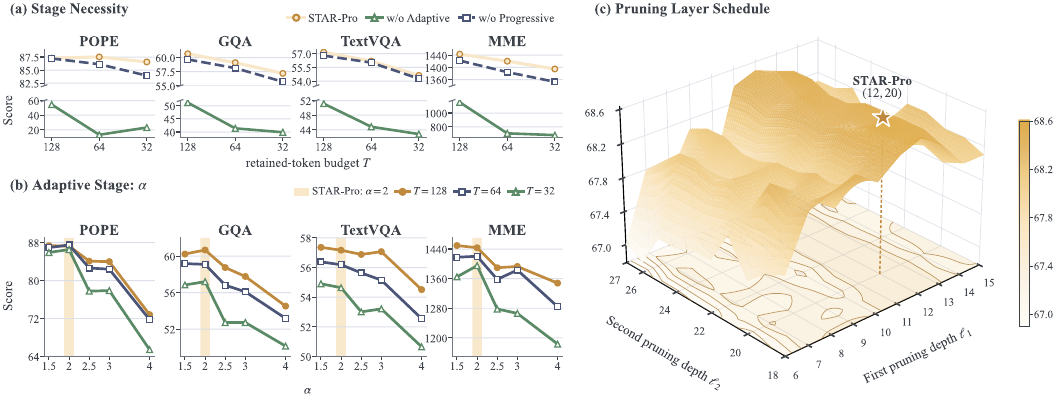}
\captionof{figure}{Ablation Studies.
(a) Stage necessity across token budgets. (b) Sensitivity to $\alpha$.
(c) Matched-compute pruning-layer sweep; height/color denote the Mean-4 score,
and the star marks the best-performing pruning-layer configuration.}
\label{fig:ablation_stage_necessity}
\end{minipage}
\end{figure*}

We measure synchronized end-to-end latency and peak allocated memory on one
NVIDIA A800-80GB GPU using 10 warm-up and 50 measured runs, and report
operator-counted TFLOPs under a common profiling protocol.
\Cref{fig:efficiency_bars} shows that STAR-Pro achieves a
consistently strong performance--efficiency trade-off, with the clearest gains
under extreme compression. At 320 tokens on LLaVA-NeXT and 32 tokens per frame
on LLaVA-Video, STAR-Pro preserves 99.0\% and 96.7\% of baseline performance,
respectively. At 160 tokens on LLaVA-NeXT
(94.4\% visual token reduction), STAR-Pro retains 97.0\% while running
2.27$\times$ faster than the unpruned baseline. Although VScan is
faster (3.56$\times$ vs.\ 2.27$\times$), STAR-Pro dominates in retained
performance (97.0\% vs.\ 90.3\%) and uses less memory (13.99 vs.\
15.46\,GB). LLaVA-Video shows the same
advantage at 16 tokens per frame (90.5\% reduction): STAR-Pro retains 92.7\%
with a 2.24$\times$ speedup over the unpruned baseline. Against its strongest
competitor, VScan, STAR-Pro gains 2.8 percentage points while delivering 11.4\% more
speedup, and uses 12.8--19.6\% fewer operator-counted FLOPs across the two
video budgets. Overall, STAR-Pro advances the accuracy--latency Pareto frontier.

To quantify Adaptive-Stage overhead, \cref{tab:video_qr_overhead} decomposes
synchronized end-to-end latency at $T{=}16$ into vision encoding and
projection, selection (pivoted QR and auxiliary operations), decoder
prefilling, and residual runtime. The 64 frame-wise $169{\rightarrow}32$ QR
calls and auxiliary operations add 40.6\,ms, only 4.0\% of the 1,013.2\,ms net
saving. Reducing the visual prefill span from 10,816 to 2,048 tokens saves
872.9\,ms in decoder prefilling. Decoder prefilling accounts for 86.2\%
of the net saving. Vision-front-end latency remains nearly unchanged
(662.8\,ms vs.\ 659.1\,ms), so the gain comes from shorter decoder prefilling
rather than preprocessing. Overall latency drops from 1,820.6\,ms to
807.4\,ms (2.24$\times$), confirming that the Adaptive-Stage cost is
small relative to downstream savings. See
\appendixref{sec:suppl_efficiency} for timing details.

\subsection{Ablation Studies}
\label{sec:ablation}
All ablations are conducted on LLaVA-1.5-7B and examine stage-wise necessity, the candidate-pool multiplier $\alpha$, and the
pruning-layer schedule. See \appendixref{sec:suppl_ablation} for details.

\noindent\textbf{Stage-wise necessity.}
At matched layer-average compute, \cref{fig:ablation_stage_necessity}(a)
compares STAR-Pro with variants that remove either stage at
$T{\in}\{128,64,32\}$. Removing the Adaptive Stage causes a severe drop across
all benchmarks. Without the Progressive Stage, the Mean-4 gap from STAR-Pro
widens from 0.62 at $T{=}128$ to 1.08 and 1.64 points at $T{=}64$ and $T{=}32$,
showing that broad feature coverage from the Adaptive Stage and refinement via
evolving text-to-visual attention in the Progressive Stage are complementary.

\noindent\textbf{Effect of $\alpha$.}
\Cref{fig:ablation_stage_necessity}(b) sweeps $\alpha$ under
matched layer-average compute. Values 1.5 and 2 are nearly tied at $T{=}128$,
while $\alpha{=}2$ performs best at $T{=}64$ and $T{=}32$. Larger pools
($\alpha{\geq}2.5$) require earlier decoder pruning and reduce performance.
Thus we use $\alpha{=}2$.

\noindent\textbf{Pruning-layer schedule.}
\Cref{fig:ablation_stage_necessity}(c) evaluates 100 matched-compute schedules
at $T{=}64$, formed by pairing the first pruning layer
$\ell_1{\in}\{6,\ldots,15\}$ with the second $\ell_2{\in}\{18,\ldots,27\}$
($10{\times}10$ combinations); together with the four Mean-4 benchmarks, this
amounts to $4{\times}100{=}400$ benchmark evaluations. The deployed $(12,20)$
pair ties for the best one-decimal Mean-4 score (68.6), with high scores across
$\ell_1{\in}\{11,12,13\}$ indicating robustness to nearby pruning depths.

\section{Conclusion}
In this work, we analyze visual token pruning and find two requirements: broad pre-fusion coverage and progressive refinement as token importance evolves. We introduce \textbf{STAR-Pro}, a training-free framework: its Adaptive Stage uses pivoted QR for coverage, while its Progressive Stage uses evolving text-to-visual attention for decoder refinement. It generalizes broadly; on LLaVA-Video-7B, STAR-Pro reduces visual tokens by 90.5\%, retains 92.7\% of baseline performance, and achieves a 2.24$\times$ inference speedup.

\section*{Acknowledgments}
This work was supported by the National Natural Science Foundation of China (62476011), and by the Beijing Natural Science Foundation (L252060).

\bibliography{main}

@String(CVPR= {IEEE Conf. Comput. Vis. Pattern Recog.})

@String(ICCV= {Int. Conf. Comput. Vis.})

@String(AAAI = {AAAI})

@String(CVPR  = {CVPR})

@String(ICCV  = {ICCV})

@article{achiam2023gpt4,
  title={Gpt-4 technical report},
  author={Achiam, Josh and Adler, Steven and Agarwal, Sandhini and Ahmad, Lama and Akkaya, Ilge and Aleman, Florencia Leoni and Almeida, Diogo and Altenschmidt, Janko and Altman, Sam and Anadkat, Shyamal and others},
  journal={arXiv preprint arXiv:2303.08774},
  year={2023}
}

@article{grattafiori2024llama3,
  title={The llama 3 herd of models},
  author={Grattafiori, Aaron and Dubey, Abhimanyu and Jauhri, Abhinav and Pandey, Abhinav and Kadian, Abhishek and Al-Dahle, Ahmad and Letman, Aiesha and Mathur, Akhil and Schelten, Alan and Vaughan, Alex and others},
  journal={arXiv preprint arXiv:2407.21783},
  year={2024}
}

@article{li2024tokenpacker,
  title={TokenPacker: Efficient Visual Projector for Multimodal LLM},
  author={Li, Wentong and Yuan, Yuqian and Liu, Jian and Tang, Dongqi and Wang, Song and Qin, Jie and Zhu, Jianke and Zhang, Lei},
  journal={International Journal of Computer Vision},
  volume={133},
  number={10},
  pages={6794--6812},
  year={2025}
}

@inproceedings{shang2024llava-prumerge,
  title={LLaVA-PruMerge: Adaptive Token Reduction for Efficient Large Multimodal Models},
  author={Shang, Yuzhang and Cai, Mu and Xu, Bingxin and Lee, Yong Jae and Yan, Yan},
  booktitle={Proceedings of the IEEE/CVF International Conference on Computer Vision (ICCV)},
  pages={22857--22867},
  year={2025}
}

@article{li2024inferencequecc,
  title={Inference optimal vlms need only one visual token but larger models},
  author={Li, Kevin Y and Goyal, Sachin and Semedo, Joao D and Kolter, J Zico},
  journal={arXiv preprint arXiv:2411.03312},
  year={2024}
}

@inproceedings{lin2025vtw,
  title={Boosting multimodal large language models with visual tokens withdrawal for rapid inference},
  author={Lin, Zhihang and Lin, Mingbao and Lin, Luxi and Ji, Rongrong},
  booktitle={Proceedings of the AAAI Conference on Artificial Intelligence},
  volume={39},
  pages={5334--5342},
  year={2025}
}

@inproceedings{chen2024imagefastv,
  title={An image is worth 1/2 tokens after layer 2: Plug-and-play inference acceleration for large vision-language models},
  author={Chen, Liang and Zhao, Haozhe and Liu, Tianyu and Bai, Shuai and Lin, Junyang and Zhou, Chang and Chang, Baobao},
  booktitle={European Conference on Computer Vision},
  pages={19--35},
  year={2024},
  organization={Springer}
}

@inproceedings{zhang2024sparsevlm,
  title={SparseVLM: Visual Token Sparsification for Efficient Vision-Language Model Inference},
  author={Zhang, Yuan and Fan, Chun-Kai and Ma, Junpeng and Zheng, Wenzhao and Huang, Tao and Cheng, Kuan and Gudovskiy, Denis and Okuno, Tomoyuki and Nakata, Yohei and Keutzer, Kurt and Zhang, Shanghang},
  booktitle={Proceedings of the 42nd International Conference on Machine Learning},
  pages={74840--74857},
  year={2025}
}

@article{dosovitskiy2020imageViT,
  title={An image is worth 16x16 words: Transformers for image recognition at scale},
  author={Dosovitskiy, Alexey and Beyer, Lucas and Kolesnikov, Alexander and Weissenborn, Dirk and Zhai, Xiaohua and Unterthiner, Thomas and Dehghani, Mostafa and Minderer, Matthias and Heigold, Georg and Gelly, Sylvain and others},
  journal={arXiv preprint arXiv:2010.11929},
  year={2020}
}

@article{team2023gemini,
  title={Gemini: a family of highly capable multimodal models},
  author={Team, Gemini and Anil, Rohan and Borgeaud, Sebastian and Alayrac, Jean-Baptiste and Yu, Jiahui and Soricut, Radu and Schalkwyk, Johan and Dai, Andrew M and Hauth, Anja and Millican, Katie and others},
  journal={arXiv preprint arXiv:2312.11805},
  year={2023}
}

@article{yang2025qwen3,
  title={Qwen3 technical report},
  author={Yang, An and Li, Anfeng and Yang, Baosong and Zhang, Beichen and Hui, Binyuan and Zheng, Bo and Yu, Bowen and Gao, Chang and Huang, Chengen and Lv, Chenxu and others},
  journal={arXiv preprint arXiv:2505.09388},
  year={2025}
}

@article{liu2024deepseek,
  title={Deepseek-v3 technical report},
  author={Liu, Aixin and Feng, Bei and Xue, Bing and Wang, Bingxuan and Wu, Bochao and Lu, Chengda and Zhao, Chenggang and Deng, Chengqi and Zhang, Chenyu and Ruan, Chong and others},
  journal={arXiv preprint arXiv:2412.19437},
  year={2024}
}

@inproceedings{goyal2017makingVQAV2,
  title={Making the v in vqa matter: Elevating the role of image understanding in visual question answering},
  author={Goyal, Yash and Khot, Tejas and Summers-Stay, Douglas and Batra, Dhruv and Parikh, Devi},
  booktitle={Proceedings of the IEEE conference on computer vision and pattern recognition},
  pages={6904--6913},
  year={2017}
}

@article{lu2022learnScienceQA,
  title={Learn to explain: Multimodal reasoning via thought chains for science question answering},
  author={Lu, Pan and Mishra, Swaroop and Xia, Tanglin and Qiu, Liang and Chang, Kai-Wei and Zhu, Song-Chun and Tafjord, Oyvind and Clark, Peter and Kalyan, Ashwin},
  journal={Advances in Neural Information Processing Systems},
  volume={35},
  pages={2507--2521},
  year={2022}
}

@inproceedings{hudson2019gqa,
  title={Gqa: A new dataset for real-world visual reasoning and compositional question answering},
  author={Hudson, Drew A and Manning, Christopher D},
  booktitle={Proceedings of the IEEE/CVF conference on computer vision and pattern recognition},
  pages={6700--6709},
  year={2019}
}

@misc{fu2023mme,
    title={MME: A Comprehensive Evaluation Benchmark for Multimodal Large Language Models},
    author={Chaoyou Fu and Peixian Chen and Yunhang Shen and Yulei Qin and Mengdan Zhang and Xu Lin and Jinrui Yang and Xiawu Zheng and Ke Li and Xing Sun and Yunsheng Wu and Rongrong Ji},
    year={2023},
    eprint={2306.13394},
    archivePrefix={arXiv},
    primaryClass={cs.CV}
}

@article{yu2023mmvet,
  title={Mm-vet: Evaluating large multimodal models for integrated capabilities},
  author={Yu, Weihao and Yang, Zhengyuan and Li, Linjie and Wang, Jianfeng and Lin, Kevin and Liu, Zicheng and Wang, Xinchao and Wang, Lijuan},
  journal={arXiv preprint arXiv:2308.02490},
  year={2023}
}

@inproceedings{xing2024pyramiddrop,
  title={PyramidDrop: Accelerating Your Large Vision-Language Models via Pyramid Visual Redundancy Reduction},
  author={Xing, Long and Huang, Qidong and Dong, Xiaoyi and Lu, Jiajie and Zhang, Pan and Zang, Yuhang and Cao, Yuhang and He, Conghui and Wang, Jiaqi and Wu, Feng and Lin, Dahua},
  booktitle={Proceedings of the IEEE/CVF Conference on Computer Vision and Pattern Recognition (CVPR)},
  year={2025}
}

@article{liu2023llava,
  title={Visual instruction tuning},
  author={Liu, Haotian and Li, Chunyuan and Wu, Qingyang and Lee, Yong Jae},
  journal={Advances in neural information processing systems},
  volume={36},
  pages={34892--34916},
  year={2023}
}

@misc{liu2024llavanext,
    title={LLaVA-NeXT: Improved reasoning, OCR, and world knowledge},
    url={https://llava-vl.github.io/blog/2024-01-30-llava-next/},
    author={Liu, Haotian and Li, Chunyuan and Li, Yuheng and Li, Bo and Zhang, Yuanhan and Shen, Sheng and Lee, Yong Jae},
    month={January},
    year={2024}
}

@article{zhang2024videollava-video,
  title={Video instruction tuning with synthetic data},
  author={Zhang, Yuanhan and Wu, Jinming and Li, Wei and Li, Bo and Ma, Zejun and Liu, Ziwei and Li, Chunyuan},
  journal={arXiv preprint arXiv:2410.02713},
  year={2024}
}

@article{bai2025qwen25vl,
  title={Qwen2.5-VL technical report},
  author={Bai, Shuai and Chen, Keqin and Liu, Xuejing and others},
  journal={arXiv preprint arXiv:2502.13923},
  year={2025}
}

@article{Qwen3-VL,
  title={Qwen3-VL Technical Report},
  author={Shuai Bai and Yuxuan Cai and Ruizhe Chen and Keqin Chen and Xionghui Chen and Zesen Cheng and Lianghao Deng and Wei Ding and Chang Gao and Chunjiang Ge and Wenbin Ge and Zhifang Guo and Qidong Huang and Jie Huang and Fei Huang and Binyuan Hui and Shutong Jiang and Zhaohai Li and Mingsheng Li and Mei Li and Kaixin Li and Zicheng Lin and Junyang Lin and Xuejing Liu and Jiawei Liu and Chenglong Liu and Yang Liu and Dayiheng Liu and Shixuan Liu and Dunjie Lu and Ruilin Luo and Chenxu Lv and Rui Men and Lingchen Meng and Xuancheng Ren and Xingzhang Ren and Sibo Song and Yuchong Sun and Jun Tang and Jianhong Tu and Jianqiang Wan and Peng Wang and Pengfei Wang and Qiuyue Wang and Yuxuan Wang and Tianbao Xie and Yiheng Xu and Haiyang Xu and Jin Xu and Zhibo Yang and Mingkun Yang and Jianxin Yang and An Yang and Bowen Yu and Fei Zhang and Hang Zhang and Xi Zhang and Bo Zheng and Humen Zhong and Jingren Zhou and Fan Zhou and Jing Zhou and Yuanzhi Zhu and Ke Zhu},
  journal={arXiv preprint arXiv:2511.21631},
  year={2025}
}

@article{li2024llavaonevision,
  title={{LLaVA-OneVision}: Easy Visual Task Transfer},
  author={Li, Bo and Zhang, Yuanhan and Guo, Dong and Zhang, Renrui and Li, Feng and Zhang, Hao and Zhang, Kaichen and Zhang, Peiyuan and Li, Yanwei and Liu, Ziwei and Li, Chunyuan},
  journal={Transactions on Machine Learning Research},
  year={2025}
}

@article{zhu2025internvl3,
  title={Internvl3: Exploring advanced training and test-time recipes for open-source multimodal models},
  author={Zhu, Jinguo and Wang, Weiyun and Chen, Zhe and Liu, Zhaoyang and Ye, Shenglong and Gu, Lixin and Tian, Hao and Duan, Yuchen and Su, Weijie and Shao, Jie and others},
  journal={arXiv preprint arXiv:2504.10479},
  year={2025}
}

@misc{guanHallusionBenchAdvancedDiagnostic2024,
  title = {{{HallusionBench}}: {{An Advanced Diagnostic Suite}} for {{Entangled Language Hallucination}} and {{Visual Illusion}} in {{Large Vision-Language Models}}},
  shorttitle = {{{HallusionBench}}},
  author = {Guan, Tianrui and Liu, Fuxiao and Wu, Xiyang and Xian, Ruiqi and Li, Zongxia and Liu, Xiaoyu and Wang, Xijun and Chen, Lichang and Huang, Furong and Yacoob, Yaser and Manocha, Dinesh and Zhou, Tianyi},
  year = 2024,
  month = mar,
  number = {arXiv:2310.14566},
  eprint = {2310.14566},
  primaryclass = {cs},
  publisher = {arXiv},
  doi = {10.48550/arXiv.2310.14566},
  urldate = {2025-10-19},
  archiveprefix = {arXiv},
}

@misc{gurariVizWizGrandChallenge2018,
  title = {{{VizWiz Grand Challenge}}: {{Answering Visual Questions}} from {{Blind People}}},
  shorttitle = {{{VizWiz}}},
  author = {Gurari, Danna and Li, Qing and Stangl, Abigale J. and Guo, Anhong and Lin, Chi and Grauman, Kristen and Luo, Jiebo and Bigham, Jeffrey P.},
  year = 2018,
  month = may,
  number = {arXiv:1802.08218},
  eprint = {1802.08218},
  primaryclass = {cs},
  publisher = {arXiv},
  doi = {10.48550/arXiv.1802.08218},
  urldate = {2025-10-19},
  archiveprefix = {arXiv},
}

@misc{kembhaviDiagramWorthDozen2016,
  title = {A {{Diagram Is Worth A Dozen Images}}},
  shorttitle = {{{AI2D}}},
  author = {Kembhavi, Aniruddha and Salvato, Mike and Kolve, Eric and Seo, Minjoon and Hajishirzi, Hannaneh and Farhadi, Ali},
  year = 2016,
  month = mar,
  number = {arXiv:1603.07396},
  eprint = {1603.07396},
  primaryclass = {cs},
  publisher = {arXiv},
  doi = {10.48550/arXiv.1603.07396},
  urldate = {2025-10-19},
  archiveprefix = {arXiv},
}

@inproceedings{liEvaluatingObjectHallucination2023a,
  title={Evaluating object hallucination in large vision-language models},
  author={Li, Yifan and Du, Yifan and Zhou, Kun and Wang, Jinpeng and Zhao, Wayne Xin and Wen, Ji-Rong},
  booktitle={Proceedings of the 2023 conference on empirical methods in natural language processing},
  pages={292--305},
  year={2023}
}

@misc{liuMMBenchYourMultimodal2024,
  title = {{{MMBench}}: {{Is Your Multi-modal Model}} an {{All-around Player}}?},
  shorttitle = {{{MMBench}} En/Cn},
  author = {Liu, Yuan and Duan, Haodong and Zhang, Yuanhan and Li, Bo and Zhang, Songyang and Zhao, Wangbo and Yuan, Yike and Wang, Jiaqi and He, Conghui and Liu, Ziwei and Chen, Kai and Lin, Dahua},
  year = 2024,
  month = aug,
  number = {arXiv:2307.06281},
  eprint = {2307.06281},
  primaryclass = {cs},
  publisher = {arXiv},
  doi = {10.48550/arXiv.2307.06281},
  urldate = {2025-10-19},
  archiveprefix = {arXiv},
  langid = {american},
}

@misc{singhVQAModelsThat2019,
  title = {Towards {{VQA Models That Can Read}}},
  shorttitle = {{{TextVQA}}},
  author = {Singh, Amanpreet and Natarajan, Vivek and Shah, Meet and Jiang, Yu and Chen, Xinlei and Batra, Dhruv and Parikh, Devi and Rohrbach, Marcus},
  year = 2019,
  month = may,
  number = {arXiv:1904.08920},
  eprint = {1904.08920},
  primaryclass = {cs},
  publisher = {arXiv},
  doi = {10.48550/arXiv.1904.08920},
  urldate = {2025-10-19},
  archiveprefix = {arXiv},
}

@misc{fuVideoMMEFirstEverComprehensive2025,
  title = {Video-{{MME}}: {{The First-Ever Comprehensive Evaluation Benchmark}} of {{Multi-modal LLMs}} in {{Video Analysis}}},
  shorttitle = {Video-{{MME}}},
  author = {Fu, Chaoyou and Dai, Yuhan and Luo, Yongdong and Li, Lei and Ren, Shuhuai and Zhang, Renrui and Wang, Zihan and Zhou, Chenyu and Shen, Yunhang and Zhang, Mengdan and Chen, Peixian and Li, Yanwei and Lin, Shaohui and Zhao, Sirui and Li, Ke and Xu, Tong and Zheng, Xiawu and Chen, Enhong and Shan, Caifeng and He, Ran and Sun, Xing},
  year = 2025,
  month = may,
  number = {arXiv:2405.21075},
  eprint = {2405.21075},
  primaryclass = {cs},
  publisher = {arXiv},
  doi = {10.48550/arXiv.2405.21075},
  urldate = {2025-10-27},
  archiveprefix = {arXiv},
  langid = {english},
}

@misc{wuLongVideoBenchBenchmarkLongcontext2024,
  title = {{{LongVideoBench}}: {{A Benchmark}} for {{Long-context Interleaved Video-Language Understanding}}},
  shorttitle = {{{LongVideoBench}}},
  author = {Wu, Haoning and Li, Dongxu and Chen, Bei and Li, Junnan},
  year = 2024,
  month = jul,
  number = {arXiv:2407.15754},
  eprint = {2407.15754},
  primaryclass = {cs},
  publisher = {arXiv},
  doi = {10.48550/arXiv.2407.15754},
  urldate = {2025-10-27},
  archiveprefix = {arXiv},
  langid = {english},
}

@misc{zhouMLVUBenchmarkingMultitask2025,
  title = {{{MLVU}}: {{Benchmarking Multi-task Long Video Understanding}}},
  shorttitle = {{{MLVU}}},
  author = {Zhou, Junjie and Shu, Yan and Zhao, Bo and Wu, Boya and Liang, Zhengyang and Xiao, Shitao and Qin, Minghao and Yang, Xi and Xiong, Yongping and Zhang, Bo and Huang, Tiejun and Liu, Zheng},
  year = 2025,
  month = jan,
  number = {arXiv:2406.04264},
  eprint = {2406.04264},
  primaryclass = {cs},
  publisher = {arXiv},
  doi = {10.48550/arXiv.2406.04264},
  urldate = {2025-10-27},
  archiveprefix = {arXiv},
  langid = {english},
}

@inproceedings{alvarDivPruneDiversitybasedVisual2025,
  title={DivPrune: Diversity-based Visual Token Pruning for Large Multimodal Models},
  author={Alvar, Saeed Ranjbar and Singh, Gursimran and Akbari, Mohammad and Zhang, Yong},
  booktitle={Proceedings of the IEEE/CVF Conference on Computer Vision and Pattern Recognition (CVPR)},
  pages={9392--9401},
  year={2025}
}

@inproceedings{yangVisionZipLongerBetter2024,
  title={VisionZip: Longer is Better but Not Necessary in Vision Language Models},
  author={Yang, Senqiao and Chen, Yukang and Tian, Zhuotao and Wang, Chengyao and Li, Jingyao and Yu, Bei and Jia, Jiaya},
  booktitle={Proceedings of the IEEE/CVF Conference on Computer Vision and Pattern Recognition (CVPR)},
  pages={19792--19802},
  year={2025}
}

@inproceedings{zhangAttentionSimilarityMaximizing2025,
  title={Beyond Attention or Similarity: Maximizing Conditional Diversity for Token Pruning in MLLMs},
  author={Zhang, Qizhe and Liu, Mengzhen and Li, Lichen and Lu, Ming and Zhang, Yuan and Pan, Junwen and She, Qi and Zhang, Shanghang},
  booktitle={Advances in Neural Information Processing Systems},
  volume={38},
  year={2025}
}

@inproceedings{zhangLLaVAMiniEfficientImage2025,
  title={LLaVA-Mini: Efficient Image and Video Large Multimodal Models with One Vision Token},
  author={Zhang, Shaolei and Fang, Qingkai and Yang, Zhe and Feng, Yang},
  booktitle={The Thirteenth International Conference on Learning Representations},
  year={2025}
}

@inproceedings{zhangTextVisualAttentionExploiting2025,
  title={Beyond Text-Visual Attention: Exploiting Visual Cues for Effective Token Pruning in VLMs},
  author={Zhang, Qizhe and Cheng, Aosong and Lu, Ming and Zhang, Renrui and Zhuo, Zhiyong and Cao, Jiajun and Guo, Shaobo and She, Qi and Zhang, Shanghang},
  booktitle={Proceedings of the IEEE/CVF International Conference on Computer Vision (ICCV)},
  pages={20857--20867},
  year={2025}
}

@inproceedings{wenStopLookingImportant2025,
  title={Stop Looking for ``Important Tokens'' in Multimodal Language Models: Duplication Matters More},
  author={Wen, Zichen and Gao, Yifeng and Wang, Shaobo and Zhang, Junyuan and Zhang, Qintong and Li, Weijia and He, Conghui and Zhang, Linfeng},
  booktitle={Proceedings of the 2025 Conference on Empirical Methods in Natural Language Processing},
  pages={9961--9980},
  year={2025}
}

@article{chen2024we,
  title={Are we on the right way for evaluating large vision-language models?},
  author={Chen, Lin and Li, Jinsong and Dong, Xiaoyi and Zhang, Pan and Zang, Yuhang and Chen, Zehui and Duan, Haodong and Wang, Jiaqi and Qiao, Yu and Lin, Dahua and others},
  journal={Advances in Neural Information Processing Systems},
  volume={37},
  pages={27056--27087},
  year={2024}
}

@article{liu2024mustdrop,
  title={Multi-Stage Vision Token Dropping: Towards Efficient Multimodal Large Language Model},
  author={Liu, Ting and Shi, Liangtao and Hong, Richang and Hu, Yue and Yin, Quanjun and Zhang, Linfeng},
  journal={arXiv preprint arXiv:2411.10803},
  year={2024}
}

@article{zhang2025vscan,
  title={VScan: Rethinking Visual Token Reduction for Efficient Large Vision-Language Models},
  author={Zhang, Ce and Ma, Kaixin and Fang, Tianqing and Yu, Wenhao and Zhang, Hongming and Zhang, Zhisong and Xie, Yaqi and Sycara, Katia and Mi, Haitao and Yu, Dong},
  journal={Transactions on Machine Learning Research},
  year={2026}
}

@inproceedings{deng2025scope,
  title={SCOPE: Saliency-Coverage Oriented Token Pruning for Efficient Multimodel LLMs},
  author={Deng, Jinhong and Li, Wen and Zhou, Joey Tianyi and He, Yang},
  booktitle={Advances in Neural Information Processing Systems (NeurIPS)},
  year={2025}
}

@inproceedings{zou2025holov,
  title={Don't Just Chase ``Highlighted Tokens'' in {MLLM}s: Revisiting Visual Holistic Context Retention},
  author={Zou, Xin and Lu, Di and Wang, Yizhou and Yan, Yibo and Lyu, Yuanhuiyi and Zheng, Xu and Zhang, Linfeng and Hu, Xuming},
  booktitle={Advances in Neural Information Processing Systems (NeurIPS)},
  year={2025},
  eprint={2510.02912},
  archivePrefix={arXiv}
}

@article{zhang2025adaptinfer,
  title={{AdaptInfer}: Adaptive Token Pruning for Vision-Language Model Inference with Dynamical Text Guidance},
  author={Zhang, Weichen and Zhu, Zhui and Li, Ningbo and Tao, Shilong and Liu, Kebin and Liu, Yunhao},
  journal={arXiv preprint arXiv:2508.06084},
  year={2025}
}

@inproceedings{li2025mob,
  title={Why 1 + 1 $<$ 1 in Visual Token Pruning: Beyond Naive Integration via Multi-Objective Balanced Covering},
  author={Li, Yangfu and Zhan, Hongjian and Chen, Tianyi and Liu, Qi and Lu, Yue},
  booktitle={Advances in Neural Information Processing Systems},
  volume={38},
  year={2025}
}

@book{horn2012matrix,
  title={Matrix Analysis},
  author={Horn, Roger A. and Johnson, Charles R.},
  edition={2nd},
  publisher={Cambridge University Press},
  year={2012}
}

@inproceedings{yue2024mmmu,
  author={Yue, Xiang and Ni, Yuansheng and Zhang, Kai and Zheng, Tianyu and Liu, Ruoqi and Zhang, Ge and Stevens, Samuel and Jiang, Dongfu and Ren, Weiming and Sun, Yuxuan and Wei, Cong and Yu, Botao and Yuan, Ruibin and Sun, Renliang and Yin, Ming and Zheng, Boyuan and Yang, Zhenzhu and Liu, Yibo and Huang, Wenhao and Sun, Huan and Su, Yu and Chen, Wenhu},
  title={{MMMU}: A Massive Multi-discipline Multimodal Understanding and Reasoning Benchmark for Expert {AGI}},
  booktitle={Proceedings of the IEEE/CVF Conference on Computer Vision and Pattern Recognition},
  pages={9556--9567},
  year={2024}
}

@inproceedings{li2024seedbench,
  author={Li, Bohao and Ge, Yuying and Ge, Yixiao and Wang, Guangzhi and Wang, Rui and Zhang, Ruimao and Shan, Ying},
  title={{SEED-Bench}: Benchmarking Multimodal Large Language Models},
  booktitle={Proceedings of the IEEE/CVF Conference on Computer Vision and Pattern Recognition},
  pages={13299--13308},
  year={2024}
}

@inproceedings{fitprune,
  title={Fit and prune: Fast and training-free visual token pruning for multi-modal large language models},
  author={Ye, Weihao and Wu, Qiong and Lin, Wenhao and Zhou, Yiyi},
  booktitle={Proceedings of the AAAI Conference on Artificial Intelligence},
  volume={39},
  number={21},
  pages={22128--22136},
  year={2025}
}

\appendix
\section*{Appendix Roadmap}

The appendix is organized into five major categories.
Appendix~\ref{sec:suppl_setup} documents the experimental setup.
Appendix~\ref{sec:suppl_method_impl} gives the model implementation details.
Appendix~\ref{sec:suppl_results} provides additional results and efficiency studies.
Appendix~\ref{sec:suppl_ablation} contains the extended ablation studies.
Appendix~\ref{sec:suppl_empirical} documents the two empirical-study protocols used in the main paper.

\section{Experimental Setup}
\label{sec:suppl_setup}

This appendix provides comprehensive details about model architectures, evaluation benchmarks, and baseline methods used in our experiments.

\subsection{Model Architectures}
\label{sec:suppl_model_arch}

\noindent\textbf{LLaVA-1.5.} \cite{liu2023llava} We evaluate the LLaVA-1.5 architecture, which combines a CLIP ViT-L/14 vision encoder with Vicuna-7B/13B language models through a two-layer MLP projector. This version refines the original LLaVA design by replacing the single linear projection layer with an MLP adapter and expanding the instruction tuning dataset, achieving substantial gains in multimodal understanding capabilities. For our experiments, we process images at $336\times 336$ resolution, yielding $576$ visual tokens ($24\times 24$ spatial grid) that serve as input to the language model. We conduct experiments on both 7B and 13B model scales to analyze performance-efficiency trade-offs.

\noindent\textbf{LLaVA-NeXT.} \cite{liu2024llavanext} This iteration introduces adaptive resolution handling to accommodate higher-quality visual inputs. The model dynamically partitions high-resolution images into multiple tiles while preserving the original aspect ratio, then processes each tile through the vision encoder independently before concatenating the resulting features for the LLM. This tiling strategy enables processing of images at up to $4\times$ higher resolution compared to LLaVA-1.5 without modifying the vision encoder architecture. For controlled evaluation across different pruning methods, we standardize the input to $672\times 672$ resolution, generating $2{,}880$ visual tokens through the tiled encoding approach.

\noindent\textbf{LLaVA-Video.} \cite{zhang2024videollava-video} We extend our evaluation to this video-specialized variant that processes temporal sequences of frames. The architecture employs SigLIP as the vision backbone and incorporates a SlowFast-inspired temporal sampling strategy to manage the token budget across video frames. Each frame is encoded at $384 \times 384$ resolution into a $27\times27$ grid of $729$ tokens. Stride-2 $2\times2$ average pooling then produces a $13\times13$ grid, or $169$ tokens per frame. Our evaluation samples exactly $64$ frames per video clip, yielding $64\times169=10{,}816$ post-pooling visual tokens. Accordingly, retaining 32 or 16 tokens per frame reduces this post-pooling token count by 81.1\% or 90.5\%, respectively.

\noindent\textbf{Qwen3-VL-8B-Instruct.} \cite{Qwen3-VL} We evaluate the dense 8B instruction-tuned Qwen3-VL model, which combines a vision transformer with a Qwen language backbone and uses enhanced interleaved MRoPE and DeepStack multi-level visual features for vision--language alignment. The single-image reference configuration used in our comparison tables contains $1{,}296$ post-merger visual tokens. We report layer-average budgets of 256 and 128 tokens, corresponding to 80.2\% and 90.1\% reductions from this reference length.

\noindent\textbf{InternVL3-8B.} \cite{zhu2025internvl3} We evaluate InternVL3, a highly capable open-source multimodal model. The architecture adopts a ViT-MLP-LLM design, integrating a Vision Transformer with the language model via an MLP connector. A distinctive feature is its native multimodal pre-training that jointly optimizes vision and language understanding from early training stages, incorporating Variable Visual Position Encoding for extended contexts. The model employs sophisticated training including supervised fine-tuning and mixed preference optimization. For our experiments, we configure the input resolution to $448\times 448$, producing $1{,}280$ visual tokens.

\subsection{Evaluation Benchmarks}
\label{sec:suppl_benchmarks}

\noindent\textbf{VQAv2.} \cite{goyal2017makingVQAV2} A large-scale visual question answering benchmark built upon COCO images, containing over 1.1 million QA pairs. Each question is paired with multiple answers to reduce language bias, and performance is measured by soft accuracy against human annotations.

\noindent\textbf{GQA.} \cite{hudson2019gqa} A visual reasoning benchmark grounded in scene graphs, featuring compositional questions about spatial relationships and object attributes. We report accuracy on the balanced test split with 12,578 questions, which mitigates answer distribution bias.

\noindent\textbf{VizWiz.} \cite{gurariVizWizGrandChallenge2018} A real-world VQA dataset collected from blind users who photographed everyday scenes and asked spoken questions. Images are often low-quality, making this benchmark a challenging testbed for robust visual understanding.

\noindent\textbf{ScienceQA.} \cite{lu2022learnScienceQA} A multimodal multiple-choice benchmark covering diverse scientific subjects across elementary and high school curricula. It requires joint reasoning over images, diagrams, and textual context.

\noindent\textbf{TextVQA.} \cite{singhVQAModelsThat2019} Evaluates reading and reasoning about text in images. The validation set has 5,000 questions requiring OCR and visual reasoning over scene text.

\noindent\textbf{POPE.} \cite{liEvaluatingObjectHallucination2023a} Polling-based Object Probing Evaluation assesses object hallucination in LVLMs using binary yes/no questions about object presence. We use the adversarial split with 3,000 questions from COCO images.

\noindent\textbf{MME.} \cite{fu2023mme} A comprehensive benchmark evaluating both perception (e.g., OCR, recognition) and cognition (e.g., reasoning, calculation) abilities. The benchmark contains 14 subtasks with a total of 2,374 question-answer pairs.

\noindent\textbf{MMBench.} \cite{liuMMBenchYourMultimodal2024} A systematically-designed benchmark covering 20 ability dimensions including object localization, social reasoning, and OCR. We report results on both English (MMB-EN) and Chinese (MMB-CN) test sets.

\noindent\textbf{MM-Vet.} \cite{yu2023mmvet} Focuses on integrated multimodal capabilities across 6 core abilities: recognition, knowledge, OCR, spatial awareness, language generation, and mathematics. The benchmark contains 218 carefully curated examples.

\noindent\textbf{MMStar.} \cite{chen2024we} A comprehensive vision-language benchmark evaluating multi-modal models across diverse capabilities including coarse and fine-grained perception, logical reasoning, and science understanding.

\noindent\textbf{AI2D.} \cite{kembhaviDiagramWorthDozen2016} A diagram QA benchmark with over 5,000 science diagrams and 15,000 multiple-choice questions, testing diagram understanding in scientific contexts.

\noindent\textbf{HallusionBench.} \cite{guanHallusionBenchAdvancedDiagnostic2024} An image-context reasoning benchmark evaluating vision-language models' susceptibility to language hallucination and visual illusion, with carefully designed examples challenging model robustness.

\noindent\textbf{MMMU.} \cite{yue2024mmmu} A college-level multimodal reasoning benchmark containing 11.5K questions across six broad disciplines and 30 subjects. Its heterogeneous visual inputs and domain-specific questions test expert knowledge, perception, and deliberate reasoning.

\noindent\textbf{SEED-Bench.} \cite{li2024seedbench} A comprehensive multiple-choice benchmark with 24K human-annotated questions spanning 27 dimensions of multimodal comprehension and generation. Its fixed answer options support objective evaluation without an external model-based judge.

\noindent\textbf{MLVU.} \cite{zhouMLVUBenchmarkingMultitask2025} A multi-task long video understanding benchmark covering diverse video genres (movies, surveillance, egocentric, etc.) across a wide range of durations. We report the mean average score (m-avg) across all subtasks.

\noindent\textbf{LongVideoBench.} \cite{wuLongVideoBenchBenchmarkLongcontext2024} A long-context video-language benchmark featuring interleaved video-language inputs of up to one hour. We report results on the validation set overall score as well as two fine-grained subtasks: perception and relation.

\noindent\textbf{Video-MME.} \cite{fuVideoMMEFirstEverComprehensive2025} A comprehensive video multimodal evaluation benchmark spanning 6 visual domains and videos of varying durations (short, medium, long). We report results in the no-subtitle setting, evaluated separately across short, medium, and long video splits.

\subsection{Baseline Methods}
\label{sec:suppl_baselines}

We compare STAR-Pro against the following visual token pruning methods.

\noindent\textbf{FastV.} \cite{chen2024imagefastv} Prunes tokens with lowest vision-text attention scores after layer 2 of the LLM. It is a simple but effective approach that identifies the inefficient visual attention phenomenon in shallow decoder layers.

\noindent\textbf{PyramidDrop.} \cite{xing2024pyramiddrop} Partitions the LLM into multiple stages and drops a fraction of visual tokens at the end of each stage with a pre-defined ratio, creating a pyramid-like visual token reduction schedule across decoder layers.

\noindent\textbf{SparseVLM.} \cite{zhang2024sparsevlm} Selects high-quality text rater tokens based on their relevance to visual content, then uses these raters' cross-modal attention patterns to guide visual token pruning within the decoder.

\noindent\textbf{VisionZip.} \cite{yangVisionZipLongerBetter2024} Identifies dominant tokens via visual attention and selects contextual tokens through clustering in the vision encoder space, combining both for the final token set.

\noindent\textbf{DivPrune.} \cite{alvarDivPruneDiversitybasedVisual2025} Reformulates token pruning as maximizing minimum pairwise distance (MMDP) to retain the most diverse subset in the pre-decoder visual feature space.

\noindent\textbf{CDPruner.} \cite{zhangAttentionSimilarityMaximizing2025} Goes beyond attention and similarity by maximizing the conditional diversity of retained tokens via determinantal point process (DPP). It defines conditional similarity between visual tokens conditioned on the instruction, producing a subset that is both visually diverse and instruction-relevant.

\noindent\textbf{SCOPE.} \cite{deng2025scope} Combines CLS-based saliency with a submodular coverage term that penalizes tokens that are redundant with their neighbors. The final score jointly considers how salient a token is and how semantically unique it is relative to other retained tokens.

\noindent\textbf{VScan.} \cite{zhang2025vscan} Rethinks visual token reduction by scanning across spatial positions to identify informative tokens while suppressing redundancy. It offers a complementary perspective to attention- and similarity-based methods by modeling visual token importance through structured spatial traversal.

\noindent\textbf{HoloV.} \cite{zou2025holov} Allocates the pruning budget adaptively across spatial crops so that the retained tokens preserve holistic visual context rather than concentrating only on highly attended regions. It is a plug-and-play pre-LLM method designed to remain robust under aggressive pruning.

\subsection{Evaluation Protocol}
\label{sec:suppl_eval_protocol}

Our accuracy evaluations use batch size 1 and deterministic greedy decoding without sampling. Within each evaluation harness, STAR-Pro budgets share the same model checkpoint, input examples, image assets, task prompt, and benchmark scorer. Baseline results are taken from the cited methods or reproduced with the corresponding task harness when a compatible rerun is required. LLaVA-1.5 and LLaVA-NeXT follow the official LLaVA evaluation scripts; LLaVA-Video uses \texttt{lmms-eval}; and Qwen3-VL and InternVL3 use VLMEvalKit. We retain each harness's task-specific metric rather than replacing it with a common surrogate: MME is reported as an unnormalised score, MMBench-EN/CN use their language-specific evaluation splits, and Video-MME uses the no-subtitle setting. Aggregate accuracy and relative-performance formulas are stated in the corresponding table captions.

For STAR-Pro and other multi-layer methods, the reported budget is the visual-token count averaged over decoder layers. For one-shot pre-LLM methods, the reported count is the number of tokens entering the decoder. All hardware-dependent latency and speedup comparisons are measured separately under the controlled protocol in \cref{sec:suppl_efficiency}; wall-clock timing from distributed accuracy-evaluation workers is not used.

\section{Model Implementation Details}
\label{sec:suppl_method_impl}

\subsection{Algorithm Details}
\label{sec:suppl_algo}

\Cref{alg:adaptive_stage,alg:progressive_stage} give the complete inference-time
procedure for the two stages.  The Adaptive Stage uses the same residual-QR
operator for every architecture.  For a single visual sequence there is one
selection group; crop- and frame-based models invoke the operator on their
natural groups, and InternVL3 uses the thumbnail/local-crop quotas described in
\cref{sec:suppl_internvl_allocation}.  This wrapper changes only the allocation of
the candidate pool, not the pivot rule.

\subsubsection{Adaptive Stage: Feature-Coverage Candidate Pool Construction}

Let $\mathcal{I}_g$ index selection group $g$, let
$\mathbf{Z}_g=\mathbf{F}_v[\mathcal{I}_g]^\top$, and let $m_g$ be its assigned
candidate quota.  For pivots $\pi_1,\ldots,\pi_{m_g}$, the Gram volume of a
selected group factorizes through the diagonal magnitudes of its QR
factorization:
\begin{equation}
\begin{aligned}
\mathcal{V}(\mathcal{C}_g)
&=\det\!\left(\mathbf{Z}_{g,\mathcal{C}_g}^{\top}
               \mathbf{Z}_{g,\mathcal{C}_g}\right)^{1/2} \\
&=\prod_{t=1}^{m_g}
  \left\|\mathbf{r}_{\pi_t}^{(g,t-1)}\right\|_2 .
\end{aligned}
\label{eq:suppl_qr_volume}
\end{equation}
Thus selecting the largest residual norm at each iteration greedily maximizes
the next multiplicative increase in covered volume.  The selected indices are
finally restored to their original token order; QR determines membership, not
the decoder's spatial sequence order.

\begin{algorithm}[!htbp]
\caption{Adaptive Stage: Feature-Coverage Candidate Pool Construction}
\label{alg:adaptive_stage}
\small
\begin{algorithmic}[1]
\REQUIRE Features $\mathbf{F}_v$; groups $\{\mathcal{I}_g\}_{g=1}^{G}$
\REQUIRE Quotas $\{m_g\}_{g=1}^{G}$
\ENSURE Candidate pool $\mathcal{C}$ with
$|\mathcal{C}|=\sum_{g=1}^{G}m_g$
\STATE $\mathcal{C} \leftarrow \varnothing$
\FOR{$g=1$ to $G$}
    \STATE $\mathbf{R}^{(g,0)}\leftarrow
    \mathbf{F}_v[\mathcal{I}_g]$;
    $\mathcal{C}_g\leftarrow\varnothing$
    \FOR{$t=1$ to $m_g$}
        \STATE $\delta_i\leftarrow\|\mathbf{r}_{i}^{(g,t-1)}\|_2$,
        $i\in\mathcal{I}_g\setminus\mathcal{C}_g$
        \STATE $\pi_t\leftarrow
        \arg\max_{i\in\mathcal{I}_g\setminus\mathcal{C}_g}\delta_i$
        \STATE $\mathbf{q}_t\leftarrow
        \mathbf{r}_{\pi_t}^{(g,t-1)} /
        \|\mathbf{r}_{\pi_t}^{(g,t-1)}\|_2$
        \STATE $\mathcal{C}_g\leftarrow\mathcal{C}_g\cup\{\pi_t\}$
        \FOR{$i\in\mathcal{I}_g\setminus\mathcal{C}_g$}
            \STATE $a_i\leftarrow
            \langle\mathbf{r}_{i}^{(g,t-1)},\mathbf{q}_t\rangle$
            \STATE $\mathbf{r}_{i}^{(g,t)}\leftarrow
            \mathbf{r}_{i}^{(g,t-1)}-a_i\mathbf{q}_t$
        \ENDFOR
    \ENDFOR
    \STATE $\mathcal{C}\leftarrow\mathcal{C}\cup\mathcal{C}_g$
\ENDFOR
\RETURN $\operatorname{sort}_{\mathrm{original}}(\mathcal{C})$
\end{algorithmic}
\end{algorithm}

\subsubsection{Progressive Stage: Refinement via Evolving Text-to-Visual Attention}

Let $m_0=|\mathcal{C}|$, set $\ell_{P+1}=L$, and let pruning after layer
$\ell_k$ produce $m_k$ survivors.  A valid schedule must satisfy both the
nested-budget condition and the exact layer-average constraint
\begin{equation}
\begin{gathered}
m_0\ge m_1\ge\cdots\ge m_P, \\
\frac{\ell_1m_0+
\sum_{k=1}^{P}(\ell_{k+1}-\ell_k)m_k}{L}=T.
\end{gathered}
\label{eq:suppl_progressive_budget}
\end{equation}
Let $\mathcal{Q}$ denote the current text-token positions and
$n_q=|\mathcal{Q}|$. At pruning layer $\ell_k$, STAR-Pro recomputes scores only
over the current survivors and applies
\begin{equation}
\begin{aligned}
r_v^{(\ell_k)}
&=\frac{1}{n_qH}\sum_{j=1}^{n_q}\sum_{h=1}^{H}
  \mathbf{A}_{j,v}^{(\ell_k),h}, \\
\mathcal{S}_k
&=\operatorname{TopK}_{m_k}
  \bigl(\{r_v^{(\ell_k)}:v\in\mathcal{S}_{k-1}\}\bigr)
  \subseteq\mathcal{S}_{k-1}.
\end{aligned}
\label{eq:suppl_progressive_update}
\end{equation}
This explicitly couples layer-specific re-estimation with irreversible,
monotone refinement: removed tokens never re-enter a later candidate set.

\begin{algorithm}[!htbp]
\caption{Progressive Stage: Refinement via Evolving Text-to-Visual Attention}
\label{alg:progressive_stage}
\small
\begin{algorithmic}[1]
\REQUIRE Pool $\mathcal{C}$; depth $L$; text-token set $\mathcal{Q}$
\REQUIRE Pruning layers $0<\ell_1<\cdots<\ell_P<L$
\REQUIRE Budgets $m_1\ge\cdots\ge m_P$
\ENSURE Nested sets $\mathcal{S}_0,\ldots,\mathcal{S}_P$; refined sequence
\STATE $\mathcal{S}_0\leftarrow\mathcal{C}$; $m_0\leftarrow|\mathcal{C}|$;
$\ell_{P+1}\leftarrow L$
\STATE \textbf{assert} Eq.~\eqref{eq:suppl_progressive_budget}
\FOR{$k=1$ to $P$}
    \STATE Continue the decoder through layer $\ell_k$ using $\mathcal{S}_{k-1}$
    \STATE $\bar{\mathbf{A}}^{(\ell_k)}\leftarrow H^{-1}
    \sum_{h=1}^{H}\mathbf{A}^{(\ell_k),h}_{\mathcal{Q},\mathcal{S}_{k-1}}$
    \FOR{$v\in\mathcal{S}_{k-1}$}
        \STATE $r_v^{(\ell_k)}\leftarrow
        |\mathcal{Q}|^{-1}\sum_{j\in\mathcal{Q}}
        \bar{\mathbf{A}}_{j,v}^{(\ell_k)}$
    \ENDFOR
    \STATE $\mathcal{S}_k\leftarrow
    \operatorname{TopK}_{m_k}(\mathbf{r}^{(\ell_k)};\mathcal{S}_{k-1})$
    \STATE Keep hidden states indexed by $\mathcal{S}_k$ and update the
    associated sequence metadata
\ENDFOR
\RETURN $\{\mathcal{S}_k\}_{k=0}^{P}$ and the refined sequence
\end{algorithmic}
\end{algorithm}

\subsection{Implementation Details}
\label{sec:suppl_impl}

\noindent\textbf{Codebase.} We implement STAR-Pro based on the official LLaVA codebase for image-based LLaVA models. For LLaVA-Video, we adopt the official LLaVA-NeXT codebase and utilize lmms-eval for video benchmark evaluation. For Qwen3-VL and InternVL3, we use VLMEvalKit for unified evaluation. Depending on the model interface, the Progressive Stage text-to-visual attention scores defined in the main paper are either read from model-native attention weights or computed directly from the current layer's text-query and visual-key projections. The latter avoids materializing a full quadratic attention matrix for long video sequences.

\noindent\textbf{Selection representation.} In the Adaptive Stage, $\mathbf{F}_v$ is the final vision-side embedding presented to the language decoder. LLaVA-1.5 and Qwen3-VL apply residual pivoted QR over the image's visual-token sequence. LLaVA-NeXT and LLaVA-Video preserve the natural crop or frame grouping and apply the same selector within each group. InternVL3 uses the thumbnail/local-crop allocation documented below. In every case, the selected indices are carried into the decoder without token merging.

\noindent\textbf{Hyperparameters.} At a fixed target layer-average budget $T$, the Adaptive Stage uses the candidate multiplier $\alpha$ (default $\alpha{=}2$), while its QR selector introduces no additional tunable parameter. The Progressive Stage is specified by its pruning layers and integer survivor counts, with every schedule checked against the layer-average constraint before evaluation. This section reports the configurations used for the main results in \cref{tab:architecture_layer_selection}. We analyze the two stages separately in \cref{sec:suppl_hyperparameter_analysis}: the Adaptive-Stage analysis varies the candidate-pool multiplier, while the Progressive-Stage analysis varies the number and placement of pruning layers and the corresponding survivor schedules under matched layer-average budgets.

\subsubsection{Architecture-Specific Pruning Layers}
\label{sec:suppl_layer_selection}

The Progressive Stage uses two decoder pruning layers, but their depths and
integer survivor counts are calibrated to each decoder architecture. We
consolidate the configurations in \cref{tab:architecture_layer_selection};
the listed counts satisfy the layer-average budget constraint introduced in
\cref{sec:suppl_schedule}, except for the unavoidable per-frame integer
rounding reported for LLaVA-Video. The Qwen3-VL row records the adopted
layer-\{14,24\} main-result configuration, while the InternVL3 rows reflect its
architecture-aware local-crop and thumbnail allocation.

The pruning-layer indices are not obtained by transferring a fixed fraction of
the unpruned decoder depth.  Once earlier pruning changes the sequence seen by
later blocks, it also changes the effective depth at which cross-modal evidence
stabilizes.  We therefore treat pruning-layer locations and survivor counts as a
coupled, architecture-specific schedule and recalibrate them jointly under the
layer-average constraint, rather than mapping a schedule by a raw depth ratio.

\begin{table*}[t]
\centering
\small
\begin{tabularx}{\textwidth}{lcccX}
\toprule
\textbf{Model} & \textbf{Decoder layers} & \textbf{Nominal budget $T$}
& \textbf{Adaptive pool $2T$} & \textbf{Progressive checkpoints} \\
\midrule
LLaVA-1.5-7B
& 32 & 128 / 64 / 32 & 256 / 128 / 64
& $\ell_{12}{\rightarrow}74/37/17,\ \ell_{20}{\rightarrow}36/18/10$ \\
LLaVA-1.5-13B
& 40 & 128 / 64 / 32 & 256 / 128 / 64
& $\ell_{15}{\rightarrow}64/32/16,\ \ell_{30}{\rightarrow}32/16/8$ \\
LLaVA-NeXT-7B
& 32 & 640 / 320 / 160 & 1,280 / 640 / 320
& $\ell_{12}{\rightarrow}367/182/91,\ \ell_{20}{\rightarrow}182/92/46$ \\
LLaVA-NeXT-13B
& 40 & 640 / 320 / 160 & 1,280 / 640 / 320
& $\ell_{15}{\rightarrow}320/160/80,\ \ell_{30}{\rightarrow}160/80/40$ \\
LLaVA-Video-7B
& 28 & 32 / 16 & 64 / 32
& $\ell_{12}{\rightarrow}11/5,\ \ell_{21}{\rightarrow}4/3$ \\
Qwen3-VL-8B-Instruct
& 36 & 256 / 128 & 512 / 256
& $\ell_{14}{\rightarrow}128/64,\ \ell_{24}{\rightarrow}64/32$ \\
InternVL3-8B
& 28 & 256 & 512
& $\ell_{12}{\rightarrow}139,\ \ell_{16}{\rightarrow}39$ \\
InternVL3-8B
& 28 & 128 & 256
& $\ell_{11}{\rightarrow}53,\ \ell_{21}{\rightarrow}34$ \\
\bottomrule
\end{tabularx}
\normalsize
\caption{Architecture-specific pruning checkpoints used by STAR-Pro.
Decoder layers are named one-based: an entry $\ell{\rightarrow}m$ prunes
after layer $\ell$, so the following layer receives $m$ visual tokens. In the
zero-based implementation this is the boundary before decoder index $\ell$.
Each schedule starts from the $2T$ candidate pool produced by the Adaptive
Stage. Slash-separated survivor counts correspond in
order to the slash-separated budgets in the same row. For LLaVA-Video, token counts are reported per
frame; integer allocation gives observed layer averages of 31.96 and 16.07 for
the nominal 32- and 16-token budgets.}
\label{tab:architecture_layer_selection}
\end{table*}

\subsubsection{InternVL3-Specific Group Allocation}
\label{sec:suppl_internvl_allocation}

InternVL3 appends a global thumbnail after its local image crops.  We therefore
allocate the Adaptive-Stage candidate pool with weight 4 for the thumbnail and
weight 1 for each local crop.  For the standard four-local-crop input, the
resulting quotas are $[64,64,64,64,256]$ at $T{=}256$ and
$[32,32,32,32,128]$ at $T{=}128$, preserving exactly $2T$ candidates in
total.  Inputs with a different number of crops or multiple images use the
same thumbnail-to-local weighting.

Within each group, STAR-Pro applies the same deterministic residual-QR
selector as in \cref{alg:adaptive_stage}.  The larger thumbnail quota protects
global scene structure, while the per-crop quotas prevent overlapping local
crops from monopolizing the pool.  Only this candidate allocation is
InternVL3-specific; its Progressive Stage follows
\cref{alg:progressive_stage} with the pruning-layer schedule already reported in
\cref{tab:architecture_layer_selection}.

\subsubsection{Token Budget Schedule Derivation}
\label{sec:suppl_schedule}

STAR-Pro targets an average visual token count per LLM decoder layer equal to the user-specified budget $T$.
The two-stage pipeline achieves this as follows:
\noindent\textbf{Adaptive Stage:} Reduces initial visual tokens $n_0$ to $\alpha T$ via pivoted-QR coverage selection.
\textbf{Progressive Stage:} Further reduces tokens across LLM decoder layers at designated pruning layers, such that the \textit{average} visual token count across all $L$ layers equals exactly $T$.

\noindent\textbf{Two-layer pruning design.}
The Progressive Stage can operate with any number of pruning layers.
Our matched-compute ablation (\cref{tab:ablation_isocompute}) compares zero, two, three, and four pruning layers and shows that the deployed two-layer schedule remains competitive with the nearby alternatives while using the simplest progressive design.
We therefore adopt \textbf{two pruning layers} $\ell_1$ and $\ell_2$ ($\ell_1 < \ell_2 < L$) as the standard design for all architectures, and use this as the running example throughout this section.
The schedules name pruning layers one-based (pruning after layer $\ell_k$);
the half-open intervals below use the equivalent zero-based decoder indices.
The resulting token distribution has three segments:
\noindent Layers $[0, \ell_1)$: $\alpha T$ tokens (\textit{inherited from the Adaptive Stage});
Layers $[\ell_1, \ell_2)$: $a$ tokens (pruned at layer $\ell_1$);
Layers $[\ell_2, L)$: $b$ tokens (pruned at layer $\ell_2$).

\noindent\textbf{General constraint.}
Setting the layer-average equal to $T$:
\begin{equation}
\begin{aligned}
\frac{\alpha T \ell_1 + a(\ell_2 - \ell_1) + b(L - \ell_2)}{L}
  &= T, \\
a(\ell_2 - \ell_1) + b(L - \ell_2)
  &= T(L - \alpha\ell_1).
\end{aligned}
\label{eq:schedule_constraint}
\end{equation}
Given the architecture ($L$), candidate multiplier $\alpha$, and pruning-layer positions $(\ell_1,\ell_2)$, this linear equation defines the feasible $(a,b)$ pairs. In practice we select integer-valued $(a,b)$ satisfying Eq.~\ref{eq:schedule_constraint} with a front-heavy constraint $a>b$ to preserve more tokens in earlier layers.

\noindent\textbf{LLaVA-1.5-7B example ($T{=}64$, $L{=}32$, two pruning layers at 12 and 20).}
For this configuration, the Adaptive Stage uses $\alpha{=}2$ and reduces $n_0{=}576$ tokens to $2T{=}128$.
Substituting $\ell_1{=}12$ and $\ell_2{=}20$ into Eq.~\ref{eq:schedule_constraint}:
\begin{align*}
(20{-}12)\,a + (32{-}20)\,b &= 64\times(32{-}24), \\
8a + 12b &= 512, \\
2a + 3b &= 2T.
\end{align*}
With $a{=}37, b{=}18$: $2{\times}37 + 3{\times}18 = 128 = 2T$ \checkmark.
The resulting token distribution:

\begin{center}
\small
\begin{tabular}{cccc}
\toprule
Layers & Visual Tokens & \# Layers & Contribution \\
\midrule
0--11   & $2T=128$ & 12 & $12\times128=1536$ \\
12--19  & $a=37$   &  8 & $8\times37\;=296$  \\
20--31  & $b=18$   & 12 & $12\times18\;=216$ \\
\midrule
\multicolumn{2}{c}{Average} & 32 & $2048/32=\mathbf{64.0}$ \checkmark \\
\bottomrule
\end{tabular}
\end{center}

\noindent\textbf{Hardware.} Accuracy evaluation jobs use available 80GB NVIDIA accelerators, but their worker timing is not used for efficiency comparisons. All reported latency and speedup measurements follow the controlled single-A800-80GB protocol in \cref{sec:suppl_efficiency}.

\noindent\textbf{Evaluation.} Detailed benchmark splits, decoding settings, and scorer conventions are provided in \cref{sec:suppl_eval_protocol}.

\noindent\textbf{Code availability.} The implementation code, evaluation configurations, and schedule-validation utilities are available at \url{https://github.com/EasonAI-5589/starpro}.

\section{Additional Results and Efficiency Studies}
\label{sec:suppl_results}
\label{sec:suppl_efficiency}

\subsection{LLaVA Series}

The four tables below provide per-benchmark results for LLaVA-1.5-7B,
LLaVA-1.5-13B, LLaVA-NeXT-7B, and LLaVA-NeXT-13B.  Qwen3-VL and InternVL3
already use the full eight-metric comparison in the main paper and are not
repeated here.  Each LLaVA table resolves the aggregate into SEED, GQA,
ScienceQA-IMG, TextVQA, POPE, MME, MMBench-EN/CN, AI2D, and MMMU at three
compression levels.  This breakdown distinguishes broad aggregate retention
from benchmark-specific trade-offs that are hidden by the summary table in the
main paper.

\begin{table*}[t]
\centering
\small
\resizebox{\textwidth}{!}{%
\begin{tabular}{l|cccccccccc|cc}
\toprule
\textbf{Method} & \textbf{SEED} & \textbf{GQA} & \textbf{SQA}$^{\text{IMG}}$ & \textbf{VQA}$^{\text{Text}}$ & \textbf{POPE} & \textbf{MME} & \textbf{MMB}$^{\text{EN}}$ & \textbf{MMB}$^{\text{CN}}$ & \textbf{AI2D} & \textbf{MMMU} & \textbf{Acc.} & \textbf{Rel.} \\
\midrule
\multicolumn{13}{c}{\textit{Upper Bound: All 576 tokens (100\%)}} \\
\midrule
Baseline & 66 & 61.9 & 69.5 & 58.2 & 85.9 & 1508.3 & 64.7 & 58.1 & 55.5 & 35 & 63.02 & 100.0 \\
\midrule
\multicolumn{13}{c}{\textit{Retain 128 Tokens (\(\downarrow\) 77.8\%)}} \\
\midrule
\rowcolor[rgb]{.90, .96, .90} FastV (ECCV24) & 59.66 & 54 & \second{69.2} & 56.4 & 68.2 & 1376.5 & \second{63.0} & 55.9 & 53.6 & 34.3 & 58.31 & 92.5 \\
\rowcolor[rgb]{.87, .94, .88} PDrop (CVPR25) & 59.74 & 57.1 & \best{70.1} & 56.7 & 77.5 & \second{1434.1} & 62.3 & 55.3 & 52.7 & 34.1 & 59.72 & 94.8 \\
\rowcolor[rgb]{.84, .92, .86} SparseVLM (ICML25) & 61.72 & 57.3 & 69.0 & 56.3 & 83.1 & 1401.3 & 62.6 & \second{56.9} & 54.7 & \second{35.9} & 60.76 & 96.4 \\
\rowcolor[rgb]{.83, .90, .96} VisionZip (CVPR25) & 61.57 & 57.6 & 68.7 & 56.9 & 83.3 & 1437.3 & 62.1 & \best{57.0} & 54.5 & \second{35.9} & 60.94 & 96.7 \\
\rowcolor[rgb]{.78, .87, .95} DivPrune (CVPR25) & 62.37 & 59.4 & 68.6 & 55.9 & 87.0 & 1411.2 & 61.5 & 54.8 & 54.2 & 35.2 & 60.95 & 96.7 \\
\rowcolor[rgb]{.75, .85, .95} CDPruner (NeurIPS25) & \second{64.02} & \second{59.9} & 69.0 & 56.2 & \best{87.7} & 1430.6 & \best{63.1} & 55.0 & 54.6 & 35.4 & 61.65 & 97.8 \\
\rowcolor[rgb]{.80, .88, .96} SCOPE (NeurIPS25) & 62.92 & 59.4 & 68.5 & \best{57.1} & 85.9 & 1436.1 & 62.7 & \best{57.0} & 54.8 & 35.6 & 61.57 & 97.7 \\
\rowcolor[rgb]{.87, .94, .88} VScan (TMLR26) & 64.19 & 59.0 & 68.6 & 57.2 & 85.0 & 1425.6 & 61.7 & 57.1 & \best{55.5} & \best{36.7} & 61.63 & 97.8 \\
\rowcolor[rgb]{.90, .96, .90} HoloV (NeurIPS25) & 61.32 & 57.4 & 68.1 & 55.7 & 82.3 & 1437.4 & 61.9 & 56.7 & 54.6 & 35.1 & 60.50 & 96.0 \\
\rowcolor[rgb]{1.0, .92, .80} \textbf{STAR-Pro (Ours)} & \best{64.4} & \best{60.8} & 68.9 & \second{57.2} & \second{87.3} & \best{1456.3} & 62.4 & 57.7 & \best{55.0} & \best{36.7} & \best{62.32} & \best{98.9} \\
\midrule
\multicolumn{13}{c}{\textit{Retain 64 Tokens (\(\downarrow\) 88.9\%)}} \\
\midrule
\rowcolor[rgb]{.90, .96, .90} FastV (ECCV24) & 46.79 & 46.0 & \best{70.1} & 51.6 & 35.5 & 977.5 & 50.1 & 42.1 & 50.6 & 34.7 & 47.64 & 75.6 \\
\rowcolor[rgb]{.87, .94, .88} PDrop (CVPR25) & 48.92 & 46.1 & 68.8 & 49.2 & 40.8 & 965.9 & 48.0 & 36.6 & 52.7 & 34.3 & 47.37 & 75.2 \\
\rowcolor[rgb]{.84, .92, .86} SparseVLM (ICML25) & 53.92 & 52.0 & \second{69.2} & 52.1 & 69.7 & 1192.8 & 58.3 & 49.6 & 52.5 & 36.3 & 55.33 & 87.8 \\
\rowcolor[rgb]{.83, .90, .96} VisionZip (CVPR25) & 57.66 & 55.1 & 69.0 & \second{55.5} & 77.0 & 1371.5 & 60.1 & \second{55.4} & 53.3 & 35.4 & 58.70 & 93.1 \\
\rowcolor[rgb]{.78, .87, .95} DivPrune (CVPR25) & 60.11 & 57.5 & 68.0 & 54.5 & 85.5 & 1352.3 & 60.1 & 52.3 & 53.8 & 34.8 & 59.42 & 94.3 \\
\rowcolor[rgb]{.75, .85, .95} CDPruner (NeurIPS25) & \second{62.54} & \second{58.6} & 68.1 & 55.3 & \second{87.5} & \second{1419.0} & \second{61.1} & 53.2 & 53.9 & 35.3 & \second{60.65} & \second{96.2} \\
\rowcolor[rgb]{.80, .88, .96} SCOPE (NeurIPS25) & 61.38 & 58.3 & 68.7 & \best{56.5} & 84.1 & 1401.8 & 61.0 & 56.0 & 53.8 & \second{35.2} & 60.51 & 96.0 \\
\rowcolor[rgb]{.87, .94, .88} VScan (TMLR26) & 61.34 & 57.9 & 69.1 & 56.1 & 85.0 & 1375.0 & 61.3 & 55.7 & 54.4 & 35.6 & 60.52 & 96.0 \\
\rowcolor[rgb]{.90, .96, .90} HoloV (NeurIPS25) & 58.22 & 55.0 & 68.6 & 54.9 & 76.8 & 1362.5 & 59.2 & 55.6 & 51.9 & 34.8 & 58.31 & 92.5 \\
\rowcolor[rgb]{1.0, .92, .80} \textbf{STAR-Pro (Ours)} & \best{62.87} & \best{59.41} & 68.62 & \second{56.26} & \best{87.52} & \best{1421.4} & 60.97 & \second{56.44} & 53.69 & 35.11 & \best{61.20} & \best{97.1} \\
\midrule
\multicolumn{13}{c}{\textit{Retain 32 Tokens (\(\downarrow\) 94.4\%)}} \\
\midrule
\rowcolor[rgb]{.83, .90, .96} VisionZip (CVPR25) & 53.2 & 51.8 & 69.1 & 53.1 & 69.4 & 1263.6 & 57.0 & \second{50.3} & 51.7 & \best{35.0} & 55.38 & 87.9 \\
\rowcolor[rgb]{.78, .87, .95} DivPrune (CVPR25) & 56.92 & 54.9 & 68.6 & 52.9 & 81.5 & 1290.3 & 57.6 & 49.1 & \second{53.0} & 34.0 & 57.30 & 90.9 \\
\rowcolor[rgb]{.75, .85, .95} CDPruner (NeurIPS25) & \best{60.84} & \second{57.0} & \best{69.5} & 53.2 & \best{87.9} & \second{1360.3} & 59.6 & 49.6 & 53.6 & 33.7 & \best{59.30} & \best{94.1} \\
\rowcolor[rgb]{.80, .88, .96} SCOPE (NeurIPS25) & \second{59.16} & 56.2 & \second{69.4} & \best{54.8} & 80.2 & \best{1362.0} & \best{60.7} & \best{52.5} & 52.8 & \second{35.1} & \second{58.90} & \second{93.5} \\
\rowcolor[rgb]{.87, .94, .88} VScan (TMLR26) & 57.3 & 54.9 & 69.3 & 53.8 & 79.8 & 1298.1 & 58.7 & 51.8 & 52.6 & 35.4 & 57.85 & 91.8 \\
\rowcolor[rgb]{.90, .96, .90} HoloV (NeurIPS25) & 54.81 & 52.8 & 69.0 & 53.7 & 70.3 & 1261.7 & 58.2 & 51.5 & 53.1 & 33.1 & 55.96 & 88.8 \\
\rowcolor[rgb]{1.0, .92, .80} \textbf{STAR-Pro (Ours)} & \best{60.3} & \best{57.2} & 68.7 & 54.6 & \second{86.6} & 1358.1 & \second{60.1} & 55.1 & \best{53.4} & 34.4 & \best{59.83} & \best{94.9} \\
\bottomrule
\end{tabular}%
}
\normalsize
\caption{Performance comparison on LLaVA-1.5-7B. Acc. denotes the average across 10 benchmarks, with MME divided by 20. Rel. (\%) is performance relative to the corresponding baseline.}
\label{tab:pruning_comparison}

%
%
%
%
%

\end{table*}

\begin{table*}[t]
\centering
\small
\resizebox{\textwidth}{!}{%
\begin{tabular}{lccccccccccccc}
\toprule
\textbf{Method} & \textbf{SEED} & \textbf{GQA} & \textbf{SQA}$^{\text{IMG}}$ & \textbf{VQA}$^{\text{Text}}$ & \textbf{POPE} & \textbf{MME} & \textbf{MMB}$^{\text{EN}}$ & \textbf{MMB}$^{\text{CN}}$ & \textbf{AI2D} & \textbf{MMMU} & \textbf{Acc.} & \textbf{Rel.} \\
\midrule
\multicolumn{13}{c}{\textit{Upper Bound: All 576 tokens (100\%)}} \\
\midrule
Baseline & 68.22 & 63.3 & 72.8 & 61.2 & 86 & 1531.2 & 68.5 & 63.5 & 60.8 & 36.4 & 65.73 & 100.0 \\
\midrule
\multicolumn{13}{c}{\textit{Retain 128 Tokens ($\downarrow$ 77.8\%)}} \\
\midrule
\rowcolor[rgb]{.90, .96, .90} FastV (ECCV24) & 55.5 & 58.3 & 74.2 & 58.6 & 75.5 & 1460.6 & 66.1 & 62.3 & 58.1 & 36.9 & 61.85 & 94.1 \\
\rowcolor[rgb]{.87, .94, .88} PDrop (CVPR25) & 65.6 & 61 & 73.3 & 60.2 & 83.4 & 1494.3 & 66.7 & 62.7 & 59.1 & 36.1 & 64.28 & 97.8 \\
\rowcolor[rgb]{.84, .92, .86} SparseVLM (ICML25) & 65.3 & 59.6 & 74.3 & 59.3 & 85 & 1487.9 & 68.4 & 62.6 & 58.1 & 37.4 & 64.44 & 98.0 \\
\rowcolor[rgb]{.83, .90, .96} VisionZip (CVPR25) & 63.7 & 57.8 & 73.8 & 58.9 & 82.5 & 1448.2 & 66.8 & 62.3 & 57 & 37.9 & 63.31 & 96.3 \\
\rowcolor[rgb]{.75, .85, .95} CDPruner (NeurIPS25) & 65 & 59.7 & 72.9 & 58.4 & 87.2 & 1468.9 & 66.9 & 61.4 & 58.4 & 35.6 & 63.89 & 97.2 \\
\rowcolor[rgb]{.80, .88, .96} SCOPE (NeurIPS25) & 64.8 & 59.3 & 73.7 & 58.9 & 85.9 & 1444.3 & 66.8 & 62.9 & 57.5 & 36.6 & 63.86 & 97.2 \\
\rowcolor[rgb]{.78, .87, .95} DivPrune (CVPR25) & 64.1 & 59.2 & 72.8 & 58 & 86.8 & 1457.7 & 66.3 & 60.7 & 57.7 & 36.8 & 63.53 & 96.7 \\
\rowcolor[rgb]{.90, .96, .90} HoloV (NeurIPS25) & 63.9 & 58.1 & 73.4 & 57.9 & 82 & 1445.1 & 66 & 61.3 & 56.8 & 36.6 & 62.83 & 95.6 \\
\rowcolor[rgb]{.87, .94, .88} VScan (TMLR26) & 65.8 & 59.3 & 73.4 & 58.6 & 85.1 & 1469.2 & 65.9 & 62.4 & 58.6 & 37.1 & 63.97 & 97.3 \\
\rowcolor[rgb]{1.0, .92, .80} \textbf{STAR-Pro} & 66.1 & 60.2 & 72.7 & 59.5 & 87.2 & 1492.7 & 67 & 63.1 & 57.9 & 35.4 & 64.37 & 97.9 \\
\multicolumn{13}{c}{\textit{Retain 64 Tokens ($\downarrow$ 88.9\%)}} \\
\midrule
\rowcolor[rgb]{.90, .96, .90} FastV (ECCV24) & 54.6 & 51.9 & 73.1 & 53.4 & 56.9 & 1246.4 & 59.2 & 55.1 & 55.5 & 36.8 & 55.88 & 85.0 \\
\rowcolor[rgb]{.87, .94, .88} PDrop (CVPR25) & 58.7 & 54 & 73.2 & 55.3 & 66.1 & 1259.6 & 62.6 & 56.6 & 57.9 & 35.6 & 58.30 & 88.7 \\
\rowcolor[rgb]{.84, .92, .86} SparseVLM (ICML25) & 60.8 & 55.9 & 73 & 57.1 & 77.9 & 1374.3 & 65.2 & 60.3 & 56.7 & 36.4 & 61.20 & 93.1 \\
\rowcolor[rgb]{.83, .90, .96} VisionZip (CVPR25) & 60.2 & 56.1 & 74.2 & 57.5 & 76 & 1397.4 & 64.3 & 61.1 & 56.8 & 35.9 & 61.20 & 93.1 \\
\rowcolor[rgb]{.75, .85, .95} CDPruner (NeurIPS25) & 64 & 59.2 & 72.6 & 57.4 & 87 & 1453.4 & 64.7 & 58.9 & 57.4 & 35.1 & 62.90 & 95.7 \\
\rowcolor[rgb]{.80, .88, .96} SCOPE (NeurIPS25) & 63.6 & 58.6 & 73.6 & 58.1 & 83.1 & 1456.9 & 66 & 62.4 & 57.6 & 35.2 & 63.10 & 96.0 \\
\rowcolor[rgb]{.78, .87, .95} DivPrune (CVPR25) & 62.2 & 57.9 & 71.7 & 57.4 & 84.5 & 1454.2 & 64.1 & 59.8 & 57 & 35.7 & 62.30 & 94.8 \\
\rowcolor[rgb]{.90, .96, .90} HoloV (NeurIPS25) & 60.5 & 56.1 & 74.3 & 57.3 & 75.7 & 1403 & 63.8 & 60.2 & 56.7 & 35.3 & 61.01 & 92.8 \\
\rowcolor[rgb]{.87, .94, .88} VScan (TMLR26) & 63.5 & 58.6 & 73.5 & 58.7 & 84.3 & 1446.8 & 64.7 & 61.5 & 57.2 & 36.9 & 63.12 & 96.0 \\
\rowcolor[rgb]{1.0, .92, .80} \textbf{STAR-Pro} & 64.3 & 59.4 & 73.1 & 58.6 & 86.8 & 1456.6 & 66.2 & 62 & 58.1 & 36.2 & 63.75 & 97.0 \\
\multicolumn{13}{c}{\textit{Retain 32 Tokens ($\downarrow$ 94.4\%)}} \\
\midrule
\rowcolor[rgb]{.83, .90, .96} VisionZip (CVPR25) & 55.8 & 52.7 & 72.7 & 55.1 & 67 & 1254.7 & 60.5 & 55.7 & 56.4 & 35.6 & 57.42 & 87.4 \\
\rowcolor[rgb]{.75, .85, .95} CDPruner (NeurIPS25) & 62.5 & 58.4 & 72.2 & 55.2 & 87.7 & 1422.2 & 63.6 & 56.2 & 57.2 & 34 & 61.81 & 94.0 \\
\rowcolor[rgb]{.80, .88, .96} SCOPE (NeurIPS25) & 61.4 & 57.4 & 72.2 & 57 & 77.7 & 1417.1 & 63.8 & 60.3 & 56.8 & 35.2 & 61.27 & 93.2 \\
\rowcolor[rgb]{.87, .94, .88} VScan (TMLR26) & 59.3 & 53.9 & 73.2 & 55.1 & 66.9 & 1282.8 & 61.9 & 56.2 & 57.3 & 34.8 & 58.27 & 88.7 \\
\rowcolor[rgb]{.78, .87, .95} DivPrune (CVPR25) & 59.5 & 56.2 & 70.9 & 54.6 & 79.3 & 1405.2 & 61.7 & 57.2 & 57.2 & 35 & 60.19 & 91.6 \\
\rowcolor[rgb]{.90, .96, .90} HoloV (NeurIPS25) & 57.2 & 53.4 & 71.5 & 55.6 & 69.1 & 1281.8 & 62.7 & 58.2 & 56.4 & 36.2 & 58.44 & 88.9 \\
\rowcolor[rgb]{1.0, .92, .80} \textbf{STAR-Pro} & 62.9 & 58 & 73.1 & 57.4 & 85.7 & 1464.1 & 65.6 & 61 & 57.8 & 36.9 & 63.16 & 96.1 \\
\bottomrule
\end{tabular}%
}
\normalsize
\caption{Performance comparison on LLaVA-1.5-13B. Acc. denotes the average across 10 benchmarks, with MME divided by 20. Rel. (\%) is performance relative to the corresponding baseline.}
\label{tab:pruning_comparison_13b}

\end{table*}

\begin{table*}[t]
\centering
\small
\resizebox{\textwidth}{!}{%
\begin{tabular}{lccccccccccccc}
\toprule
\textbf{Method} & \textbf{SEED} & \textbf{GQA} & \textbf{SQA}$^{\text{IMG}}$ & \textbf{VQA}$^{\text{Text}}$ & \textbf{POPE} & \textbf{MME} & \textbf{MMB}$^{\text{EN}}$ & \textbf{MMB}$^{\text{CN}}$ & \textbf{AI2D} & \textbf{MMMU} & \textbf{Acc.} & \textbf{Rel.} \\
\midrule
\multicolumn{13}{c}{\textit{Upper Bound: All 2880 tokens (100\%)}} \\
\midrule
Baseline & 69.66 & 62.5 & 67.5 & 60.3 & 86.8 & 1511.8 & 65.8 & 57.3 & 64.7 & 35.2 & 64.54 & 100.0 \\
\midrule

\multicolumn{13}{c}{\textit{Retain 640 Tokens ($\downarrow$ 77.8\%)}} \\
\midrule
\rowcolor[rgb]{.90, .96, .90} FastV (ECCV24) & 64.25 & 58.9 & 67.4 & 58.1 & 79.5 & 1412.6 & 63.1 & 53.5 & 65.1 & 35 & 61.55 & 95.4 \\
\rowcolor[rgb]{.87, .94, .88} PDrop (CVPR25) & 64.68 & 60 & 66.7 & 57.8 & 83.8 & 1475.9 & 64.1 & 55.2 & 64.6 & 35.1 & 62.58 & 97.0 \\
\rowcolor[rgb]{.84, .92, .86} SparseVLM (ICML25) & 67.17 & 61.2 & 67.6 & 59.7 & 85.3 & 1456.8 & 65.9 & 58.6 & 64.7 & 35.1 & 63.81 & 98.9 \\
\rowcolor[rgb]{.83, .90, .96} VisionZip (CVPR25) & 66.75 & 61.2 & 68.1 & 60 & 86.1 & 1493.4 & 65.4 & 58.1 & 65.3 & 35.9 & 64.15 & 99.4 \\
\rowcolor[rgb]{.75, .85, .95} CDPruner (NeurIPS25) & 68.9 & 62.5 & 67.7 & 58.3 & 87.2 & 1468.9 & 65.9 & 57.4 & 65.3 & 35.2 & 64.18 & 99.5 \\
\rowcolor[rgb]{.80, .88, .96} SCOPE (NeurIPS25) & 67.77 & 62.1 & 68 & 60.1 & 86.7 & 1485.5 & 66.2 & 58.2 & 64.9 & 34.6 & \second{64.28} & \second{99.6} \\
\rowcolor[rgb]{.78, .87, .95} DivPrune (CVPR25) & 67.58 & 61.9 & 67.8 & 57 & 86.9 & 1469.7 & 65.8 & 57.3 & 65.6 & 35.4 & 63.88 & 99.0 \\
\rowcolor[rgb]{.90, .96, .90} HoloV (NeurIPS25) & 66.8 & 60.5 & 66.5 & 54.8 & 85.6 & 1476.6 & 63.9 & 57.3 & 62.8 & 34.7 & 62.67 & 97.1 \\
\rowcolor[rgb]{.87, .94, .88} VScan (TMLR26) & 66.7 & 62.4 & 67.9 & 58.3 & 85.6 & 1473.8 & 66.3 & 59.8 & 66.7 & 34.9 & 64.22 & 99.5 \\
\rowcolor[rgb]{1.0, .92, .80} \textbf{STAR-Pro} & 68.95 & 62.4 & 68.8 & 59.1 & 87.6 & 1479.1 & 66 & 58 & 65.6 & 32.2 & 64.26 & 99.6 \\

\multicolumn{13}{c}{\textit{Retain 320 Tokens ($\downarrow$ 88.9\%)}} \\
\midrule
\rowcolor[rgb]{.90, .96, .90} FastV (ECCV24) & 52.76 & 49.8 & 66.6 & 52.2 & 49.5 & 1099 & 53.4 & 42.5 & 62.2 & 34.9 & 51.88 & 80.4 \\
\rowcolor[rgb]{.87, .94, .88} PDrop (CVPR25) & 53.83 & 50.4 & 66.7 & 49 & 60.8 & 1171.5 & 55.5 & 44.7 & 62.5 & 34.1 & 53.61 & 83.1 \\
\rowcolor[rgb]{.84, .92, .86} SparseVLM (ICML25) & 62.14 & 57.9 & 67.2 & 56.5 & 76.9 & 1386.1 & 63.1 & 56.7 & 63 & 34 & 60.67 & 94.0 \\
\rowcolor[rgb]{.83, .90, .96} VisionZip (CVPR25) & 63.44 & 58.9 & 67.5 & 58.8 & 82.2 & 1397.1 & 62.8 & 55.6 & 63.2 & 35.8 & 61.81 & 95.8 \\
\rowcolor[rgb]{.75, .85, .95} CDPruner (NeurIPS25) & 67.35 & 61.5 & 67.3 & 57.4 & 87.3 & 1456.3 & 64.9 & 55.9 & 63.8 & 34.6 & 63.29 & 98.1 \\
\rowcolor[rgb]{.80, .88, .96} SCOPE (NeurIPS25) & 66.22 & 60.9 & 68 & 58.3 & 85 & 1477.0 & 65 & 57.6 & 63.4 & 34.9 & 63.32 & 98.1 \\
\rowcolor[rgb]{.78, .87, .95} DivPrune (CVPR25) & 65.57 & 61.1 & 67.7 & 56.2 & 84.7 & 1423.3 & 63.9 & 55.7 & 65.8 & 35.4 & 62.72 & 97.2 \\
\rowcolor[rgb]{.90, .96, .90} HoloV (NeurIPS25) & 64.6 & 59.2 & 67.1 & 55.5 & 82.8 & 1465.8 & 63.8 & 55.7 & 62.2 & 36.1 & 62.03 & 96.1 \\
\rowcolor[rgb]{.87, .94, .88} VScan (TMLR26) & 63.3 & 59.9 & 68.4 & 57.2 & 83.0 & 1411.1 & 63.5 & 56.4 & 63.5 & 35.4 & 62.11 & 96.2 \\
\rowcolor[rgb]{1.0, .92, .80} \textbf{STAR-Pro} & 67.4 & 61.8 & 67.9 & 58.1 & 87.5 & 1468.1 & 65.4 & 58.3 & 65.3 & 33.8 & \best{63.89} & \best{99.0} \\

\multicolumn{13}{c}{\textit{Retain 160 Tokens ($\downarrow$ 94.4\%)}} \\
\midrule
\rowcolor[rgb]{.83, .90, .96} VisionZip (CVPR25) & 58.11 & 55.2 & 67.9 & 56 & 74.9 & 1327.8 & 58.2 & 50.4 & 62.8 & 34 & 58.39 & 90.5 \\
\rowcolor[rgb]{.75, .85, .95} CDPruner (NeurIPS25) & 65.92 & 60.7 & 67.2 & 55.4 & 86.9 & 1432.2 & 63.6 & 53.7 & 62.7 & 35.4 & \second{62.31} & \second{96.6} \\
\rowcolor[rgb]{.80, .88, .96} SCOPE (NeurIPS25) & 64.35 & 60 & 67.1 & 56.8 & 81.3 & 1402.2 & 63.3 & 56.4 & 62.8 & 34.6 & 61.68 & 95.6 \\
\rowcolor[rgb]{.78, .87, .95} DivPrune (CVPR25) & 62.91 & 59.3 & 67.1 & 54.1 & 80 & 1356.6 & 62.9 & 53.7 & 65.5 & 34.4 & 60.77 & 94.2 \\
\rowcolor[rgb]{.90, .96, .90} HoloV (NeurIPS25) & 60.7 & 57 & 67.2 & 55.4 & 78.1 & 1349.4 & 62.1 & 54.2 & 60.7 & 34.4 & 59.73 & 92.5 \\
\rowcolor[rgb]{.87, .94, .88} VScan (TMLR26) & 58.3 & 56.6 & 68.4 & 53.1 & 78.1 & 1255.5 & 59.2 & 49.1 & 62.4 & 35.0 & 58.29 & 90.3 \\
\rowcolor[rgb]{1.0, .92, .80} \textbf{STAR-Pro} & 65.21 & 60.4 & 67.7 & 56.4 & 86.2 & 1421.9 & 64.4 & 56.9 & 63.3 & 34.3 & \best{62.59} & \best{97.0} \\
\bottomrule
\end{tabular}%
}
\normalsize
\caption{Performance comparison on LLaVA-NeXT-7B. Acc. denotes the average across 10 benchmarks, with MME divided by 20. Rel. (\%) is performance relative to the corresponding baseline.}
\label{tab:pruning_comparison_next}

\end{table*}

\begin{table*}[t]
\centering
\small
\resizebox{\textwidth}{!}{%
\begin{tabular}{lccccccccccccc}
\toprule
\textbf{Method} & \textbf{SEED} & \textbf{GQA} & \textbf{SQA}$^{\text{IMG}}$ & \textbf{VQA}$^{\text{Text}}$ & \textbf{POPE} & \textbf{MME} & \textbf{MMB}$^{\text{EN}}$ & \textbf{MMB}$^{\text{CN}}$ & \textbf{AI2D} & \textbf{MMMU} & \textbf{Acc.} & \textbf{Rel.} \\
\midrule
\multicolumn{13}{c}{\textit{Upper Bound: All 2880 tokens (100\%)}} \\
\midrule
Baseline & 71.58 & 64.4 & 73.1 & 63.2 & 85.3 & 1539.5 & 68.5 & 61.2 & 70.1 & 35.7 & 67.0 & 100.0 \\
\midrule
\multicolumn{13}{c}{\textit{Retain 640 Tokens ($\downarrow$ 77.8\%)}} \\
\midrule
\rowcolor[rgb]{.90, .96, .90} FastV (ECCV24) & 67.4 & 60.9 & 71.7 & 60.7 & 80.2 & 1516.7 & 65.5 & 59.9 & 67.7 & 35.8 & 64.6 & 96.4 \\
\rowcolor[rgb]{.87, .94, .88} PDrop (CVPR25) & 68.8 & 62.8 & 71.7 & 62.1 & 84.4 & 1559.1 & 66.6 & 60.8 & 69 & 35.9 & 66.0 & 98.5 \\
\rowcolor[rgb]{.84, .92, .86} SparseVLM (ICML25) & 69.7 & 62.7 & 72.5 & 62.8 & 85.6 & 1562.7 & 68.5 & 64.0 & 68.1 & 37.6 & 67.0 & 99.9 \\
\rowcolor[rgb]{.83, .90, .96} VisionZip (CVPR25) & 68.7 & 62.9 & 70.8 & 62.1 & 57.8 & 1549.2 & 67.5 & 62.6 & 68.3 & 36.9 & 63.5 & 94.8 \\
\rowcolor[rgb]{.75, .85, .95} CDPruner (NeurIPS25) & 70.6 & 64 & 71.8 & 61 & 87.5 & 1560.2 & 68.2 & 62.2 & 68.7 & 37.8 & 67.0 & 100.0 \\
\rowcolor[rgb]{.80, .88, .96} SCOPE (NeurIPS25) & 69.9 & 63.5 & 71.7 & 62.4 & 86.5 & 1573.1 & 67.6 & 63.2 & 68.6 & 36.9 & 66.9 & 99.8 \\
\rowcolor[rgb]{.78, .87, .95} DivPrune (CVPR25) & 69.4 & 63.5 & 72.2 & 59.2 & 86.5 & 1526.1 & 67.5 & 62.9 & 68.4 & 37.8 & 66.4 & 99.1 \\
\rowcolor[rgb]{.90, .96, .90} HoloV (NeurIPS25) & 68.7 & 61.7 & 70.3 & 57.3 & 85.5 & 1497.7 & 67.5 & 61.9 & 66.6 & 37 & 65.1 & 97.2 \\
\rowcolor[rgb]{.87, .94, .88} VScan (TMLR26) & 68.2 & 62.8 & 72.2 & 61.8 & 85.2 & 1553.6 & 67.6 & 62.7 & 69.9 & 36.6 & 66.5 & 99.2 \\
\rowcolor[rgb]{1.0, .92, .80} \textbf{STAR-Pro} & 71.33 & 64.2 & 72.8 & 62.2 & 87.4 & 1545.2 & 68.5 & 63.1 & 69.8 & 35.8 & 67.2 & 100.3 \\
\multicolumn{13}{c}{\textit{Retain 320 Tokens ($\downarrow$ 88.9\%)}} \\
\midrule
\rowcolor[rgb]{.90, .96, .90} FastV (ECCV24) & 59.2 & 54.6 & 70.5 & 55.4 & 63.6 & 1279 & 59.8 & 54.4 & 65 & 35.8 & 58.2 & 86.9 \\
\rowcolor[rgb]{.87, .94, .88} PDrop (CVPR25) & 62.4 & 57.7 & 72.1 & 56.2 & 74.6 & 1386.3 & 62.8 & 55.3 & 66.7 & 35.7 & 61.3 & 91.5 \\
\rowcolor[rgb]{.84, .92, .86} SparseVLM (ICML25) & 65.8 & 60.9 & 70.9 & 60.0 & 81.5 & 1491.6 & 67.3 & 63.5 & 66.8 & 37.4 & 64.9 & 96.8 \\
\rowcolor[rgb]{.83, .90, .96} VisionZip (CVPR25) & 65.3 & 60.7 & 70.2 & 60.7 & 82.2 & 1487.3 & 65.9 & 62.3 & 67.1 & 37.2 & 64.6 & 96.4 \\
\rowcolor[rgb]{.75, .85, .95} CDPruner (NeurIPS25) & 68.8 & 63.2 & 71.3 & 58.8 & 87.7 & 1500.1 & 65.6 & 61.9 & 68.3 & 37.3 & 65.8 & 98.2 \\
\rowcolor[rgb]{.80, .88, .96} SCOPE (NeurIPS25) & 67.9 & 63 & 71 & 60.8 & 85.2 & 1509.3 & 66.6 & 63.1 & 68.3 & 36.6 & 65.8 & 98.2 \\
\rowcolor[rgb]{.78, .87, .95} DivPrune (CVPR25) & 67.2 & 61.8 & 72.3 & 57.6 & 85.2 & 1473 & 65.9 & 61.9 & 67.7 & 37.2 & 65.0 & 97.1 \\
\rowcolor[rgb]{.90, .96, .90} HoloV (NeurIPS25) & 66.5 & 60.7 & 69.9 & 58.2 & 83.2 & 1496.6 & 66.3 & 62.2 & 66.5 & 35.8 & 64.4 & 96.1 \\
\rowcolor[rgb]{.87, .94, .88} VScan (TMLR26) & 64.9 & 61.0 & 72.4 & 59.3 & 82.0 & 1496.4 & 65.2 & 59.8 & 66.9 & 36.3 & 64.3 & 95.9 \\
\rowcolor[rgb]{1.0, .92, .80} \textbf{STAR-Pro} & 69.19 & 63.4 & 72 & 60.6 & 87.2 & 1524.2 & 67.3 & 63.7 & 68.2 & 35.2 & 66.3 & 98.9 \\
\multicolumn{13}{c}{\textit{Retain 160 Tokens ($\downarrow$ 94.4\%)}} \\
\midrule
\rowcolor[rgb]{.83, .90, .96} VisionZip (CVPR25) & 61.2 & 57.8 & 69.7 & 58.5 & 76.8 & 1393.9 & 64.1 & 60 & 65.7 & 37 & 62.0 & 92.6 \\
\rowcolor[rgb]{.75, .85, .95} CDPruner (NeurIPS25) & 67 & 62.1 & 70.5 & 56.7 & 88.2 & 1478.1 & 65.8 & 60.1 & 67.6 & 37.6 & 65.0 & 96.9 \\
\rowcolor[rgb]{.80, .88, .96} SCOPE (NeurIPS25) & 66.1 & 61.3 & 71.2 & 59.2 & 82.7 & 1473.7 & 66.2 & 62.9 & 67 & 36.8 & 64.7 & 96.6 \\
\rowcolor[rgb]{.78, .87, .95} DivPrune (CVPR25) & 64.5 & 60 & 71.4 & 56.3 & 81.9 & 1436.7 & 65.1 & 60.9 & 67.3 & 36.6 & 63.6 & 94.9 \\
\rowcolor[rgb]{.90, .96, .90} HoloV (NeurIPS25) & 62.5 & 58.5 & 70 & 57.7 & 78.5 & 1424.5 & 65.7 & 61.6 & 65.4 & 36.9 & 62.8 & 93.7 \\
\rowcolor[rgb]{.87, .94, .88} VScan (TMLR26) & 61.5 & 58.2 & 70.3 & 55.0 & 75.5 & 1385.1 & 61.5 & 53.4 & 63.4 & 34.1 & 60.2 & 89.9 \\
\rowcolor[rgb]{1.0, .92, .80} \textbf{STAR-Pro} & 66.94 & 62.3 & 71.4 & 59.3 & 86 & 1519.8 & 65.3 & 62.7 & 68 & 35.4 & 65.3 & 97.5 \\
\bottomrule
\end{tabular}%
}
\normalsize
\caption{Performance comparison on LLaVA-NeXT-13B. Acc. denotes the average across 10 benchmarks, with MME divided by 20. Rel. (\%) is performance relative to the corresponding baseline.}
\label{tab:pruning_comparison_next13b}

\end{table*}

\paragraph{Cross-benchmark overview.}
Across the twelve model--budget settings, STAR-Pro has the highest aggregate
in ten, ties the best reported aggregate on LLaVA-NeXT-7B at 640 tokens, and is
only 0.1 relative-performance points behind SparseVLM on LLaVA-1.5-13B at 128
tokens.  Its separation from the strongest alternative is generally clearer
at the tightest budget: the margins are 0.8, 2.1, 0.4, and 0.6 points on
LLaVA-1.5-7B, LLaVA-1.5-13B, LLaVA-NeXT-7B, and LLaVA-NeXT-13B, respectively.
At benchmark level, GQA and MMBench-CN are recurring strengths, whereas POPE
and MME sometimes favor CDPruner or SCOPE, especially on LLaVA-NeXT-7B.  The
ScienceQA-IMG, AI2D, and MMMU winners are also model dependent.  Thus the
aggregate trend is an observed balance across complementary tasks rather than
uniform per-column dominance; the model-specific results below identify where
that balance changes.

\paragraph{LLaVA-1.5-7B}
STAR-Pro obtains the highest aggregate at all three budgets, retaining 98.9\%,
97.1\%, and 94.9\% of the baseline performance at 128, 64, and 32 tokens,
respectively.  The margin over the strongest competing aggregate is 1.1 points
at 128 tokens, 0.9 points at 64 tokens, and 0.8 points at 32 tokens in relative
performance.  The gradual rather than abrupt decline across budgets is also
consistent with STAR-Pro's coverage-first allocation: the method remains close
to the full-model profile even after retaining only 5.6\% of the original
visual tokens.

\noindent\textbf{Per-benchmark breakdown.}
On SEED and GQA, STAR-Pro leads the 128-token rows with 64.4 and 60.8 and the
64-token rows with 62.87 and 59.41; at 32 tokens, it remains best on GQA (57.2)
and is second to CDPruner on SEED (60.3 versus 60.84).  ScienceQA-IMG changes
little across the three STAR-Pro budgets, staying in the narrow 68.6--68.9
range, although PDrop/FastV and CDPruner are higher
at individual budgets.  TextVQA follows a similarly stable pattern: STAR-Pro
ties the best 128-token score at 57.2 and is within 0.24 and 0.2 points of SCOPE
at 64 and 32 tokens.  On the hallucination and perception pair, STAR-Pro is
near the best POPE value at 128 tokens, leads POPE at 64 tokens, and remains
second at 32 tokens; its MME score is best at 128 and 64 tokens and only 3.9
points below SCOPE at 32 tokens.  The multilingual results are asymmetric:
MMBench-CN is strongest for STAR-Pro at all three budgets (57.7, 56.44, and
55.1), whereas MMBench-EN remains competitive but is led by CDPruner/VScan or
SCOPE depending on the budget.  Finally, STAR-Pro is near the best AI2D result
throughout and ties the best MMMU score at 128 tokens, while its lower-budget
MMMU values illustrate that the aggregate lead does not require winning every
reasoning benchmark.

\paragraph{LLaVA-1.5-13B}
At 128 retained tokens, STAR-Pro preserves 97.9\% of the baseline aggregate;
SparseVLM is marginally higher at 98.0\%, while STAR-Pro ties the highest
POPE score.  STAR-Pro becomes the aggregate leader at 64 tokens (97.0\%) and
widens its advantage under the most aggressive 32-token setting, where its
96.1\% relative performance exceeds CDPruner by 2.1 points.  The scale transfer
is therefore most visible under the tightest budget rather than only at the
easier operating point.

\noindent\textbf{Per-benchmark breakdown.}
SEED is consistently strong: STAR-Pro leads at 128, 64, and 32 tokens with
66.1, 64.3, and 62.9, respectively.  It also leads GQA at 64 tokens, while
PDrop is 0.8 points higher at 128 tokens and CDPruner is 0.4 points higher at
32 tokens.  ScienceQA-IMG remains stable near 73 across the sweep, with other
methods occasionally higher, whereas TextVQA changes from a competitive 59.5
at 128 tokens to the best reported score, 57.4, at 32 tokens.  On POPE,
STAR-Pro ties CDPruner at 128 tokens and stays close at 64 tokens, but trails
CDPruner by 2.0 points at 32 tokens.  MME exhibits the opposite behavior:
STAR-Pro is close to the best values at 128 and 64 tokens and becomes the clear
leader at 32 tokens with 1464.1.  Across MMBench, its Chinese score is best at
128 tokens, its English score is best at 64 tokens, and both language splits
are best at 32 tokens.  AI2D is strongest for STAR-Pro at 64 and 32 tokens,
while MMMU becomes strongest at 32 tokens (36.9).  Altogether, the 32-token row
leads seven of the ten individual benchmarks, explaining why the aggregate
margin grows as compression becomes more aggressive.

\paragraph{LLaVA-NeXT-7B}
With five image crops and 2,880 starting tokens, STAR-Pro retains 99.6\% of
baseline performance at 640 tokens, tied at the reported precision with the
best aggregate.  It then becomes the sole aggregate leader at 320 tokens
(99.0\%) and remains first at 160 tokens (97.0\%), 0.4 relative-performance
points above CDPruner.  These results are notable because the any-resolution
input contains five times as many initial visual tokens as LLaVA-1.5, yet the
same two-stage principle remains effective across all three compression
levels.

\noindent\textbf{Per-benchmark breakdown.}
At 640 tokens, STAR-Pro leads SEED (68.95), ScienceQA-IMG (68.8), and POPE
(87.6), while its GQA score is only 0.1 below CDPruner.  At 320 tokens it leads
both SEED and GQA, and at 160 tokens it remains within 0.71 and 0.3 points of
CDPruner on those two benchmarks.  TextVQA is not the source of the aggregate
advantage: SCOPE or VisionZip is modestly higher at each budget.  The same
qualification holds for MME, where VisionZip, SCOPE, or CDPruner leads the
corresponding budget.  POPE, by contrast, is strongest for STAR-Pro at 640 and
320 tokens and remains close to CDPruner at 160 tokens.  On the multilingual
pair, STAR-Pro is competitive but not best at 640 tokens, then leads both
MMBench-EN and MMBench-CN at 320 and 160 tokens.  AI2D follows the same
late-budget pattern, becoming best for STAR-Pro at 160 tokens, while MMMU is
consistently led by another method.  The aggregate ranking therefore comes
from balanced preservation of general VQA, hallucination, multilingual, and
diagram understanding rather than from the MMMU column.

\paragraph{LLaVA-NeXT-13B}
This model starts from 2,880 visual tokens.  STAR-Pro retains 100.3\% of the
baseline aggregate at 640 tokens and 97.5\% at 160 tokens, the highest reported
aggregates at both budgets.  At 320 tokens, STAR-Pro also leads the aggregate
by 0.7 relative-performance points over CDPruner and SCOPE.  The fact that the
640-token aggregate slightly exceeds the unpruned reference should be read as
benchmark-level variance under moderate token filtering, not as a claim that
pruning universally improves the base model.

\noindent\textbf{Per-benchmark breakdown.}
STAR-Pro leads SEED and GQA at 640 and 320 tokens; at 160 tokens it remains best
on GQA and is only 0.06 points below CDPruner on SEED.  ScienceQA-IMG is best
for STAR-Pro at 640 tokens, remains competitive at 320 tokens, and ties
DivPrune at 160 tokens.  TextVQA is led by SparseVLM or SCOPE at the two larger
budgets, but STAR-Pro becomes best at 160 tokens with 59.3.  POPE stays close
to CDPruner at 640 and 320 tokens but shows a larger 2.2-point deficit at 160
tokens.  MME shows the reverse trend: SCOPE is higher at 640 tokens, whereas
STAR-Pro leads at 320 and 160 tokens with 1524.2 and 1519.8.  On MMBench-EN,
STAR-Pro ties the best 640- and 320-token scores but trails SCOPE at 160 tokens;
on MMBench-CN, it is below SparseVLM at 640 tokens, best at 320 tokens, and
second to SCOPE by 0.2 points at 160 tokens.  AI2D is consistently near the
top and becomes best at 160 tokens, while MMMU remains below CDPruner or
SparseVLM across the three budgets.  The per-column pattern therefore supports
strong cross-task transfer while preserving the expected method-dependent
trade-offs.

\subsection{LLaVA-Video}

\begin{table*}[t]
\centering
\small
\resizebox{\textwidth}{!}{%
\begin{tabular}{l|c|ccc|cccc|cc}
\toprule
\textbf{Method} & \textbf{MLVU} & \multicolumn{3}{c|}{\textbf{LongVideoBench}} & \multicolumn{4}{c|}{\textbf{Video-MME}} & \textbf{Acc.} & \textbf{Rel.} \\
\textbf{Metric} & \textbf{m-avg} & \textbf{val} & \textbf{perception} & \textbf{relation} & \textbf{w/o sub} & \textbf{short} & \textbf{medium} & \textbf{long} & & \\
\midrule
\multicolumn{11}{c}{\textit{Baseline ($64 \times 169$ tokens, 100\%)}} \\
\midrule
Baseline & 67.7 & 59.0 & 65.0 & 53.8 & 63.6 & 76.6 & 61.2 & 53.1 & 63.4 & 100.0 \\
\midrule
\multicolumn{11}{c}{\textit{Retain $64 \times 32$ Tokens ($\downarrow$ 81.1\%)}} \\
\midrule
\rowcolor[rgb]{.90, .96, .90}
FastV (ECCV24) & 58.5 & 52.4 & 57.0 & 48.5 & 56.0 & 63.8 & 55.9 & 48.4 & 55.6 & 87.7 \\
\rowcolor[rgb]{.84, .92, .86}
SparseVLM (ICML25) & 60.7 & 53.7 & 58.1 & 49.9 & 59.0 & 69.8 & 56.9 & 50.3 & 57.8 & 91.1 \\
\rowcolor[rgb]{.80, .88, .95}
CDPruner (NeurIPS25) & \second{63.0} & 56.5 & 61.0 & \second{52.7} & 60.5 & 71.9 & \second{58.6} & 51.0 & 60.0 & 94.6 \\
\rowcolor[rgb]{.78, .87, .95}
DivPrune (CVPR25) & 61.5 & 56.4 & \second{62.1} & 51.4 & 59.3 & 69.9 & 57.9 & 50.2 & 59.1 & 93.1 \\
\rowcolor[rgb]{.87, .94, .88}
VScan (TMLR26) & 62.4 & \best{57.8} & \second{63.2} & \best{53.1} & \second{60.9} & \second{72.7} & 58.2 & \best{51.7} & \second{60.4} & \second{95.2} \\
\rowcolor[rgb]{1.0, .95, .88}
\textbf{STAR-Pro (Ours)} & \best{65.4} & \second{57.0} & \best{63.7} & 51.1 & \best{61.6} & \best{74.4} & \best{59.1} & \second{51.1} & \best{61.3} & \best{96.7} \\
\midrule
\multicolumn{11}{c}{\textit{Retain $64 \times 16$ Tokens ($\downarrow$ 90.5\%)}} \\
\midrule
\rowcolor[rgb]{.90, .96, .90}
FastV (ECCV24) & 52.8 & 46.6 & 48.8 & 44.7 & 50.0 & 55.0 & 50.0 & 45.0 & 49.8 & 78.5 \\
\rowcolor[rgb]{.84, .92, .86}
SparseVLM (ICML25) & 52.0 & 47.6 & 53.0 & 42.8 & 49.8 & 53.8 & 49.3 & 46.3 & 49.8 & 78.5 \\
\rowcolor[rgb]{.80, .88, .95}
CDPruner (NeurIPS25) & 58.9 & 52.7 & 57.4 & 48.5 & \second{57.3} & 66.2 & \second{56.0} & \second{49.6} & 56.3 & 88.8 \\
\rowcolor[rgb]{.78, .87, .95}
DivPrune (CVPR25) & 58.6 & 52.1 & 57.6 & 47.2 & 56.7 & 67.7 & 54.2 & 48.2 & 55.8 & 88.0 \\
\rowcolor[rgb]{.87, .94, .88}
VScan (TMLR26) & \second{59.2} & \second{54.8} & \second{60.6} & \second{49.7} & 57.1 & \second{68.0} & 54.8 & 48.4 & \second{57.0} & \second{89.9} \\
\rowcolor[rgb]{1.0, .95, .88}
\textbf{STAR-Pro (Ours)} & \best{60.3} & \best{56.1} & \best{62.4} & \best{50.6} & \best{60.0} & \best{71.0} & \best{57.6} & \best{51.6} & \best{58.8} & \best{92.7} \\
\bottomrule
\end{tabular}%
}
\normalsize
\caption{Full LLaVA-Video-7B results with 64 frames per video. Acc. uses the same three benchmark-level scores as the main paper (MLVU m-avg, LongVideoBench val, and Video-MME w/o subtitle); Rel. (\%) normalizes Acc. by the corresponding unpruned mean (63.4). The remaining columns provide the associated sub-metrics.}
\label{tab:video_comparison_full}
\end{table*}


\paragraph{Aggregate video trend}
At $64\times32$ retained tokens, STAR-Pro preserves 96.7\% of baseline
performance, 1.5 percentage points above VScan.  When the budget is halved to
$64\times16$, STAR-Pro retains 92.7\% and its margin over the runner-up widens
to 2.8 points.  The aggregate uses only MLVU m-avg, LongVideoBench val, and
Video-MME without subtitles; the additional columns in
\cref{tab:video_comparison_full} reveal whether this ranking is shared by the
underlying perception, relation, and duration-specific subsets.

\paragraph{MLVU}
At 32 tokens per frame, STAR-Pro obtains 65.4 m-avg, 2.4 points above CDPruner
and 3.0 points above VScan.  At 16 tokens per frame it remains first at 60.3,
with VScan second at 59.2.  The gap narrows under the tighter budget, but the
ranking is unchanged: STAR-Pro is the strongest method on MLVU at both
compression levels, contributing a consistent rather than budget-specific
advantage to the aggregate.

\paragraph{LongVideoBench}
The 32-token result exposes a useful internal trade-off.  STAR-Pro is best on
the perception subset (63.7) but lower on relation reasoning (51.1 versus
VScan's 53.1), leaving its validation score second at 57.0.  At 16 tokens per
frame, this relation gap reverses: STAR-Pro reaches 50.6, 0.9 points above
VScan, while its perception score of 62.4 is 1.8 points above the runner-up.
Consequently, STAR-Pro leads the validation score at 56.1, a 1.3-point margin
over VScan.  This observed reversal is consistent with progressive refinement
being more valuable under aggressive compression, when both perceptual
evidence and cross-frame relations must survive a smaller token budget;
because the compared methods differ in other selection choices, the table does
not isolate that mechanism causally.

\paragraph{Video-MME}
At 32 tokens per frame, STAR-Pro leads the no-subtitle score (61.6) and the
short- and medium-duration subsets (74.4 and 59.1).  Its long-duration score is
51.1, second to VScan's 51.7, showing that the moderate-compression gain is not
uniform across temporal length.  At 16 tokens per frame, STAR-Pro leads all
four Video-MME columns: 60.0 without subtitles, 71.0 on short videos, 57.6 on
medium videos, and 51.6 on long videos.  Relative to the strongest competing
value in each column, these are margins of 2.7, 3.0, 1.6, and 2.0 points,
respectively.  The long-duration score even increases from the 32-token result,
whereas the other Video-MME subsets decline moderately, illustrating that
budget sensitivity is not identical across temporal regimes.

\paragraph{Cross-benchmark takeaway}
At 32 tokens per frame, STAR-Pro's overall lead comes from first-place MLVU and
Video-MME together with second-place LongVideoBench; the detailed columns
localize the remaining weakness to relation and long-duration evaluation.  At
16 tokens per frame, STAR-Pro ranks first on all eight reported benchmark and
subtask columns.  The widening aggregate margin is therefore corroborated by
the full breakdown rather than being caused by one favorable video benchmark.

\subsection{Efficiency Protocol and Adaptive-Stage Overhead}
\label{sec:suppl_adaptive_overhead}

\noindent\textbf{Operator-counted TFLOPs.}
Operator-counted TFLOPs are averaged over three profiled calls after one
warm-up call and cover PyTorch matrix-multiplication and 2D-convolution
operators. Fused attention kernels without an exposed matrix multiplication
and selection operators such as top-$k$ and QR are not counted, so these values
are used only for within-protocol comparisons.

\noindent\textbf{Grouped complexity.}
The implementation does not factorize all visual tokens as one monolithic
matrix.  For groups indexed by $g$, with $n_g$ candidate tokens, $m_g$
retained pivots, and selector feature dimension $d_s$, the dominant pivoted-QR
work scales as
$\mathcal{O}\!\left(d_s\sum_g n_gm_g\right)$.
LLaVA-NeXT applies the selector crop-wise (five groups with $n_g{=}576$ and
$d_s{=}4096$ for the profiled five-crop inputs), whereas LLaVA-Video applies
it frame-wise.  At the profiled 64-frame, $T{=}16$ point, the latter consists
of 64 independent $169{\rightarrow}32$ selector calls with $d_s{=}3584$,
rather than one factorization over the complete video sequence.

\noindent\textbf{Measured exclusive timing.}
The exclusive-timing protocol compares the full-token and pruned pipelines
using the same LLaVA-Video-7B model, 64 sampled frames, VideoMME inputs, one
A800-80GB GPU, batch size 1, 10 warm-up samples, and 50 measured samples.
Timings use CUDA events and one synchronization after
\texttt{model.generate}; CPU video decoding and preprocessing are excluded.
Importantly, \emph{LLM prefill} is timed at the decoder core after multimodal
assembly and therefore sees the actual post-selection sequence: the visual
span decreases from 10,816 tokens for Full to 2,048 tokens for STAR-Pro.
Residual is the synchronized end-to-end latency not covered by the named CUDA
events.

The 872.9\,ms reduction in LLM prefill is substantially larger than the
40.6\,ms one-time selection cost: QR plus the remaining Adaptive Stage
accounts for only 4.6\% of the saved prefill time.  This yields a measured
$2.24{\times}$ end-to-end speedup while leaving the vision encoder cost
essentially unchanged.  Because the efficiency protocol generates one token,
the first output token is produced by the prefill forward pass and no
subsequent cached-decoding call occurs.  In a separate 16-token diagnostic
run at $T{=}64$, the 15 cached-decoding steps take 47.2\,ms in total; the
Adaptive Stage and QR are still executed only once.

\section{Extended Ablation Studies}
\label{sec:suppl_ablation}

\subsection{Pruning-Schedule Robustness at a Fixed Average-Token Budget}
\label{sec:suppl_schedule_robustness}

\noindent\textbf{Controlled schedule sweep.}
This study asks whether STAR-Pro depends on a narrowly tuned placement of its
two Progressive-Stage pruning layers.  We use LLaVA-1.5-7B at the
layer-average budget $T{=}64$ and fix the Adaptive-Stage multiplier at
$\alpha{=}2$.  The first pruning layer ranges over
$\ell_1\in\{6,\ldots,15\}$ and the second over
$\ell_2\in\{18,\ldots,27\}$, producing a complete $10\times10$ grid of 100
measured schedules.  For every pair, the integer survivor counts after the two
pruning layers are recomputed to preserve the same layer-average visual-token
budget.  This matching is important: moving a pruning layer later would otherwise
retain more tokens for more decoder layers and receive an unintended compute
advantage.  Performance is summarized by Mean-4, the average of POPE, GQA,
TextVQA, and MME$/20$.

\begin{figure*}[!t]
\centering
\begin{minipage}[b]{0.49\textwidth}\centering
\includegraphics[width=\linewidth]{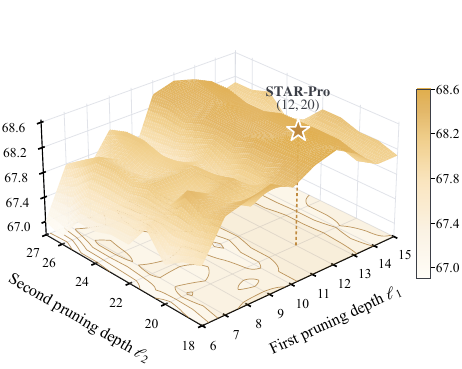}
\end{minipage}\hfill
\begin{minipage}[b]{0.49\textwidth}\centering
\includegraphics[width=\linewidth]{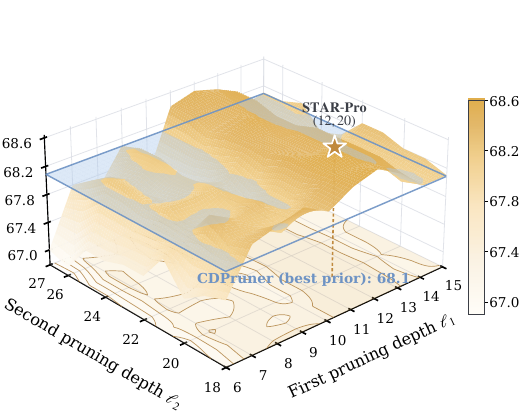}
\end{minipage}

\vspace{3pt}
\includegraphics[width=0.94\textwidth]{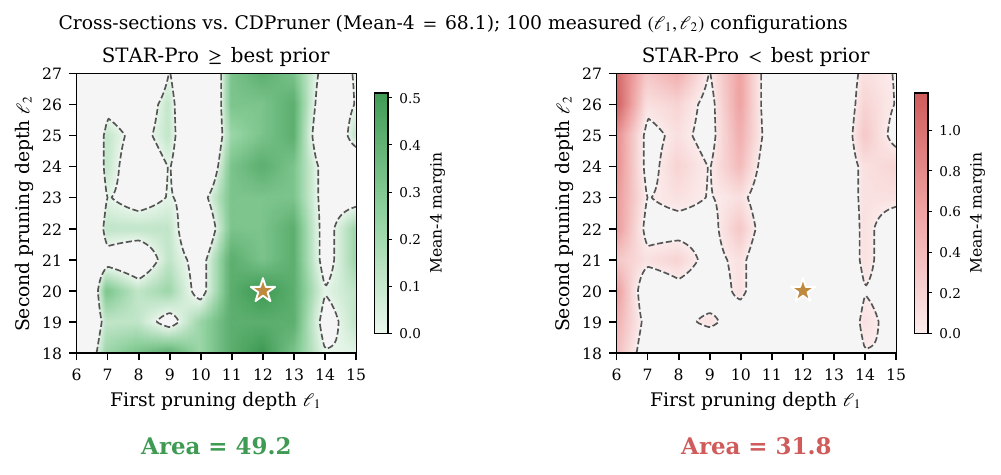}
\caption{\textbf{Pruning-schedule robustness on LLaVA-1.5-7B at $T{=}64$.}
\textbf{(a)} Mean-4 over the 100 measured pairs, shown directly
(left) and with the CDPruner Mean-4 reference plane at $68.1$ (right); the star
marks the deployed $(12,20)$ schedule.  \textbf{(b)} Cross-sections of the same
surface, with regions at or above the reference shown in green and regions
below it shown in red.}
\label{fig:suppl_layer_depth}
\end{figure*}

\noindent\textbf{Reading the performance surfaces.}
The upper-left panel of \cref{fig:suppl_layer_depth} shows the Mean-4 landscape
over the two pruning-layer coordinates, with the star marking the deployed
$(12,20)$ schedule.  The upper-right panel overlays the strongest prior method
at the same token budget as a horizontal reference plane: CDPruner obtains
Mean-4 $68.1$.  This view converts the sweep from a search for a single maximum
into a robustness test: a useful schedule should remain competitive over a
neighborhood of nearby pruning-layer choices.  The deployed schedule reaches
Mean-4 $68.6$, ties the best measured score at the reported one-decimal
precision, and lies about $0.5$ points above the reference plane.

\noindent\textbf{Where the robustness comes from.}
The lower panel projects the same surface onto the
$(\ell_1,\ell_2)$ plane and separates schedules at or above CDPruner from those
below it.  The at-or-above region occupies area $49.2$ out of the total search
area $81$, versus $31.8$ below the reference; thus approximately 61\% of the domain is stronger than the prior-method
baseline.  More importantly, the competitive region is contiguous around the
deployed point rather than being formed by isolated peaks.  For each
$\ell_1\in\{11,12,13\}$, every tested
$\ell_2\in\{18,\ldots,27\}$ remains at or above the reference.  Once the first
pruning layer lies in this central band, the second pruning layer can therefore move
across a broad range without falling below the prior-method reference.

\noindent\textbf{Interpretation and scope.}
Together, the panels show that $(12,20)$ is a representative point inside a
broad high-performing region, not a lone favorable cell---which is precisely
what lets STAR-Pro deploy a single pruning-layer schedule without
benchmark-specific tuning.  We keep the quantitative claims grounded in the
$100$ measured configurations: both the best-score tie and the central-band
coverage are read directly off the measurements, while the
bilinear-interpolated surfaces serve only to render and partition that grid.
Accordingly, the evidence is deliberately scoped to LLaVA-1.5-7B at $T{=}64$
over the displayed search range: it establishes robustness to nearby
pruning-layer choices, not that $(12,20)$ is globally optimal across
architectures or token budgets.

\FloatBarrier

\subsection{Hyperparameter Analysis of the Adaptive and Progressive Stages}
\label{sec:suppl_hyperparameter_analysis}

We analyze three complementary choices under matched layer-average budgets:
the Adaptive-Stage candidate-pool multiplier, the candidate selection
operator, and the number of Progressive-Stage pruning layers.  We first
visualize sensitivity to the candidate-pool multiplier, and then compare the
selector and pruning-layer-count controls in a common per-benchmark table.

\label{sec:suppl_alpha_ablation}
\noindent\textbf{Candidate-pool multiplier.}
We sweep the Adaptive-Stage multiplier
$\alpha\in\{1,1.5,2,2.5,3\}$ while fixing the two Progressive-Stage
pruning layers at L12/L20.  Integer retention schedules are selected under the
same layer-average token budget used by the main method; the detailed table
reports the actual average whenever the strict constraint is infeasible.
Across all four models and three compression budgets, STAR-Pro with
$\alpha\in\{1.5,2,2.5\}$ matches or exceeds the strongest prior method at the
same budget in nearly every setting---and at the deployed $\alpha{=}2$ in
every setting.  Within this range, the aggregate Rel.\ spans at most $1.4$
points for every model and budget.  Performance generally peaks near
$\alpha{=}2$ and degrades for both an under-sized pool ($\alpha{=}1$) and an
over-sized one ($\alpha{=}3$), showing that the deployed value lies in a
stable neighborhood rather than at an isolated optimum.

\begin{figure*}[!t]
\centering
\includegraphics[width=0.88\textwidth]{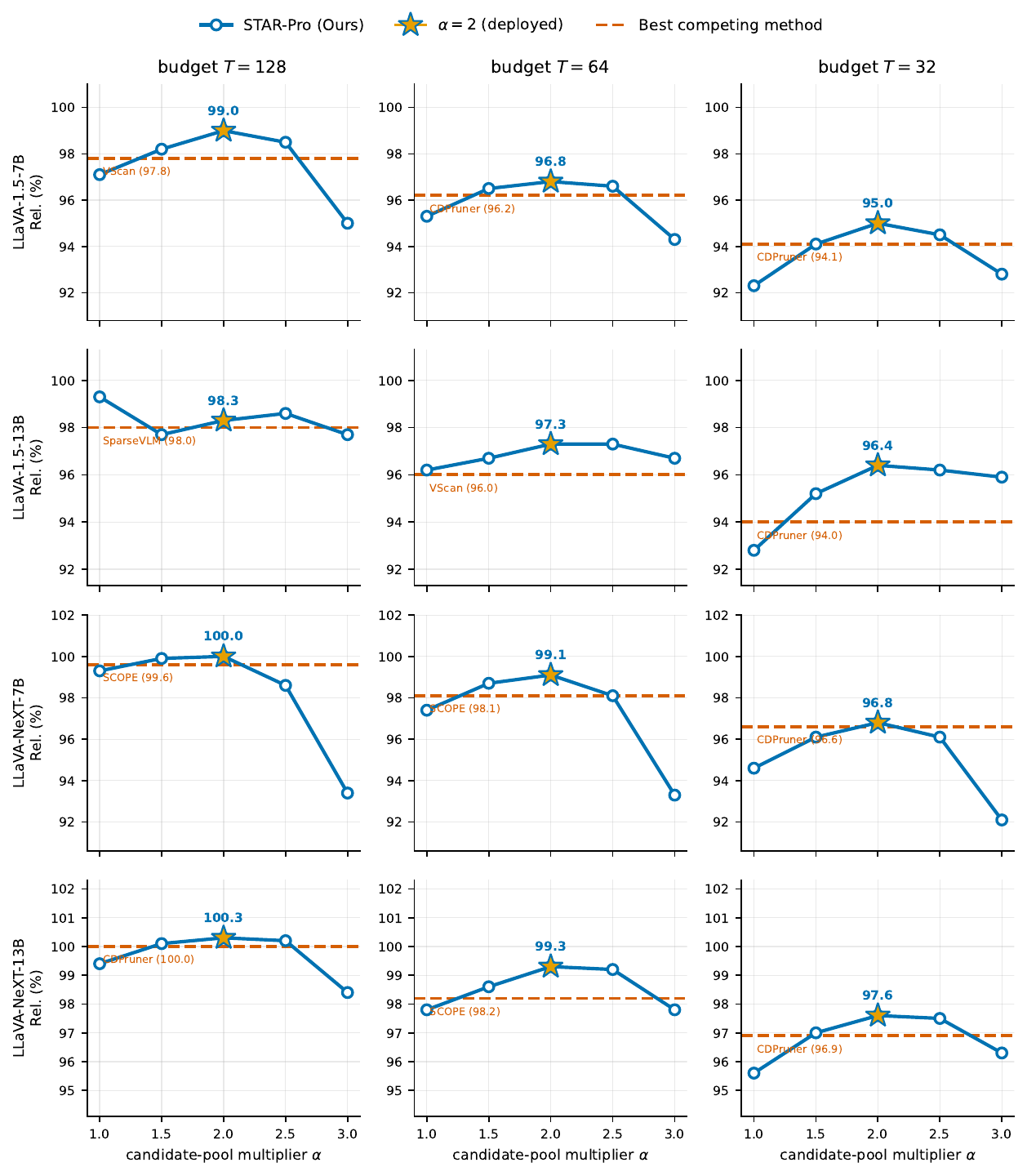}
\caption{\textbf{Robustness and stability to the candidate-pool multiplier $\alpha$.}
Each panel plots STAR-Pro's aggregate Rel.\ (\%) as $\alpha$ sweeps
$\{1,1.5,2,2.5,3\}$, for one model (rows) at one layer-averaged token budget
(columns). The star marks $\alpha{=}2$, the deployed configuration used throughout the
paper. All plotted points---including the $\alpha{=}2$ star---are the per-model
candidate-multiplier sweep values reported in
\cref{tab:alpha_sweep_llava15_7b,tab:alpha_sweep_llava15_13b,tab:alpha_sweep_llavanext7b,tab:alpha_sweep_llavanext13b},
where the $\alpha{=}2$ row is highlighted in each.
The dashed line is the best competing method at
the same budget in the main comparison tables. Rel.\ is the 10-benchmark Acc.\
(MME/20) normalized by each model's unpruned baseline; exact values are listed
in those tables.}
\label{fig:suppl_alpha_rel_grid}
\end{figure*}

\noindent\textbf{Deployed and diagnostic schedules.}
The deployed LLaVA-1.5-7B configuration uses pruning layers $(12,20)$, as listed
in \cref{tab:architecture_layer_selection}.  The full matched-compute schedule
sweep in the main paper evaluates all 100 measured pairs
$\ell_1\in\{6,\ldots,15\}$ and $\ell_2\in\{18,\ldots,27\}$ at $T{=}64$.
Its score is
$\operatorname{mean}(\mathrm{POPE},\mathrm{GQA},\mathrm{TextVQA},
\mathrm{MME}/20)$; the displayed surface rounds measured scores to one decimal
and uses bilinear interpolation only for visualization.  The deployed
$(12,20)$ pair ties the best measured score at that reporting precision.

\noindent\textbf{Adaptive-Stage selector control.}
Block~(a) of \cref{tab:ablation_isocompute} evaluates two alternative pools at
$\alpha{=}2$ under the same diagnostic Progressive-Stage schedule and
layer-average budget.  Random removes adaptive structure altogether, whereas
Stride uses a uniform spatial grid.  For context, both controls trail the
complete configuration on POPE, GQA, and TextVQA at every budget; MME is the
exception at $T{=}128$,
where Stride is higher, so the comparison should be read across metrics rather
than through that unnormalised score alone.

\noindent\textbf{Number of Progressive-Stage pruning layers.}
Block~(b) reports $P{=}0$, $P{=}3$, and $P{=}4$ controls.  The $P{=}0$ arm uses
only the Adaptive Stage and makes its terminal QR decision before the decoder.
For $P{=}3$ and $P{=}4$, the listed equal-depth placements and integer survivor
counts preserve the same layer-average token budget.  The highlighted Ours row
completes the comparison with $P{=}2$ at the deployed $(12,20)$ pruning layers.
Its Adaptive Stage uses STAR-Pro's query-agnostic pivoted-QR rule, selecting at
each step the token with the largest residual after projection onto the span
of the selected tokens.  It therefore targets complementary feature coverage
without a positive or negative query-relevance weight.

\begin{table*}[t]
\centering
\small
\resizebox{\textwidth}{!}{%
\begin{tabular}{ll|cccc|cccc|cccc}
\toprule
\multirow{2}{*}{\textbf{Arm}} & \multirow{2}{*}{\textbf{Configuration}}
  & \multicolumn{4}{c|}{$T{=}128$ ($\downarrow$77.8\%)}
  & \multicolumn{4}{c|}{$T{=}64$ ($\downarrow$88.9\%)}
  & \multicolumn{4}{c}{$T{=}32$ ($\downarrow$94.4\%)} \\
\cmidrule(lr){3-6}\cmidrule(lr){7-10}\cmidrule(lr){11-14}
 & & POPE & GQA & VQA$^{\text{T}}$ & MME
   & POPE & GQA & VQA$^{\text{T}}$ & MME
   & POPE & GQA & VQA$^{\text{T}}$ & MME \\
\midrule
\multicolumn{14}{l}{\textit{(a) Adaptive Stage selection operator ($\alpha{=}2$, fixed diagnostic Progressive schedule)}} \\
\midrule
Random         & no structure                     & 85.13 & 59.84 & 54.21 & 1447.6 & 82.88 & 57.97 & 51.43 & 1372.5 & 77.40 & 54.64 & 48.17 & 1265.2 \\
Stride         & uniform grid                     & 85.16 & 60.13 & 54.47 & 1471.3 & 82.94 & 57.73 & 51.08 & 1380.8 & 77.40 & 55.41 & 48.12 & 1221.2 \\
\midrule
\multicolumn{14}{l}{\textit{(b) Number of Progressive Stage pruning layers $P$ ($\alpha{=}2$)}} \\
\midrule
$P{=}0$ & Adaptive only                  & 87.21 & 59.68 & 56.79 & 1422.5 & 86.12 & 58.13 & 56.03 & 1384.7 & 83.94 & 55.72 & 54.28 & 1353.1 \\
$P{=}3$ & $\ell{=}(8,16,24)$        & 86.63 & 60.34 & 57.61 & 1441.8 & 86.76 & 58.90 & 56.94 & 1427.0 & 85.33 & 57.01 & 55.64 & 1374.3 \\
$P{=}4$ & $\ell{=}(6,13,19,26)$     & 86.63 & 60.51 & 57.54 & 1440.1 & 86.86 & 59.09 & 56.63 & 1428.2 & 85.45 & 57.66 & 55.52 & 1389.8 \\
\midrule
\multicolumn{14}{l}{\textit{Ours}} \\
\midrule
\rowcolor[rgb]{1.0, .92, .80} \textbf{STAR-Pro} & pivoted-QR coverage, $P{=}2$, $\ell{=}(12,20)$ & 87.30 & 60.80 & 57.20 & 1456.3 & 87.52 & 59.41 & 56.26 & 1421.4 & 86.60 & 57.20 & 54.60 & 1358.1 \\
\bottomrule
\end{tabular}%
}
\normalsize
\caption{Matched-compute controls on LLaVA-1.5-7B. Every arm has the nominal
layer-average visual-token budget $T$. Block~(a) reports Random and Stride
candidate pools under a fixed diagnostic Progressive schedule. Block~(b) varies the number of
Progressive-Stage pruning layers; $P{=}0$ denotes Adaptive-only selection, and
the listed $P\in\{3,4\}$ placements are diagnostic equal-depth schedules. The Ours
row uses query-agnostic pivoted-QR coverage with the deployed $P{=}2$,
$\ell{=}(12,20)$ schedule and reports the corresponding main-paper scores.
MME is unnormalised and should be read alongside POPE, GQA, and
VQA$^{\text{Text}}$.}
\label{tab:ablation_isocompute}
\end{table*}

\subsection{Robustness to the Per-Layer Token Allocation}
\label{sec:suppl_budget_ablation}
We probe whether STAR-Pro depends on a finely tuned per-layer token schedule. For every model we keep its \emph{deployed} pruning layers and the candidate multiplier $\alpha{=}2$ (pool $=2T$) unchanged, and only re-allocate how many tokens survive each of the two pruning layers: we make the split moderately more front-heavy than the deployed schedule (retained ratio $K_1{:}K_2\approx2.5{:}1$ vs.\ the deployed ${\approx}2{:}1$) while keeping the layer-average visual-token count \emph{exactly} at each budget. \emph{Depl.} is the deployed schedule (\cref{tab:architecture_layer_selection}); \emph{Token-var.} is this re-allocated schedule. The \emph{Config} column lists the surviving tokens $2T{\to}K_1{\to}K_2$ in $[0,\ell_1),[\ell_1,\ell_2),[\ell_2,L)$.

Across all twelve model--budget pairs, reallocating tokens changes Rel.\ by
only $-0.3$ to $+0.5$ percentage points relative to the deployed schedule.
Nine variants improve, one is unchanged, and the two decreases are at most
$0.3$ points. This narrow range holds across all four architectures and all
three tested budgets, indicating that the measured performance is robust to a
moderate redistribution of survivors rather than tied to one exact split.
\begin{table*}[t]
\centering
\small
\setlength{\tabcolsep}{3.2pt}
\resizebox{\textwidth}{!}{%
\begin{tabular}{llcccccccccccc}
\toprule
\textbf{Schedule} & \textbf{Config ($2T{\to}K_1{\to}K_2$)} & \textbf{SEED} & \textbf{GQA} & \textbf{SQA}$^{\text{IMG}}$ & \textbf{VQA}$^{\text{Text}}$ & \textbf{POPE} & \textbf{MME} & \textbf{MMB}$^{\text{EN}}$ & \textbf{MMB}$^{\text{CN}}$ & \textbf{AI2D} & \textbf{MMMU} & \textbf{Acc.} & \textbf{Rel.} \\
\midrule
\multicolumn{14}{c}{\textbf{LLaVA-1.5-7B} \;---\; deployed pruning layers $\ell_{12},\ell_{20}$, $\alpha{=}2$} \\
\midrule
\multicolumn{14}{c}{\textit{Layer-average budget $T{=}128$ ($\downarrow$ 77.8\%)}} \\
\rowcolor[rgb]{.93,.93,.93} Depl. & $256{\to}74{\to}36$ & 64.4 & 60.8 & 68.9 & 57.2 & 87.3 & 1456.3 & 62.4 & 57.7 & 55.0 & 36.7 & 62.32 & 98.9 \\
\rowcolor[rgb]{1.0,.92,.80} Token-var. & $256{\to}80{\to}32$ & 64.4 & 60.8 & 68.7 & 57.4 & 87.2 & 1464.3 & 62.7 & 57.7 & 55.1 & 36.8 & 62.41 & 99.0 \\
\cmidrule(lr){1-14}
\multicolumn{14}{c}{\textit{Layer-average budget $T{=}64$ ($\downarrow$ 88.9\%)}} \\
\rowcolor[rgb]{.93,.93,.93} Depl. & $128{\to}37{\to}18$ & 62.87 & 59.41 & 68.62 & 56.26 & 87.52 & 1421.4 & 60.97 & 56.44 & 53.69 & 35.11 & 61.20 & 97.1 \\
\rowcolor[rgb]{1.0,.92,.80} Token-var. & $128{\to}40{\to}16$ & 62.9 & 59.3 & 68.6 & 56.3 & 87.4 & 1383.5 & 61.1 & 56.4 & 53.7 & 34.9 & 60.97 & 96.8 \\
\cmidrule(lr){1-14}
\multicolumn{14}{c}{\textit{Layer-average budget $T{=}32$ ($\downarrow$ 94.4\%)}} \\
\rowcolor[rgb]{.93,.93,.93} Depl. & $64{\to}17{\to}10$ & 60.3 & 57.2 & 68.7 & 54.6 & 86.6 & 1358.1 & 60.1 & 55.1 & 53.4 & 34.4 & 59.83 & 94.9 \\
\rowcolor[rgb]{1.0,.92,.80} Token-var. & $64{\to}20{\to}8$ & 60.4 & 57.4 & 68.6 & 54.9 & 86.5 & 1370.4 & 59.9 & 55.0 & 53.0 & 34.7 & 59.89 & 95.0 \\
\midrule
\multicolumn{14}{c}{\textbf{LLaVA-1.5-13B} \;---\; deployed pruning layers $\ell_{15},\ell_{30}$, $\alpha{=}2$} \\
\midrule
\multicolumn{14}{c}{\textit{Layer-average budget $T{=}128$ ($\downarrow$ 77.8\%)}} \\
\rowcolor[rgb]{.93,.93,.93} Depl. & $256{\to}64{\to}32$ & 66.1 & 60.2 & 72.7 & 59.5 & 87.2 & 1492.7 & 67.0 & 63.1 & 57.9 & 35.4 & 64.37 & 97.9 \\
\rowcolor[rgb]{1.0,.92,.80} Token-var. & $256{\to}68{\to}26$ & 66.1 & 60.4 & 72.7 & 59.4 & 87.1 & 1499.7 & 67.0 & 63.2 & 57.9 & 35.4 & 64.43 & 98.0 \\
\cmidrule(lr){1-14}
\multicolumn{14}{c}{\textit{Layer-average budget $T{=}64$ ($\downarrow$ 88.9\%)}} \\
\rowcolor[rgb]{.93,.93,.93} Depl. & $128{\to}32{\to}16$ & 64.3 & 59.4 & 73.1 & 58.6 & 86.8 & 1456.6 & 66.2 & 62.0 & 58.1 & 36.2 & 63.75 & 97.0 \\
\rowcolor[rgb]{1.0,.92,.80} Token-var. & $128{\to}34{\to}13$ & 64.3 & 59.3 & 73.1 & 58.6 & 86.8 & 1461.0 & 66.3 & 62.0 & 58.2 & 36.1 & 63.78 & 97.0 \\
\cmidrule(lr){1-14}
\multicolumn{14}{c}{\textit{Layer-average budget $T{=}32$ ($\downarrow$ 94.4\%)}} \\
\rowcolor[rgb]{.93,.93,.93} Depl. & $64{\to}16{\to}8$ & 62.9 & 58.0 & 73.1 & 57.4 & 85.7 & 1464.1 & 65.6 & 61.0 & 57.8 & 36.9 & 63.16 & 96.1 \\
\rowcolor[rgb]{1.0,.92,.80} Token-var. & $64{\to}18{\to}5$ & 62.8 & 58.2 & 73.1 & 57.5 & 85.7 & 1471.2 & 65.6 & 61.1 & 57.7 & 38.4 & 63.37 & 96.4 \\
\midrule
\multicolumn{14}{c}{\textbf{LLaVA-NeXT-7B} \;---\; deployed pruning layers $\ell_{12},\ell_{20}$, $\alpha{=}2$} \\
\midrule
\multicolumn{14}{c}{\textit{Layer-average budget $T{=}128$ ($\downarrow$ 77.8\%)}} \\
\rowcolor[rgb]{.93,.93,.93} Depl. & $1280{\to}367{\to}182$ & 68.95 & 62.4 & 68.8 & 59.1 & 87.6 & 1479.1 & 66.0 & 58.0 & 65.6 & 32.2 & 64.26 & 99.6 \\
\rowcolor[rgb]{1.0,.92,.80} Token-var. & $1280{\to}400{\to}160$ & 69.0 & 62.5 & 68.7 & 59.3 & 87.5 & 1479.1 & 65.8 & 58.2 & 65.7 & 34.2 & 64.50 & 99.9 \\
\cmidrule(lr){1-14}
\multicolumn{14}{c}{\textit{Layer-average budget $T{=}64$ ($\downarrow$ 88.9\%)}} \\
\rowcolor[rgb]{.93,.93,.93} Depl. & $640{\to}182{\to}92$ & 67.4 & 61.8 & 67.9 & 58.1 & 87.5 & 1468.1 & 65.4 & 58.3 & 65.3 & 33.8 & 63.89 & 99.0 \\
\rowcolor[rgb]{1.0,.92,.80} Token-var. & $640{\to}200{\to}80$ & 67.4 & 61.7 & 67.9 & 58.2 & 87.5 & 1464.1 & 65.4 & 58.3 & 65.2 & 35.3 & 64.02 & 99.2 \\
\cmidrule(lr){1-14}
\multicolumn{14}{c}{\textit{Layer-average budget $T{=}32$ ($\downarrow$ 94.4\%)}} \\
\rowcolor[rgb]{.93,.93,.93} Depl. & $320{\to}91{\to}46$ & 65.21 & 60.4 & 67.7 & 56.4 & 86.2 & 1421.9 & 64.4 & 56.9 & 63.3 & 34.3 & 62.59 & 97.0 \\
\rowcolor[rgb]{1.0,.92,.80} Token-var. & $320{\to}100{\to}40$ & 65.3 & 60.2 & 67.4 & 56.9 & 86.1 & 1406.4 & 64.3 & 57.0 & 63.4 & 34.3 & 62.53 & 96.9 \\
\midrule
\multicolumn{14}{c}{\textbf{LLaVA-NeXT-13B} \;---\; deployed pruning layers $\ell_{15},\ell_{30}$, $\alpha{=}2$} \\
\midrule
\multicolumn{14}{c}{\textit{Layer-average budget $T{=}128$ ($\downarrow$ 77.8\%)}} \\
\rowcolor[rgb]{.93,.93,.93} Depl. & $1280{\to}320{\to}160$ & 71.33 & 64.2 & 72.8 & 62.2 & 87.4 & 1545.2 & 68.5 & 63.1 & 69.8 & 35.8 & 67.2 & 100.3 \\
\rowcolor[rgb]{1.0,.92,.80} Token-var. & $1280{\to}336{\to}136$ & 71.3 & 64.1 & 73.0 & 62.0 & 87.5 & 1544.9 & 68.4 & 63.1 & 69.9 & 37.1 & 67.37 & 100.5 \\
\cmidrule(lr){1-14}
\multicolumn{14}{c}{\textit{Layer-average budget $T{=}64$ ($\downarrow$ 88.9\%)}} \\
\rowcolor[rgb]{.93,.93,.93} Depl. & $640{\to}160{\to}80$ & 69.19 & 63.4 & 72.0 & 60.6 & 87.2 & 1524.2 & 67.3 & 63.7 & 68.2 & 35.2 & 66.3 & 98.9 \\
\rowcolor[rgb]{1.0,.92,.80} Token-var. & $640{\to}168{\to}68$ & 69.2 & 63.3 & 72.0 & 60.7 & 87.1 & 1522.1 & 67.4 & 63.7 & 68.4 & 38.0 & 66.60 & 99.4 \\
\cmidrule(lr){1-14}
\multicolumn{14}{c}{\textit{Layer-average budget $T{=}32$ ($\downarrow$ 94.4\%)}} \\
\rowcolor[rgb]{.93,.93,.93} Depl. & $320{\to}80{\to}40$ & 66.94 & 62.3 & 71.4 & 59.3 & 86.0 & 1519.8 & 65.3 & 62.7 & 68.0 & 35.4 & 65.3 & 97.5 \\
\rowcolor[rgb]{1.0,.92,.80} Token-var. & $320{\to}84{\to}34$ & 67.0 & 62.6 & 71.4 & 59.2 & 86.1 & 1518.3 & 65.1 & 62.7 & 68.1 & 37.1 & 65.52 & 97.8 \\
\bottomrule
\end{tabular}%
}
\normalsize
\caption{\textbf{Robustness to the per-layer token allocation.} For each model the pruning layers and $\alpha{=}2$ are fixed at their deployed values; only the per-layer retained counts change. \emph{Depl.}\ is the deployed schedule and \emph{Token-var.}\ a moderately more front-heavy re-allocation ($K_1{:}K_2\approx2.5{:}1$) with the \emph{same} layers and the \emph{same} exact layer-average budget. The \emph{Config} column gives $2T{\to}K_1{\to}K_2$. Acc.\ is the average over the 10 benchmarks with MME divided by 20; Rel.\ (\%) is Acc.\ relative to the corresponding unpruned baseline.}
\label{tab:suppl_budget_ablation}
\end{table*}

\begin{table*}[t]
\centering
\small
\setlength{\tabcolsep}{4pt}
\resizebox{\textwidth}{!}{%
\begin{tabular}{lrrrrrrrrrrrrr}
\toprule
\textbf{Config.} & $T_{\text{avg}}$ & \textbf{SEED} & \textbf{GQA} & \textbf{SQA}$^{\text{IMG}}$ & \textbf{VQA}$^{\text{Text}}$ & \textbf{POPE} & \textbf{MME} & \textbf{MMB}$^{\text{EN}}$ & \textbf{MMB}$^{\text{CN}}$ & \textbf{AI2D} & \textbf{MMMU} & \textbf{Acc.} & \textbf{Rel.} \\
\midrule
\multicolumn{14}{c}{\textit{Layer-average budget $T{=}128$ ($\downarrow$ 77.8\%)}} \\
\midrule
$\alpha{=}1$   & 128 & 62.99 & 59.68 & 68.67 & 56.79 & 87.21 & 1388.6 & 61.31 & 56.19 & 54.05 & 35.44 & 61.18 & 97.1 \\
$\alpha{=}1.5$ & 128 & 63.87 & 60.10 & 68.62 & 57.57 & 87.47 & 1421.7 & 62.16 & 57.39 & 54.66 & 36.00 & 61.89 & 98.2 \\
\rowcolor[rgb]{1.0, .92, .80} \textbf{STAR-Pro ($\alpha{=}2$)} & 128 & 64.4 & 60.8 & 68.9 & 57.2 & 87.3 & 1456.3 & 62.4 & 57.7 & 55.0 & 36.7 & 62.32 & 98.9 \\
$\alpha{=}2.5$ & 128 & 65.07 & 58.31 & 68.47 & 54.95 & 87.06 & 1480.1 & 63.52 & 58.33 & 54.79 & 36.33 & 62.08 & 98.5 \\
$\alpha{=}3$   & 144 & 64.10 & 50.99 & 68.52 & 50.61 & 80.00 & 1469.4 & 62.84 & 57.13 & 55.21 & 36.00 & 59.89 & 95.0 \\
\midrule
\multicolumn{14}{c}{\textit{Layer-average budget $T{=}64$ ($\downarrow$ 88.9\%)}} \\
\midrule
$\alpha{=}1$   & 64 & 60.47 & 58.13 & 68.57 & 56.03 & 86.12 & 1368.5 & 60.03 & 54.98 & 53.17 & 34.44 & 60.04 & 95.3 \\
$\alpha{=}1.5$ & 64 & 62.18 & 59.12 & 68.96 & 56.46 & 87.04 & 1367.1 & 60.97 & 55.84 & 54.08 & 35.00 & 60.80 & 96.5 \\
\rowcolor[rgb]{1.0, .92, .80} \textbf{STAR-Pro ($\alpha{=}2$)} & 64 & 62.87 & 59.41 & 68.62 & 56.26 & 87.52 & 1421.4 & 60.97 & 56.44 & 53.69 & 35.11 & 61.20 & 97.1 \\
$\alpha{=}2.5$ & 64 & 63.11 & 57.00 & 68.47 & 54.03 & 87.37 & 1395.6 & 61.39 & 57.73 & 54.24 & 35.44 & 60.86 & 96.6 \\
$\alpha{=}3$   & 72 & 62.50 & 50.12 & 68.96 & 50.20 & 83.34 & 1439.2 & 60.88 & 56.62 & 54.34 & 35.11 & 59.40 & 94.3 \\
\midrule
\multicolumn{14}{c}{\textit{Layer-average budget $T{=}32$ ($\downarrow$ 94.4\%)}} \\
\midrule
$\alpha{=}1$   & 32 & 57.74 & 55.72 & 68.27 & 54.28 & 83.94 & 1328.7 & 58.33 & 51.46 & 52.66 & 33.00 & 58.18 & 92.3 \\
$\alpha{=}1.5$ & 32 & 59.55 & 57.04 & 68.17 & 54.91 & 86.03 & 1348.7 & 59.78 & 53.69 & 52.62 & 34.00 & 59.32 & 94.1 \\
\rowcolor[rgb]{1.0, .92, .80} \textbf{STAR-Pro ($\alpha{=}2$)} & 32 & 60.3 & 57.2 & 68.7 & 54.6 & 86.6 & 1358.1 & 60.1 & 55.1 & 53.4 & 34.4 & 59.83 & 94.9 \\
$\alpha{=}2.5$ & 32 & 61.15 & 54.54 & 68.86 & 53.35 & 86.73 & 1338.3 & 60.46 & 55.58 & 53.08 & 35.11 & 59.58 & 94.5 \\
$\alpha{=}3$   & 36 & 60.77 & 49.43 & 68.77 & 49.86 & 83.22 & 1377.9 & 60.20 & 55.84 & 53.47 & 34.11 & 58.46 & 92.8 \\
\bottomrule
\end{tabular}}
\caption{Candidate-multiplier sweep on \textbf{LLaVA-1.5-7B} with the two
Progressive-Stage checkpoints fixed at $\{12,20\}$. The highlighted
\textbf{STAR-Pro ($\alpha{=}2$)} row is the deployed configuration used
throughout the paper; its scores are taken from the main comparison table
(\cref{tab:pruning_comparison}). $T_{\text{avg}}$ is the realized layer-average
visual-token count (it exceeds $T$ only where the strict average constraint is
infeasible, at $\alpha{=}3$). Acc.\ is the 10-benchmark average with MME/20,
and Rel.\ (\%) normalizes Acc.\ by the full-token baseline.}
\label{tab:alpha_sweep_llava15_7b}
\end{table*}

\begin{table*}[t]
\centering
\small
\setlength{\tabcolsep}{4pt}
\resizebox{\textwidth}{!}{%
\begin{tabular}{lrrrrrrrrrrrrr}
\toprule
\textbf{Config.} & $T_{\text{avg}}$ & \textbf{SEED} & \textbf{GQA} & \textbf{SQA}$^{\text{IMG}}$ & \textbf{VQA}$^{\text{Text}}$ & \textbf{POPE} & \textbf{MME} & \textbf{MMB}$^{\text{EN}}$ & \textbf{MMB}$^{\text{CN}}$ & \textbf{AI2D} & \textbf{MMMU} & \textbf{Acc.} & \textbf{Rel.} \\
\midrule
\multicolumn{14}{c}{\textit{Layer-average budget $T{=}128$ ($\downarrow$ 77.8\%)}} \\
\midrule
$\alpha{=}1$   & 128 & 64.34 & 59.21 & 73.08 & 58.65 & 86.83 & 1761.8 & 66.33 & 62.03 & 58.00 & 36.44 & 65.30 & 99.3 \\
$\alpha{=}1.5$ & 128 & 65.50 & 59.93 & 73.08 & 59.53 & 87.01 & 1493.7 & 66.24 & 62.11 & 57.71 & 36.11 & 64.19 & 97.7 \\
\rowcolor[rgb]{1.0, .92, .80} \textbf{STAR-Pro ($\alpha{=}2$)} & 128 & 66.1 & 60.2 & 72.7 & 59.5 & 87.2 & 1492.7 & 67.0 & 63.1 & 57.9 & 35.4 & 64.37 & 97.9 \\
$\alpha{=}2.5$ & 128 & 66.97 & 60.88 & 73.23 & 59.87 & 86.94 & 1493.4 & 67.35 & 62.29 & 59.49 & 36.33 & 64.80 & 98.6 \\
$\alpha{=}3$   & 128 & 67.22 & 58.73 & 73.33 & 57.48 & 86.30 & 1488.1 & 67.09 & 62.29 & 59.29 & 36.33 & 64.25 & 97.7 \\
\midrule
\multicolumn{14}{c}{\textit{Layer-average budget $T{=}64$ ($\downarrow$ 88.9\%)}} \\
\midrule
$\alpha{=}1$   & 64 & 62.76 & 58.32 & 73.18 & 57.88 & 85.79 & 1466.8 & 65.39 & 60.91 & 57.74 & 37.33 & 63.26 & 96.2 \\
$\alpha{=}1.5$ & 64 & 63.88 & 59.18 & 72.24 & 58.30 & 86.58 & 1493.3 & 65.56 & 61.51 & 57.35 & 36.67 & 63.59 & 96.7 \\
\rowcolor[rgb]{1.0, .92, .80} \textbf{STAR-Pro ($\alpha{=}2$)} & 64 & 64.3 & 59.4 & 73.1 & 58.6 & 86.8 & 1456.6 & 66.2 & 62.0 & 58.1 & 36.2 & 63.75 & 97.0 \\
$\alpha{=}2.5$ & 64 & 64.79 & 58.91 & 72.63 & 58.89 & 86.56 & 1509.2 & 66.75 & 61.94 & 57.51 & 36.11 & 63.96 & 97.3 \\
$\alpha{=}3$   & 64 & 65.38 & 57.74 & 72.58 & 56.61 & 86.62 & 1498.7 & 66.33 & 61.51 & 58.32 & 35.67 & 63.57 & 96.7 \\
\midrule
\multicolumn{14}{c}{\textit{Layer-average budget $T{=}32$ ($\downarrow$ 94.4\%)}} \\
\midrule
$\alpha{=}1$   & 32 & 60.25 & 56.94 & 72.68 & 56.31 & 82.46 & 1417.2 & 61.82 & 57.47 & 56.61 & 34.67 & 61.01 & 92.8 \\
$\alpha{=}1.5$ & 32 & 62.47 & 58.43 & 72.83 & 57.73 & 84.93 & 1464.4 & 63.86 & 59.36 & 56.77 & 36.00 & 62.56 & 95.2 \\
\rowcolor[rgb]{1.0, .92, .80} \textbf{STAR-Pro ($\alpha{=}2$)} & 32 & 62.9 & 58.0 & 73.1 & 57.4 & 85.7 & 1464.1 & 65.6 & 61.0 & 57.8 & 36.9 & 63.16 & 96.1 \\
$\alpha{=}2.5$ & 32 & 63.47 & 57.97 & 71.99 & 57.78 & 85.78 & 1463.1 & 65.99 & 61.77 & 58.26 & 36.22 & 63.24 & 96.2 \\
$\alpha{=}3$   & 32 & 63.61 & 56.38 & 72.04 & 55.80 & 86.03 & 1510.1 & 65.82 & 61.43 & 57.64 & 35.78 & 63.00 & 95.9 \\
\bottomrule
\end{tabular}}
\caption{Candidate-multiplier sweep on \textbf{LLaVA-1.5-13B} with the two
Progressive-Stage checkpoints fixed at $\{15,30\}$. The highlighted
\textbf{STAR-Pro ($\alpha{=}2$)} row is the deployed configuration; its scores
are taken from the main comparison table (\cref{tab:pruning_comparison_13b}).
$T_{\text{avg}}$ is the realized layer-average visual-token count. Acc.\ is the
10-benchmark average with MME/20, and Rel.\ (\%) normalizes Acc.\ by the
full-token baseline.}
\label{tab:alpha_sweep_llava15_13b}
\end{table*}

\begin{table*}[t]
\centering
\small
\setlength{\tabcolsep}{4pt}
\resizebox{\textwidth}{!}{%
\begin{tabular}{lrrrrrrrrrrrrr}
\toprule
\textbf{Config.} & $T_{\text{avg}}$ & \textbf{SEED} & \textbf{GQA} & \textbf{SQA}$^{\text{IMG}}$ & \textbf{VQA}$^{\text{Text}}$ & \textbf{POPE} & \textbf{MME} & \textbf{MMB}$^{\text{EN}}$ & \textbf{MMB}$^{\text{CN}}$ & \textbf{AI2D} & \textbf{MMMU} & \textbf{Acc.} & \textbf{Rel.} \\
\midrule
\multicolumn{14}{c}{\textit{Layer-average budget $T{=}640$ ($\downarrow$ 77.8\%)}} \\
\midrule
$\alpha{=}1$   & 640 & 67.66 & 61.97 & 67.97 & 58.19 & 87.31 & 1472.0 & 65.48 & 57.90 & 65.22 & 35.33 & 64.06 & 99.3 \\
$\alpha{=}1.5$ & 640 & 68.53 & 62.62 & 68.37 & 59.51 & 87.56 & 1491.5 & 65.31 & 57.99 & 65.09 & 34.89 & 64.44 & 99.9 \\
\rowcolor[rgb]{1.0, .92, .80} \textbf{STAR-Pro ($\alpha{=}2$)} & 640 & 68.95 & 62.4 & 68.8 & 59.1 & 87.6 & 1479.1 & 66.0 & 58.0 & 65.6 & 32.2 & 64.26 & 99.6 \\
$\alpha{=}2.5$ & 640 & 69.06 & 59.30 & 69.01 & 53.40 & 87.95 & 1488.2 & 65.65 & 58.08 & 64.93 & 34.33 & 63.61 & 98.6 \\
$\alpha{=}3$   & 720 & 67.27 & 49.05 & 68.72 & 43.81 & 81.91 & 1471.8 & 63.69 & 56.01 & 64.35 & 34.33 & 60.27 & 93.4 \\
\midrule
\multicolumn{14}{c}{\textit{Layer-average budget $T{=}320$ ($\downarrow$ 88.9\%)}} \\
\midrule
$\alpha{=}1$   & 320 & 65.45 & 60.88 & 67.67 & 58.00 & 86.08 & 1417.4 & 64.46 & 57.30 & 63.18 & 34.44 & 62.83 & 97.4 \\
$\alpha{=}1.5$ & 320 & 66.83 & 61.46 & 68.22 & 58.10 & 87.03 & 1464.1 & 64.54 & 57.47 & 64.41 & 35.56 & 63.68 & 98.7 \\
\rowcolor[rgb]{1.0, .92, .80} \textbf{STAR-Pro ($\alpha{=}2$)} & 320 & 67.4 & 61.8 & 67.9 & 58.1 & 87.5 & 1468.1 & 65.4 & 58.3 & 65.3 & 33.8 & 63.89 & 99.0 \\
$\alpha{=}2.5$ & 320 & 67.79 & 58.79 & 68.82 & 52.67 & 88.24 & 1487.8 & 65.05 & 57.73 & 65.09 & 34.89 & 63.35 & 98.1 \\
$\alpha{=}3$   & 360 & 66.61 & 49.34 & 68.37 & 43.81 & 82.40 & 1438.8 & 64.20 & 56.36 & 64.22 & 34.89 & 60.21 & 93.3 \\
\midrule
\multicolumn{14}{c}{\textit{Layer-average budget $T{=}160$ ($\downarrow$ 94.4\%)}} \\
\midrule
$\alpha{=}1$   & 160 & 62.57 & 59.23 & 68.12 & 55.63 & 81.96 & 1362.1 & 62.76 & 54.90 & 63.41 & 34.00 & 61.07 & 94.6 \\
$\alpha{=}1.5$ & 160 & 64.41 & 59.97 & 67.48 & 56.62 & 84.52 & 1401.2 & 63.78 & 56.19 & 62.73 & 34.78 & 62.05 & 96.1 \\
\rowcolor[rgb]{1.0, .92, .80} \textbf{STAR-Pro ($\alpha{=}2$)} & 160 & 65.21 & 60.4 & 67.7 & 56.4 & 86.2 & 1421.9 & 64.4 & 56.9 & 63.3 & 34.3 & 62.59 & 97.0 \\
$\alpha{=}2.5$ & 160 & 65.50 & 57.30 & 68.37 & 50.71 & 87.01 & 1428.0 & 64.12 & 57.39 & 63.47 & 34.78 & 62.01 & 96.1 \\
$\alpha{=}3$   & 180 & 65.01 & 48.82 & 68.62 & 43.61 & 78.30 & 1419.0 & 62.93 & 56.44 & 64.31 & 35.44 & 59.44 & 92.1 \\
\bottomrule
\end{tabular}}
\caption{Candidate-multiplier sweep on \textbf{LLaVA-NeXT-7B} with the two
Progressive-Stage checkpoints fixed at $\{12,20\}$. The highlighted
\textbf{STAR-Pro ($\alpha{=}2$)} row is the deployed configuration; its scores
are taken from the main comparison table (\cref{tab:pruning_comparison_next}).
$T_{\text{avg}}$ is the realized layer-average visual-token count (it exceeds
$T$ only where the strict average constraint is infeasible, at $\alpha{=}3$).
Acc.\ is the 10-benchmark average with MME/20, and Rel.\ (\%) normalizes Acc.\
by the full-token baseline.}
\label{tab:alpha_sweep_llavanext7b}
\end{table*}

\begin{table*}[t]
\centering
\small
\setlength{\tabcolsep}{4pt}
\resizebox{\textwidth}{!}{%
\begin{tabular}{lrrrrrrrrrrrrr}
\toprule
\textbf{Config.} & $T_{\text{avg}}$ & \textbf{SEED} & \textbf{GQA} & \textbf{SQA}$^{\text{IMG}}$ & \textbf{VQA}$^{\text{Text}}$ & \textbf{POPE} & \textbf{MME} & \textbf{MMB}$^{\text{EN}}$ & \textbf{MMB}$^{\text{CN}}$ & \textbf{AI2D} & \textbf{MMMU} & \textbf{Acc.} & \textbf{Rel.} \\
\midrule
\multicolumn{14}{c}{\textit{Layer-average budget $T{=}640$ ($\downarrow$ 77.8\%)}} \\
\midrule
$\alpha{=}1$   & 640 & 69.22 & 63.50 & 72.19 & 60.79 & 86.61 & 1529.1 & 67.26 & 63.75 & 68.17 & 38.00 & 66.59 & 99.4 \\
$\alpha{=}1.5$ & 640 & 70.36 & 64.30 & 72.88 & 61.75 & 86.78 & 1528.3 & 68.11 & 63.14 & 68.91 & 38.00 & 67.06 & 100.1 \\
\rowcolor[rgb]{1.0, .92, .80} \textbf{STAR-Pro ($\alpha{=}2$)} & 640 & 71.33 & 64.2 & 72.8 & 62.2 & 87.4 & 1545.2 & 68.5 & 63.1 & 69.8 & 35.8 & 67.2 & 100.3 \\
$\alpha{=}2.5$ & 640 & 71.38 & 63.72 & 73.03 & 62.30 & 87.00 & 1548.0 & 67.60 & 62.63 & 69.46 & 36.89 & 67.14 & 100.2 \\
$\alpha{=}3$   & 640 & 71.30 & 60.51 & 72.88 & 55.15 & 86.42 & 1554.3 & 67.77 & 62.46 & 69.62 & 35.44 & 65.93 & 98.4 \\
\midrule
\multicolumn{14}{c}{\textit{Layer-average budget $T{=}320$ ($\downarrow$ 88.9\%)}} \\
\midrule
$\alpha{=}1$   & 320 & 66.99 & 62.63 & 71.49 & 59.53 & 85.84 & 1530.3 & 64.97 & 62.54 & 68.04 & 36.89 & 65.54 & 97.8 \\
$\alpha{=}1.5$ & 320 & 68.12 & 63.01 & 71.00 & 60.27 & 86.45 & 1526.6 & 66.67 & 62.54 & 68.36 & 37.78 & 66.05 & 98.6 \\
\rowcolor[rgb]{1.0, .92, .80} \textbf{STAR-Pro ($\alpha{=}2$)} & 320 & 69.19 & 63.4 & 72.0 & 60.6 & 87.2 & 1524.2 & 67.3 & 63.7 & 68.2 & 35.2 & 66.3 & 98.9 \\
$\alpha{=}2.5$ & 320 & 69.85 & 63.42 & 71.89 & 61.10 & 87.05 & 1505.6 & 67.26 & 62.37 & 68.56 & 37.89 & 66.47 & 99.2 \\
$\alpha{=}3$   & 320 & 70.02 & 60.29 & 72.38 & 54.38 & 87.11 & 1502.8 & 67.18 & 62.11 & 68.91 & 37.78 & 65.53 & 97.8 \\
\midrule
\multicolumn{14}{c}{\textit{Layer-average budget $T{=}160$ ($\downarrow$ 94.4\%)}} \\
\midrule
$\alpha{=}1$   & 160 & 64.24 & 59.99 & 71.54 & 57.47 & 82.80 & 1441.4 & 65.82 & 61.86 & 66.74 & 37.67 & 64.02 & 95.6 \\
$\alpha{=}1.5$ & 160 & 65.95 & 61.51 & 72.24 & 58.64 & 84.44 & 1488.0 & 65.39 & 61.94 & 67.55 & 37.56 & 64.96 & 97.0 \\
\rowcolor[rgb]{1.0, .92, .80} \textbf{STAR-Pro ($\alpha{=}2$)} & 160 & 66.94 & 62.3 & 71.4 & 59.3 & 86.0 & 1519.8 & 65.3 & 62.7 & 68.0 & 35.4 & 65.3 & 97.5 \\
$\alpha{=}2.5$ & 160 & 67.13 & 62.08 & 71.05 & 59.15 & 86.01 & 1496.1 & 65.99 & 62.63 & 67.84 & 36.89 & 65.36 & 97.5 \\
$\alpha{=}3$   & 160 & 67.17 & 59.25 & 70.10 & 53.69 & 86.42 & 1506.6 & 65.82 & 62.54 & 67.91 & 37.11 & 64.53 & 96.3 \\
\bottomrule
\end{tabular}}
\caption{Candidate-multiplier sweep on \textbf{LLaVA-NeXT-13B} with the two
Progressive-Stage checkpoints fixed at $\{15,30\}$. The highlighted
\textbf{STAR-Pro ($\alpha{=}2$)} row is the deployed configuration; its scores
are taken from the main comparison table (\cref{tab:pruning_comparison_next13b}).
$T_{\text{avg}}$ is the realized layer-average visual-token count. Acc.\ is the
10-benchmark average with MME/20, and Rel.\ (\%) normalizes Acc.\ by the
full-token baseline.}
\label{tab:alpha_sweep_llavanext13b}
\end{table*}

\section{Additional Details for the Empirical Study}
\label{sec:suppl_empirical}

This section documents the metrics and protocols behind the two empirical-study
panels in the main paper.  The first panel evaluates feature-space coverage
before cross-modal fusion, while the second evaluates how the set of attended
visual tokens changes across decoder depth.

\subsection{Feature-Retention Metric and PCA Protocol}
\label{sec:suppl_metric}

\noindent\textbf{Feature-retention metric.}
For one image, let
$\mathbf{X}=[\mathbf{x}_1,\ldots,\mathbf{x}_n]
\in\mathbb{R}^{d_s\times n}$ contain the $n{=}576$ non-CLS visual-token
features as columns, and let
$\mathbf{X}_{S}$ contain the columns indexed by a selected set $S$.  Let
$r=\operatorname{rank}(\mathbf{X}_S)$ and write its compact singular-value
decomposition as
\[
  \mathbf{X}_S=\mathbf{U}_r\mathbf{\Sigma}_r\mathbf{V}_r^\top,
  \qquad \mathbf{\Sigma}_r=\operatorname{diag}(\sigma_1,\ldots,\sigma_r),
  \quad \sigma_j>0.
\]
The Moore--Penrose pseudoinverse of the Gram matrix is therefore
$(\mathbf{X}_S^\top\mathbf{X}_S)^\dagger
=\mathbf{V}_r\mathbf{\Sigma}_r^{-2}\mathbf{V}_r^\top$, and hence
\[
  \begin{aligned}
  \mathbf{P}_{S}
  &:=\mathbf{X}_{S}
     (\mathbf{X}_{S}^{\top}\mathbf{X}_{S})^{\dagger}
     \mathbf{X}_{S}^{\top}
    =\mathbf{U}_r\mathbf{U}_r^\top,\\
  \mathbf{P}_S^\top&=\mathbf{P}_S,\qquad
  \mathbf{P}_S^2=\mathbf{P}_S,\\
  \operatorname{range}(\mathbf{P}_S)
  &=\operatorname{span}(\mathbf{X}_S).
  \end{aligned}
\]
Thus $\mathbf{P}_S$ is the unique orthogonal projector onto the selected
feature subspace, including when the selected features are linearly dependent.

\noindent\textbf{Least-squares and energy interpretation.}
The pseudoinverse also gives the minimum-Frobenius-norm solution of the
multi-response least-squares problem:
\[
  \begin{aligned}
  \mathbf{A}^\star
  &=\mathbf{X}_S^\dagger\mathbf{X}
    =(\mathbf{X}_S^\top\mathbf{X}_S)^\dagger
      \mathbf{X}_S^\top\mathbf{X},\\
  \mathbf{A}^\star
  &\in\operatorname*{arg\,min}_{\mathbf{A}}
      \|\mathbf{X}-\mathbf{X}_S\mathbf{A}\|_F^2,\\
  \mathbf{X}_S\mathbf{A}^\star&=\mathbf{P}_S\mathbf{X}.
  \end{aligned}
\]
Because $\mathbf{P}_S\mathbf{X}$ and
$(\mathbf{I}-\mathbf{P}_S)\mathbf{X}$ are orthogonal in the Frobenius inner
product, the Pythagorean identity gives
\[
  \|\mathbf{X}\|_F^2
  =\|\mathbf{P}_S\mathbf{X}\|_F^2
   +\|(\mathbf{I}-\mathbf{P}_S)\mathbf{X}\|_F^2.
\]
We therefore define
\[
  \begin{aligned}
  \eta(S)
  &:=\frac{\|\mathbf{P}_{S}\mathbf{X}\|_F^2}
           {\|\mathbf{X}\|_F^2}\\
  &=1-\frac{\displaystyle\min_{\mathbf{A}}
                \|\mathbf{X}-\mathbf{X}_S\mathbf{A}\|_F^2}
               {\|\mathbf{X}\|_F^2}\in[0,1].
  \end{aligned}
\]
Hence $\eta(S)$ is exactly the fraction of total feature energy recoverable by
optimal linear reconstruction from the span of the selected tokens.

\noindent\textbf{Monotonicity and incremental form.}
If $S\subseteq S'$, then
$\operatorname{span}(\mathbf{X}_S)\subseteq
 \operatorname{span}(\mathbf{X}_{S'})$ and
$\mathbf{P}_{S'}-\mathbf{P}_S$ is the orthogonal projector onto the component
of $\operatorname{span}(\mathbf{X}_{S'})$ orthogonal to
$\operatorname{span}(\mathbf{X}_S)$, and is therefore positive semidefinite.
Consequently,
\[
  \eta(S')-\eta(S)
  =\frac{\operatorname{tr}\!\left[
      \mathbf{X}^\top(\mathbf{P}_{S'}-\mathbf{P}_S)\mathbf{X}\right]}
     {\|\mathbf{X}\|_F^2}\geq 0.
\]
For a candidate token $\mathbf{x}_i$, let
$\mathbf{r}_i=(\mathbf{I}-\mathbf{P}_S)\mathbf{x}_i$.  If
$\mathbf{r}_i\neq\mathbf{0}$, set
$\mathbf{q}_i=\mathbf{r}_i/\|\mathbf{r}_i\|_2$.  The enlarged subspace is an
orthogonal direct sum, which yields
\[
  \mathbf{P}_{S\cup\{i\}}
  =\mathbf{P}_S+\mathbf{q}_i\mathbf{q}_i^\top,
  \qquad
  \eta(S\cup\{i\})-\eta(S)
  =\frac{\|\mathbf{q}_i^\top\mathbf{X}\|_2^2}
         {\|\mathbf{X}\|_F^2}.
\]
If $\mathbf{r}_i=\mathbf{0}$, the candidate is already in the selected span
and contributes zero gain.  This establishes the algebraic equivalence between
the closed-form projector and our implementation, which incrementally builds
an orthonormal basis along each method's selection order and accumulates the
captured squared projection energy.

\noindent\textbf{Spectral upper envelope.}
Let $\mathbf{Q}_S$ be any orthonormal basis for the selected span, and order
the singular values of $\mathbf{X}$ nonincreasingly.  By Ky Fan's maximum
principle, every rank-$r$ selected subspace satisfies
\[
  \begin{aligned}
  \eta(S)
  &=\frac{\operatorname{tr}(\mathbf{Q}_S^\top
      \mathbf{X}\mathbf{X}^\top\mathbf{Q}_S)}{\|\mathbf{X}\|_F^2}\\
  &\leq
    \frac{\sum_{j=1}^{r}\sigma_j^2(\mathbf{X})}
         {\|\mathbf{X}\|_F^2}.
  \end{aligned}
\]
The upper bound is attained by the unconstrained leading rank-$r$ left-singular
subspace of $\mathbf{X}$.  In contrast, a token selector is constrained to a
subspace spanned by actual columns of $\mathbf{X}$; for a particular $S$, the
gap to this envelope quantifies its loss relative to the unconstrained
rank-$r$ optimum.

\noindent\textbf{Coverage protocol.}
The coverage curves use raw, uncentered CLIP ViT-L/14 features from the
penultimate hidden layer, before the multimodal projector.  For each token
budget $K$, we take the first $K$ entries of each method's selection order and
report the mean with standard-error bands over 30 MME images.  Thus $\eta$ is a
within-representation coverage diagnostic: it measures how much feature-space
energy is recoverable from the selected span.  It is not an accuracy metric and
should only be compared at the same feature layer, preprocessing, image set,
and token budget.

\noindent\textbf{PCA protocol.}
For the qualitative insets, we use one representative image from the same
30-image cohort.  Writing
$\mathbf{z}_j=\mathbf{x}_j/\|\mathbf{x}_j\|_2$ and
$\bar{\mathbf{z}}=n^{-1}\sum_{j=1}^n\mathbf{z}_j$, we form the centered matrix
$\mathbf{Z}=[\mathbf{z}_1-\bar{\mathbf{z}},\ldots,
\mathbf{z}_n-\bar{\mathbf{z}}]$.  If $\mathbf{U}_2$ contains its two leading
left singular vectors, the displayed coordinate of token $j$ is
\[
  \mathbf{y}_j=\mathbf{U}_2^\top
  (\mathbf{z}_j-\bar{\mathbf{z}})\in\mathbb{R}^2.
\]
This PCA basis is computed once from all 576 tokens and shared across methods;
each inset only marks the first six tokens in the corresponding selection
order.  The insets visualize how quickly each selector reaches distinct
feature groups.  Because PCA uses normalized, centered features whereas
$\eta$ uses the raw, uncentered features above, the insets are qualitative and
are not used in the computation of $\eta$.

\subsection{Selection-Continuity Metric and Protocol}
\label{sec:suppl_continuity}

We analyze LLaVA-1.5-7B, which has 32 decoder layers, on 500 MME images.  At
each layer $\ell$, let $S_\ell$ be the $K{=}64$ visual tokens with the highest
text-to-visual decoder-attention scores.  For two layers, set agreement is
measured by intersection over union,
\[
  \operatorname{IoU}(S_i,S_j)
  =\frac{|S_i\cap S_j|}{|S_i\cup S_j|}.
\]
For a layer gap $d$, the continuity score at layer $\ell$ is
\[
  C_\ell(d)=\frac{1}{2}\left[
    \operatorname{IoU}(S_\ell,S_{\ell-d})+
    \operatorname{IoU}(S_\ell,S_{\ell+d})
  \right].
\]
At boundary layers where only one of the two neighbors exists, we use that
available side rather than dividing it by two.  We first average the complete
$32{\times}32$ pairwise-IoU matrix over the 500 images, then report the
per-layer curves for $d\in\{1,2,3,4,5\}$ and their mean.  A value below one
means that the membership of the attended token set changes with depth; the
metric compares set membership, not attention magnitudes or task accuracy.

\end{document}